\documentclass{article}

\usepackage[preprint]{neurips_2025}

\usepackage[T1]{fontenc}
\usepackage[utf8]{inputenc}
\usepackage{amssymb}
\usepackage[pdftex]{graphicx}
\usepackage[export]{adjustbox}
\usepackage{color}
\usepackage{enumitem}
\usepackage{geometry}
\usepackage{bm}
\usepackage{cancel}
\usepackage{xcolor}
\usepackage{mathrsfs,amsmath}
\usepackage{hyperref}
\hypersetup{hypertex=true, colorlinks=true, linkcolor=blue, anchorcolor=blue, citecolor=blue,urlcolor=blue}

\usepackage{url}            
\usepackage{subfigure}
\usepackage{relsize}
\usepackage{placeins}
\usepackage{url}            
\usepackage{booktabs}       
\usepackage{amsfonts}       
\usepackage{nicefrac}       
\usepackage{microtype}      
\usepackage{xcolor}         

\newcommand{\mlg}[1]{\mathlarger{#1}}
\newcommand{\mlgr}[1]{\textcolor{red}{\mathlarger{#1}}}
\newcommand{\mlgg}[1]{\textcolor{blue}{\mathlarger{#1}}}
\newcommand{\eps}{\varepsilon}
\newcommand{\bah}{\bar{\alpha}}

\title{Rethinking Diffusion Model in High Dimension}

\author{Zhenxin Zheng \\ MTLab, Meitu Inc. \\ \texttt{blair.star@163.com} \And Zhenjie Zheng \\ Civil Engineering, University of Hong Kong \\ \texttt{zjzheng@connect.hku.hk}}

\begin{document}
	
	\maketitle
	
	\begin{abstract}
		\textbf{Curse of Dimensionality} is an unavoidable challenge in statistical probability models, yet diffusion models seem to overcome this limitation, achieving impressive results in high-dimensional data generation. Diffusion models assume that they can learn the statistical quantities of the underlying probability distribution, enabling sampling from this distribution to generate realistic samples. But is this really how they work? We argue not, based on the following observations: 1) In high-dimensional sparse scenarios, the fitting target of the diffusion model's objective function degrades from a \textbf{weighted sum of multiple samples} to a \textbf{single sample}, which we believe hinders the model’s ability to effectively learn essential statistical quantities such as posterior, score, or velocity field. 2) Most inference methods can \textbf{be unified within a simple framework} which involves no statistical concepts, aligns with the degraded objective function, and provides an novel and intuitive perspective on the inference process. \textit{Code is available at https://github.com/blairstar/NaturalDiffusion.}
	\end{abstract}
	
	\section{Introduction} \label{sec:intro}
	Diffusion models exhibit remarkable competitiveness in high-dimensional data generation scenarios, particularly in image generation \cite{rombach2022high}. Beyond delivering outstanding performance, diffusion models also possess various elegant mathematical formulations. \cite{sohl2015deep} first introduced the diffusion model approach, which utilizes a Markov chain to transform complex data distributions into simple normal distributions and then learns the posterior probability distribution corresponding to the transformation process. \cite{ho2020denoising} refined the objective function of diffusion models and introduced Denoised DPM. \cite{song2020score} generalized the noise addition process of diffusion models from discrete to continuous and formulated it as a stochastic differential equation (SDE). \cite{lipman2022flow} proposed a new optimization perspective based on flow matching, enabling the model to directly learn the velocity field of probability flows.
	
	All three of the aforementioned models assume that the diffusion model can learn the statistical quantities of the data distribution. In the Markov Chain formulation, it is assumed that the model can learn the posterior probability distribution. In the SDE formulation, it is assumed that the model can learn the score of the marginal distribution. In the flow matching approach, it is assumed that the model can learn the velocity field. However, this assumption contradicts conventional understandings. Traditionally, it is believed that in high-dimensional sparse scenarios, machine learning models cannot effectively learn complex hidden probability distributions and their essential statistical quantities. This discrepancy prompts a fundamental inquiry: \textbf{This discrepancy raises a fundamental question: Do diffusion models truly learn these complex distributions and their statistical quantities as theoretically assumed? If not, why are they still able to generate high-quality samples? Could it be that diffusion models operate via a different underlying mechanism?}
	
	\textbf{We argue that diffusion models do not learn these statistical quantities; instead, they operate via a different mechanism}.

	To support this conclusion, this paper provides a detailed analysis of the objective function and inference methods of diffusion models.

	Section~\ref{section:sparsity} focuses on the objective function. We identify a phenomenon that emerges in high-dimensional spaces: due to data sparsity, the fitting target of the diffusion model’s objective function \textbf{degrades from a weighted sum to a single sample}. Under such conditions, we argue that the model cannot effectively learn the essential statistical quantities of the underlying data distribution, including the posterior, score, and velocity field.

	Section~\ref{section:infer_framework} focuses on the inference method. We propose a novel inference framework that not only aligns with the degraded objective function but also unifies most existing inference methods, including DDPM Ancestral Sampling, DDIM \citep{song2020denoising}, Euler, DPM-Solver \citep{lu2022dpm}, DPM-Solver++ \citep{lu2022dpmpp}, and DEIS \citep{zhang2022fast}. Furthermore, this framework provides an entirely new and intuitive perspective for understanding the inference process, without relying on any statistical concepts.

	Section~\ref{section:application} focues on the applications of the new inference framework. By leveraging this framework, we are able to reinterpret several key phenomena, such as why higher-order inference methods often exhibit acceleration. Moreover, within this framework, it is possible to search for other, more optimal parameter configurations that yield better generative performance.

	This work makes the following key contributions: 
	\begin{itemize}
    \item We present the \textbf{first rigorous analysis} of the diffusion model objective in high-dimensional sparse scenarios, demonstrating that its fitting target \textbf{degrades from a weighted sum of multiple samples to a single sample}. This degradation prevents the model from effectively capturing the underlying data distribution and its associated statistical quantities (posterior, score, velocity field). 
    \item We further introduce a novel inference framework that \textbf{unifies most existing inference methods}, encompassing both stochastic and deterministic approaches. This framework provides an entirely new way of understanding the inference process—free from any reliance on statistical concepts—while remaining fully consistent with the degraded objective function. 
    \item Taken together, these contributions offer a \textbf{complete and fundamentally new perspective} on high-dimensional diffusion models, covering both their training objectives and inference mechanisms. This perspective is simple, intuitive, and free from statistical concepts, opening up a promising new direction for advancing diffusion models in high-dimensional settings. 
	\end{itemize}	
	
	\section{Background} \label{section:background}
	Given a batch of sampled data \({X^{0}_{0}, X^{1}_{0}, \dots, X^{N}_{0}}\) from the random variable \(X_0\), the diffusion model mixes the data with random noise in different proportions, forming a sequence of new variables \({X_1, X_2, \cdots, X_T}\). The signal-to-noise ratio (SNR), which represents the ratio of data to noise, gradually decreases, and by the final variable \(X_T\), it almost consists entirely of random noise.
	
	For the original diffusion model and VP SDE, they mix in the following way:
	\begin{align} \label{eq_vp_merge}
		X_t = \sqrt{\bah_t}\cdot X_0 + \sqrt{1-\bah_t} \cdot \varepsilon
	\end{align} 
	Where \(\bah_t\) gradually decreases from 1 to 0, and \(t\) takes discrete values from 1 to \(T\).
	
	For flow matching, it mixes in the following way:
	\begin{align} \label{eq_flow_merge}	
		X_t = (1-\sigma_t) \cdot X_0 +  \sigma_t \cdot \varepsilon
	\end{align}
	Where \(\sigma_t\) also gradually increases from 0 to 1, and in practice, \(\sigma_t = t\) is often set. \(t\) takes continuous values, \(t \in [0, 1]\).
	
	\paragraph{Markov Chain-based diffusion model} For the Markov Chain-based diffusion model, its core lies in learning the conditional posterior probability \(p(x_{t-1}|x_t)\). Since the posterior probability is approximately a Gaussian function, and its variance is relatively fixed, we can focus on learning the mean of the conditional posterior probability \(E_{p(x_{t-1}|x_t)}\left(x_{t-1}\right)\). According to the Total Law of Expectation\cite{ross2010first}, the mean can be expressed in another form:
	\begin{align}
		E_{p(x_{t-1}|x_t)}\left(x_{t-1}\right) = \int p(x_0|x_t)\  E_{p(x_{t-1}|x_0,x_t)}\left(x_{t-1}\right)dx_0
	\end{align}
	As seen from equation (7) in \cite{ho2020denoising}, the mean of \(p(x_{t-1}|x_0,x_t)\) can be expressed as a linear combination of \(x_0\) and \(x_t\), i.e.	
	\begin{align}
		  E_{p(x_{t-1}|x_0,x_t)}\left(x_{t-1}\right) = \underbrace{\frac{\sqrt{\bah_{t-1}}\beta_t}{1-\bah_t}}_{const=C_0} \cdot x_0 + \underbrace{\frac{\sqrt{\alpha_t}(1-\bah_{t-1})}{1-\bah_t}}_{const=C_t} \cdot x_t 
	\end{align}
	Thus, the mean of \(p(x_{t-1}|x_t)\) can be further expressed as
	\begin{align}
		E_{p(x_{t-1}|x_t)}\left(x_{t-1}\right) = \int p(x_0|x_t) \left(C_0\cdot x_0 + C_t \cdot x_t\right) dx_0 = C_0 \int p(x_0|x_t)\ x_0\ dx_0 + C_t \cdot x_t
	\end{align}
	Therefore, the objective of Markov Chain-based diffusion model can be considered as learning the mean of \(p(x_0|x_t)\), i.e.
	\begin{align}
		\min_\theta \int p(x_t) \left\|f_\theta(x_t) - \int p(x_0|x_t)\ x_0\ dx_0\right\|^2 dx_t
	\end{align}
	where \(f_\theta(x_t)\) is a learnable neural network function, with input \(x_t\).
	
	\paragraph{Score-based diffusion model} For the score-based diffusion model, its core lies in learning the score of the marginal distribution \(p(x_t)\) \(\left(\frac{\partial \log p(x_t)}{\partial x_t}\right)\). Similar to the Markov chain-based diffusion model, by introducing another variable \(X_0\), the score can be expressed in another form:	 
	\begin{alignat}{2}
		\frac{\partial \log p(x_t)}{\partial x_t} = \int p(x_0|x_t)\ \frac{\partial \log p(x_t|x_0)}{\partial x_t} dx_0
	\end{alignat}
	The proof of this relationship can be found in Appendix \ref{section:conditional_score}.
	
	Since \(p(x_t|x_0) \sim \mathcal{N}\left(x_t; \sqrt{\bah_t}x_0, \sqrt{1-\bah_t}\right)\), the score of \(p(x_t|x_0)\) can be expressed as
	\begin{align}
		\frac{\partial \log p(x_t|x_0)}{\partial x_t} = -\frac{x_t - \sqrt{\bah_t}x_0}{1-\bah_t} =  \underbrace{\frac{\sqrt{\bah_t}}{1-\bah_t}}_{const=S_0}\cdot x_0 + \underbrace{\frac{-1}{1-\bah_t}}_{const=S_t}\cdot x_t 
	\end{align}
	Thus, the score of \(p(x_t)\) can be expressed as
	\begin{align}
		\frac{\partial \log p(x_t)}{\partial x_t} = \int p(x_0|x_t) ( S_0 \cdot x_0 + S_t \cdot x_t)\ dx_0 = S_0 \int p(x_0|x_t)\ x_0\ dx_0 + S_t \cdot x_t
	\end{align}
	Therefore, the objective of score-based diffusion model can also be considered as learning the mean of \(p(x_0|x_t)\).
	
    \paragraph{Flow Matching-based diffusion model} The core of the flow matching-based diffusion model lies in learning the velocity field of the probability flow. According to Theorem 1 in \cite{lipman2022flow}, the velocity field \(u(x_t)\) can be expressed as a weighted sum of the conditional velocity field \(u(x_t|x_0)\), i.e.
	\begin{align}
		u(x_t) = \int p(x_0|x_t) u(x_t|x_0) dx_0
	\end{align}
	From equation \ref{eq_flow_merge}, we know that the conditional velocity field \(u(x_t|x_0)\) is
	\begin{align}
		u(x_t|x_0) \triangleq \frac{\mathrm{d}x_t}{\mathrm{d}t} = \varepsilon - x_0
	\end{align}
	Thus, the velocity field \(u(x_t)\) can be expressed as
	\begin{align}
		u(x_t) = \int p(x_0|x_t) (\varepsilon - x_0) dx_0 = \varepsilon - \int p(x_0|x_t) x_0 dx_0
	\end{align}
	Therefore, the objective of flow matching-based diffusion model can also be considered as learning the mean of \(p(x_0|x_t)\).
	
	\paragraph{Equivalent to predicting $X_0$} Fitting the mean of \(p(x_0|x_t)\) is equivalent to \textbf{predicting \(X_0\)}, i.e.
	\begin{align*}
		\min_\theta \int p(x_t) \left\|f_\theta(x_t) - \int p(x_0|x_t)\ x_0\ dx_0\right\|^2 dx_t \iff \min_\theta \iint p(x_0, x_t) \left\|f_\theta(x_t) - \ x_0\right\|^2 dx_0 dx_t 
	\end{align*}
	The specific proof can be found in Appendix \ref{section:mean_eq_x0}. The integrals above cannot be computed exactly and are typically approximated using Monte Carlo integration. In practice, the required samples are typically obtained via \textbf{Ancestral Sampling}. The detailed procedure is as follows: sample \(X_0\) from \(p(x_0)\), and then sample \(X_t\) from \(p(x_t|x_0)\). The pair \((X_0, X_t)\) follows the joint distribution \(p(x_0, x_t)\), and the individual \(X_t\) follows \(p(x_t)\), and the individual \(X_0\) follows \(p(x_0|x_t=X_t)\).

	\section{Impact of sparsity on the objective function} \label{section:sparsity}
	We first show the form of the posterior probability distribution \(p(x_0|x_t)\).
	
	\subsection{Form of the posterior $p(x_0|x_t)$}
	For convenience, we use a unified form to represent the two mixing ways in Eq. \eqref{eq_vp_merge} and Eq. \eqref{eq_flow_merge} as follows: $x_t = c_0 \cdot x_0 + c_1 \cdot \varepsilon $. When \(c_0^2 + c_1^2 = 1\), this represents the mixing way of Markov Chain-based   and Score-based diffusion model. When \(c_0 + c_1 = 1\), it represents the Flow Matching mixing way. Under this representation, \(p(x_t|x_0) \sim \mathcal{N}(x_t; c_0x_0, c_1^2)\).
	
	From the analysis in Appendix \ref{section:posterior_form}, the posterior \(p(x_0|x_t)\) has the following form:
	\begin{align}	\label{eq_pos_x0}
		p(x_0|x_t) &= \text{Normalize}\left(\exp{\frac{-(x_0 - \mu)^2}{2\sigma^2}}\ p(x_0)\right) \qquad \text{where}\ \mu= \frac{x_t}{c_0} \quad \sigma = \frac{c_1}{c_0}
	\end{align}
	Here, \(p(x_0)\) is the hidden data distribution, which is unknown and cannot be sampled directly. It can only be randomly selected from the existing samples \(\{X_0^0, X_0^1, \dots, X_0^N\}\) (\(X_0^i \sim p(x_0)\)). The selection process can be considered as sampling from the following mixed Dirac delta distribution:$ p(x_0) = \frac{1}{N}\sum_{i=0}^{N}{\delta\left(x_0 - X^i_0\right)}$. Substituting this into Equation \eqref{eq_pos_x0}, we get:
	\begin{align}
		p(x_0|x_t) = \frac{1}{Z_c} \exp{\frac{-(x_0 - \mu)^2}{2\sigma^2}} \sum_{i=0}^{N}{\delta\left(x_0 - X^i_0\right)}
	\end{align}
	Here, $\mu = \frac{x_t}{c_0}$, $\sigma = \frac{c_1}{c_0}$, and $Z_c$ is normalization factor. It can be seen that when \(p(x_0)\) is discrete, \(p(x_0|x_t)\) is also discrete, and the probability of each discrete value \(X_0^i\) is \textbf{inversely} proportional to \textbf{the distance between \(X_0^i\) and \(\mu\)}. A similar conclusion is also presented in Appendix B of \cite{karras2022elucidating}, although the derivation method differ.
	
	\subsection{Weighted Sum Degradation phenomenon} \label{section:degradation}

	we further analyze the characteristics of the mean of $p(x_0|x_t)$. According to the definition of expectation, the mean of $p(x_0|x_t)$ can be expressed as:
	\begin{align}
		\int &\ x_0\ p(x_0|x_t) dx_t = \frac{1}{Z_c} \sum_{i=0}^{N} {X^i_0\ \exp{\frac{-(X^i_0 - \mu)^2}{2\sigma^2}}}
	\end{align}
	The mean of $p(x_0|x_t)$ is a weighted sum of all $X^i_0$ samples, and the weight is inversely proportional to the distance between $X^i_0$ and $\mu$.  If one sample is much closer than all others, the weighted sum degrades to that single sample. This is more likely with sparse data.
	
	Figure \ref{f_sparse} presents an example with sparse data ($X_0$, \textcolor{blue}{blue}), and small noise std (\textcolor{green}{green} circle). In this case, most of $X_t$ remain near its origin data sample. This make $p(x_0|x_t)$ highly peaked at the closest $X_0$, causing its mean to degrade from a weighted sum to that single sample. We call this phenomenon \textbf{weighted sum degradation} and argue it potentially hinders the model learning the true data distribution.
	
	Next, we analyze \emph{weighted sum degradation} for conditional ImageNet-256 and ImageNet-512\cite{deng2009imagenet}. Both datasets have high pixel dims (196608 and 786432) and retain high latent dims (4096 and 16480) after VAE\cite{kingma2013auto, rombach2022high} compression. As compression is typical, we will only consider the compressed case below.
	
	We calculate the proportion of degradation. First, we sample $X_t$ as in training (first randomly select an $X_0$, then sample $X_t$ from $p(x_t|x_0=X_0)$). Then, we determine whether $p(x_0|x_t = X_t)$ is degraded. If there exists an $X^\prime_0$ such that $p(x_0 = X^\prime_0 | x_t = X_t) > 0.9$, then we consider \emph{weighted sum degradation} to be present; if $X^\prime_0 = X_0$, it is called \emph{weighted sum degradation to $X_0$}.
	
	Since noise level also affects weighted sum degradation, we calculate degradation rates separately for different $t$. We calculate the proportions of both \emph{weighted sum degradation} and \emph{degradation to $X_0$} under two noise mixing schemes: VP (Equation \eqref{eq_vp_merge}) and Flow Matching (Equation \eqref{eq_flow_merge}).

	\begin{table}
		\centering
		\caption{Statistics of ImageNet-256(weighted sum degradation / weighted sum degradation to $X_0$)}
		\label{table:img256}
		\begin{adjustbox}{width=1\textwidth}
		\begin{tabular}{c|cccccccc}
			\textbf{merging\textbackslash{}time} & \textbf{200} & \textbf{300} & \textbf{400} & \textbf{500} & \textbf{600} & \textbf{700} & \textbf{800} & \textbf{900} \\ \hline
			\textbf{vp}                          & 1.00/1.00    & 1.00/1.00    & 1.00/0.98    & 0.91/0.57    & 0.41/0.01    & 0.02/0.00    & 0.00/0.00    & 0.00/0.00    \\
			\textbf{flow}                        & 1.00/1.00    & 1.00/1.00    & 1.00/1.00    & 1.00/1.00    & 1.00/0.95    & 0.97/0.69    & 0.76/0.15    & 0.09/0.00    \\
		\end{tabular}
		\end{adjustbox}
	\end{table}
	\begin{table}
		\centering
		\caption{Statistics of ImageNet-512(weighted sum degradation / weighted sum degradation to $X_0$)}
		\label{table:img512}
		\begin{adjustbox}{width=1\textwidth}
		\begin{tabular}{c|cccccccc}
			\textbf{merging\textbackslash{}time} & \textbf{200} & \textbf{300} & \textbf{400} & \textbf{500} & \textbf{600} & \textbf{700} & \textbf{800} & \textbf{900} \\ \hline
			\textbf{vp}                          & 1.00/1.00    & 1.00/1.00    & 1.00/0.98    & 0.98/0.57    & 0.87/0.08    & 0.50/0.00    & 0.03/0.00    & 0.00/0.00    \\ 
			\textbf{flow}                        & 1.00/1.00    & 1.00/1.00    & 1.00/1.00    & 1.00/1.00    & 1.00/0.94    & 0.99/0.67    & 0.95/0.20    & 0.71/0.01    \\
		\end{tabular}
		\end{adjustbox}
	\end{table}
	Tables \ref{table:img256} and \ref{table:img512} present statistics for both datasets, showing several clear patterns:
	\begin{itemize}[]
		\item As $t$ decreases, the \emph{weighted sum degradation} phenomenon becomes more pronounced.
		\item The degradation rate of Flow Matching is higher than that of VP.
		\item The higher the dimension, the greater the proportion of degradation.
	\end{itemize}
	Besides, we observe severe degradation in both datasets for both VP and Flow Matching, especially for $t<600$. Furthermore, due to limited sampling during training, each $p(x_0|x_t=X_t)$ cannot be sufficiently sampled, so \textbf{the actual degradation ratio should be higher than the statistics show}.
	
	In high dimensions, each $p(x_0|x_t=X_t)$ should be complex. When \emph{weighted sum degradation} occurs, it is equivalent to using a single sample as an estimator of the mean, which typically have large error. If we cannot provide an accurate fitting target, we argue that the model is unlikely to learn the ideal target accurately. Therefore, \textbf{we argue that diffusion models are uncapable to effectively learn the underlying data distribution, and it is necessary to re-examine their operational mechanisms}.
		
	\subsection{A simple way to understand the objective function} \label{section:new_target_view}
	
	As shown previously, weighted sum degradation is significant in high dimensions, which reduces the fitting target to the original data sample ($X_0$). Therefore, we can understand the objective in a simple way:\textbf{predict the original data sample ($X_0$) from the noise-mixed sample ($X_t$) }.	
	
	From the perspective of the frequency spectrum, we can further understand the principle \cite{dieleman2024spectral}.

\begin{figure}[!ht]
		\begin{minipage}{0.48\linewidth}
			\includegraphics[width=1.0\textwidth]{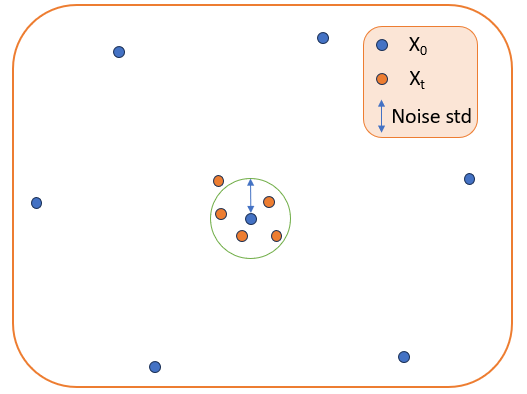} \caption{Impact of data sparsity on posterior probability distribution} \label{f_sparse}
		\end{minipage}
		\hfill
		\begin{minipage}{0.40\linewidth}
			\includegraphics[width=1.0\textwidth]{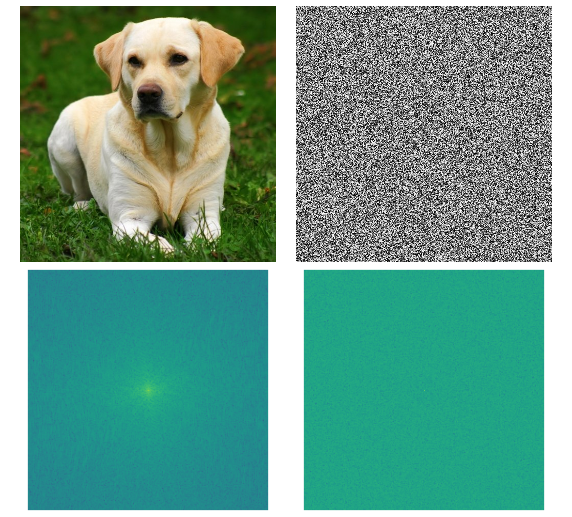}\caption{Left: Natural image and its spectrum. Right: noise and its spectrum} \label{f_spec}
		\end{minipage}
	\end{figure}

\begin{figure}[!ht]
		\begin{minipage}{0.46\linewidth}
			\includegraphics[width=1.0\textwidth]{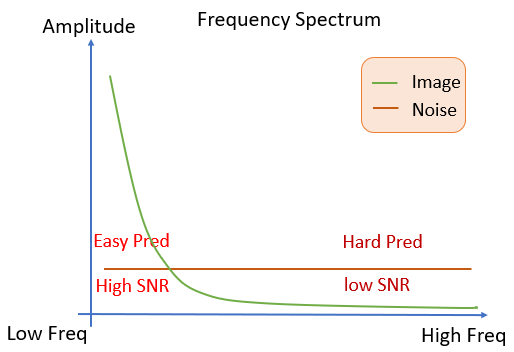}\caption{Image and noise frequency spectrum} \label{figure:freq_spec}
		\end{minipage}
		\hfill
		\begin{minipage}{0.43\linewidth}
			\includegraphics[width=1.0\textwidth]{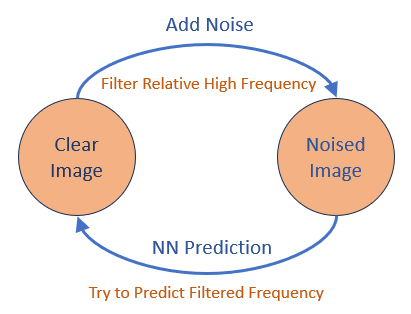}\caption{New perspective of training object function} \label{figure:new_target_view}
		\end{minipage}
	\end{figure}

	As seen in Figure \ref{f_spec}, natural image spectra concentrate energy in low frequencies (bright centrally, dark peripherally), while noise have a uniform spectrum. Thus, when mixed with noise, high frequencies always have lower SNR (signal-noise-ratio) than low frequencies. As noise grows, high frequencies are submerged first, then low frequencies (Figure \ref{figure:freq_spec}).

	When training a model to predict $X_0$ from noise-mixed samples, the model prioritizes frequencies based on their SNR. It easily predicts non-submerged frequencies (likely copying them).For submerged frequencies, it prioritizes predicting the lower-frequency components, as they have relatively higher SNR and larger amplitudes (giving them more weight in the Euclidean loss).

	Thus, the objective can be further understood as \textbf{filtering higher-frequency components – completing the filtered frequency components} (Figure \ref{figure:new_target_view}). At large $t$, even some low frequencies are submerged, so the model prioritizes predicting low frequencies. At small $t$, only high frequencies are submerged, and the model works on predicting these details.  This frequency-dependent process is confirmed during inference: early steps (large $t$) generate contours, while later steps (small $t$) add details. Since the model compensates for the submerged frequencies, it can also be regarded as an \textbf{information enhancement operator}.	

	\section{A unified inference framework-Natural Inference} \label{section:infer_framework}
	We know that the inference methods of diffusion models rely on an assumption that the model can learn the hidden probability distributions or statistical quantities. However, as pointed out in the Section~\ref{section:degradation}, in high-dimensional spaces the degradation phenomenon prevents the model from effectively learning these quantities. Therefore, it is necessary to attempt to reinterpret existing inference methods from a new perspective. Moreover, we have also seen in the Section~\ref{section:new_target_view} that the degraded objective function can be understood in a simple way - predicting the original image $x_0$ from a noisy image $x_t$. Based on the principle of \textbf{train-test matching}, this naturally leads us to ask: can current inference methods also be understood in a similar, simpler way?
	
	The answer is yes. Below, we will reveal that most of inference methods can be unified into a simple framework based on predicting $x_0$, including Ancestor Sampling, DDIM, Euler, DPMsolver, DPMSolver++, DEIS, and Flow Matching solvers, among others.
	
	We first introduce a class of key operations contained in the new framework.
	
	\subsection{Self Guidence} \label{section:self_guide}
	
	Following the concept of Classifier Free Guidance \cite{ho2022classifier}, we introduce a new operation called Self Guidance. The principle of Classifier Free Guidance can be summarized as follows:
	\begin{align} \label{eq_guide_first}
		I_{out} = I_{bad} + \lambda \cdot (I_{good} - I_{bad})
	\end{align}	
	Where $I_{bad}$ is the output of a less capable model, $I_{good}$ is the output of a more capable model, and both models share the same input. $\lambda$ controls the degree of guidance.
	
	In fact, Classifier Free Guidance is somewhat similar to Unsharp Masking algorithm in traditional image enhancing processing\cite{gonzalez2017digital,scikitimage}. In Unsharp Masking algorithm, $I_{good}$ is the original image, and $I_{bad}$ is the image after Gaussian blur. The term $(I_{good} - I_{bad})$ provides the edge information, which, when added to the original image $I_{good}$, results in an image with sharper edges. Therefore, Classifier Free Guidance can also be considered as an \textbf{image enhancement} operation.
	
	In the diffusion model inference process, a series of predicted $x_0$ are generated, where the quality of $x_0$ starts poor and improves over time. If an earlier predicted $x_0$ is used as $I_{bad}$ and a later predicted $x_0$ is used as $I_{good}$, then in this paper, we refer to this operation as \textbf{Self Guidance}, because both $I_{bad}$ and $I_{good}$ are outputs of the same model, and no additional model is needed.
	
\begin{figure}
		\centering
		\begin{subfigure}[]{\includegraphics[width=0.16\textwidth]{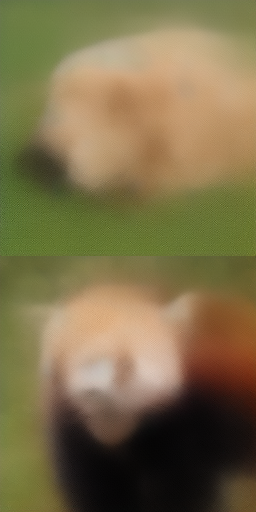}} \end{subfigure}
		\begin{subfigure}[]{\includegraphics[width=0.16\textwidth]{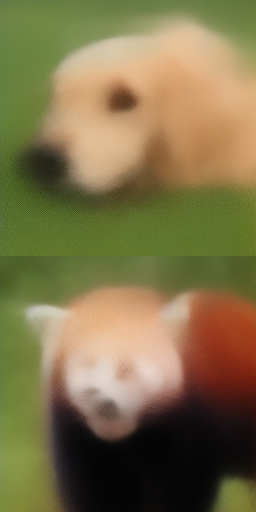}} \end{subfigure}
		\begin{subfigure}[]{\includegraphics[width=0.16\textwidth]{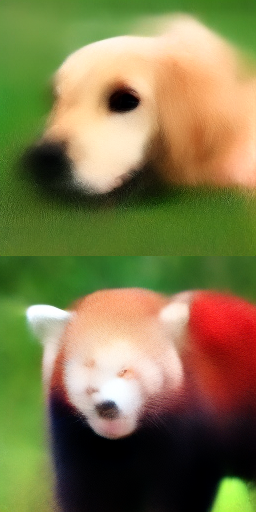}} \end{subfigure}
		\begin{subfigure}[]{\includegraphics[width=0.16\textwidth]{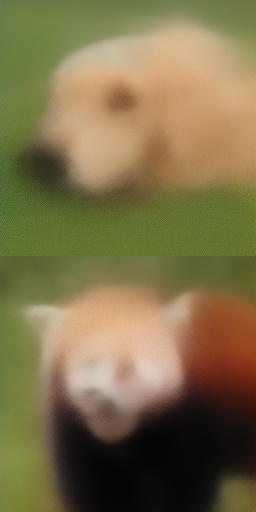}} \end{subfigure}
		\begin{subfigure}[]{\includegraphics[width=0.16\textwidth]{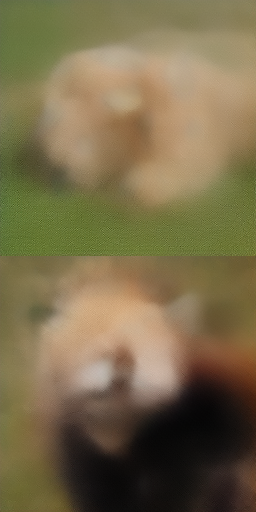}} \end{subfigure}
		\caption{(a) Model output on t=540 ($I_{bad}$) \quad (b) Model output on t=500 ($I_{good}$) \quad (c) Output of Fore Self Guidence \quad (d) Output of Mid Self Guidence \quad (e) Output of Back Self Guidence }

		\label{figure:Self Guidence}
	\end{figure}

	Based on the value of $\lambda$, we further classify Self Guidance as follows:
	\begin{itemize}[]
		\item When $\lambda > 1$, it is called \textbf{Fore Self Guidance}, where the output improves the quality. See Fig. \ref{figure:Self Guidence}(c).
		\item When $0 < \lambda < 1$, it is called \textbf{Mid Self Guidance}, where the output is a linear interpolation between $I_{bad}$ and $I_{good}$, with a quality worse than $I_{good}$ but better than $I_{bad}$. See Fig. \ref{figure:Self Guidence}(d).
		\item When $\lambda < 0$, it is called \textbf{Back Self Guidance}, where the output is not only worse than $I_{good}$, but also worse than $I_{bad}$. See Fig. \ref{figure:Self Guidence}(e). 
	 \end{itemize}
	 
	 As shown in Appendix \ref{section:self_guide_appendix}, \textbf{the linear combination of any two model outputs can be viewed as a single Self Guidance, while the linear combination of multiple model outputs can be viewed as a composition of multiple Self Guidances}.
	 
	\subsection{Natural Inference}

	\begin{figure}[h]
		\centering \includegraphics[width=1.0\textwidth]{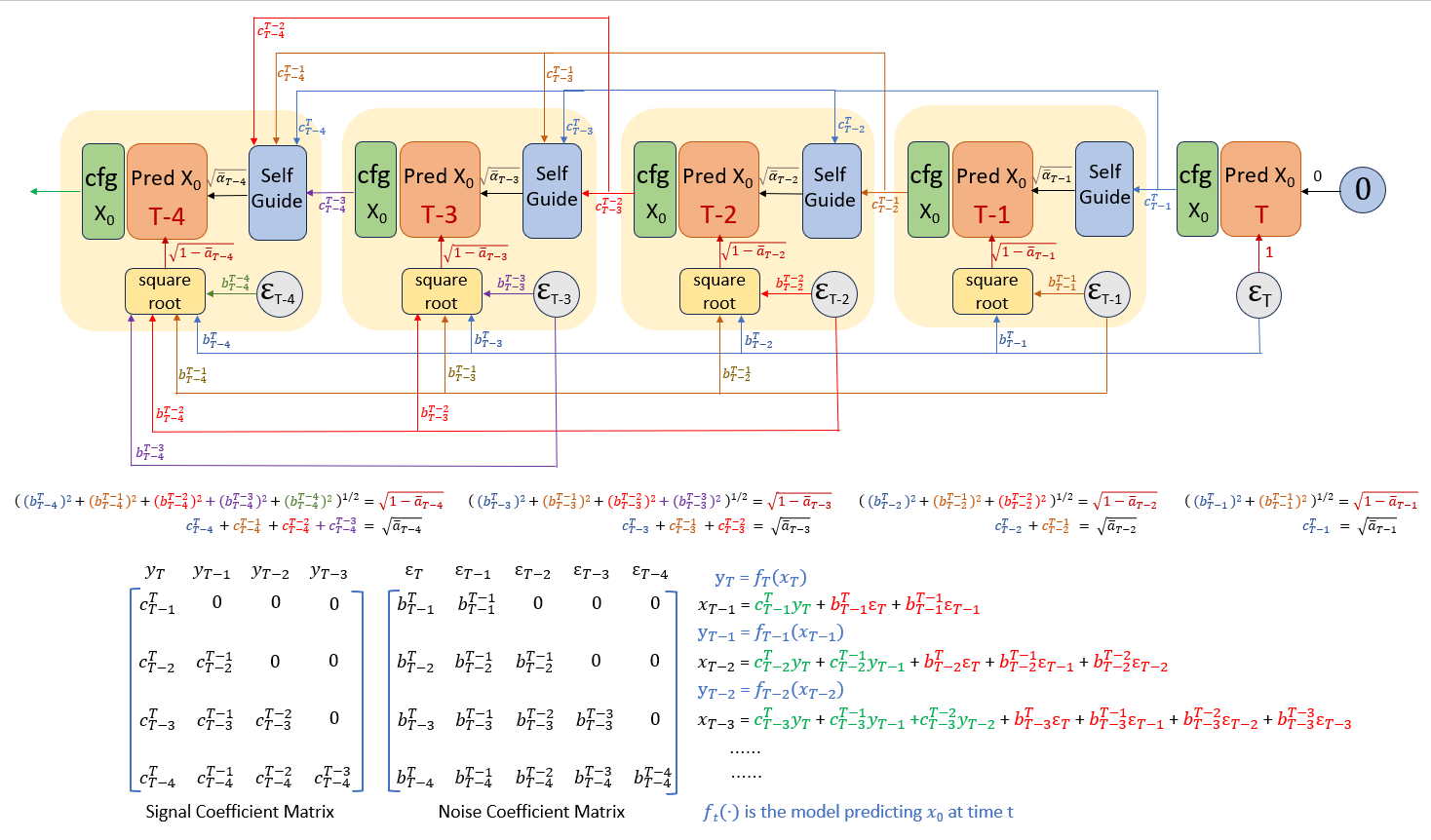} \centering \caption{A new inference framework - Natural Inference} \label{f_infer_scheme}
	\end{figure}
	
	The new inference framework is illustrated in Figure~\ref{f_infer_scheme}, with the core ideas summarized as follows:

\begin{itemize}
	\item It consists of $T$ models that predict $X_0$; 
	\item Each model takes two part inputs: signal (image) and noise;
	\item The image signal is a linear combination of outputs from previous models, while the noise is a linear combination of previous noise and newly added noise; \item At time $t$, the sum of the coefficients corresponding to the image signal ($\sum_{i=t+1}^{T}\mlg{c}^i_{t}$) equals $\sqrt{\bah_{t}}$, and the square root of the sum of the squared noise coefficients ($\sqrt{\sum_{i=t}^T(\mlg{b}^i_{t})^2}$) equals $\sqrt{1 - \bah_{t}}$. This means that \textbf{the magnitudes of the signal and noise remain consistent with those used during the training phase}.
\end{itemize}

	As shown in Section~\ref{section:self_guide}, \textbf{the linear combination of image signals can be interpreted as a composition of multiple Self Guidance operations}. The linear combination of independent noise is still noise\cite{Taboga2021Marco}. Since the input signal of each model depends only on the output signals of previous models, the inference framework exhibits an \textbf{\textcolor{red}{autoregressive}} structure.

	In this paper, we refer to $\sqrt{\bah_{t}}$ as the \textbf{marginal signal coefficient}, and $\sqrt{1-\bah_{t}}$ as the \textbf{marginal noise coefficient}. The term $\sum_{i=t+1}^{T}\mlg{c}^i_{t}$ is referred to as the \textbf{equivalent marginal signal coefficient}, and $\sqrt{\sum_{i=t}^T(\mlg{b}^i_{t})^2}$ is called the \textbf{equivalent marginal noise coefficient}. For clarity, all coefficients are organized into matrix form, as shown in the lower part of Figure~\ref{f_infer_scheme}. Due to the autoregressive property, the signal coefficient matrix has a lower triangular structure.

	\subsection{Represent sampling methods with Natural Inference Framework}
	
	This section briefly demonstrates how various sampling methods can be reformulated within the Natural Inference Framework. For more detailed explanations, please refer to Appendix \ref{section:represent}.

	For first-order sampling methods (including DDPM, DDIM, ODE Euler, SDE Euler, and Flow Matching ODE Euler), their iterative procedures can all be expressed in the following form:
	\begin{align} 
			\mlg{y}_t &= \mlg{f}_t(\mlg{x}_t) \\
			 \mlg{x}_{t-1} &= \textcolor{blue}{\mlg{d}_{t-1}} \cdot \mlg{x}_{t} + \textcolor{blue}{\mlg{e}_{t-1}} \cdot \mlg{y}_{t} + \textcolor{blue}{\mlg{g}_{t-1}} \cdot \mlg{\eps_{t-1}}
	\end{align} 
	Here, $f_t$ is the model function predicting $x_0$ at step $t$. $x_t$, $y_t$, and $\epsilon_{t-1}$ are vectors, while $d_{t-1}$, $e_{t-1}$, and $g_{t-1}$ are fixed scalars. For deterministic methods, $g_{t-1}$ is zero.

	Starting from $x_T$, we can iterate according to the above equation to further determine the expressions of $x_{T-1}$, $x_{T-2}$, $\cdots$, $x_1$, and $x_0$. Each $x_t$ can be represented as two components: one is a linear combination of $\{y_i\}_{i=t+1}^T$, and the other is a linear combination of $\{\varepsilon_i\}_{i=t}^T$. Since $d_{t-1}$, $e_{t-1}$, and $g_{t-1}$ are all known constants, the weights for each element in $\{y_i\}_{i=t+1}^T$ and $\{\varepsilon_i\}_{i=t}^T$ can be calculated.
	
	The calculation results show that the sum of the coefficients corresponding to $\{y_i\}_{i=t+1}^T$ is approximately equal to $\sqrt{\bar{\alpha}_t}$, and the square root of the sum of squared coefficients for $\{\varepsilon_i\}_{i=t}^T$ is approximately $\sqrt{1-\bar{\alpha}_t}$. Moreover, the approximation error decreases as the number of sampling steps increases (see Figures \ref{figure:ddpm_equiv_coeff}-\ref{figure:flow_euler_equiv_coeff} and Figures \ref{figure:sde_euler_equiv_coeff}-\ref{figure:ode_euler_equiv_coeff}). Therefore, these sampling methods can be represented in the form of the Natural Inference framework.

	The above computation can be quite complex, especially when the number of sampling steps is large. Therefore, it is necessary to seek more efficient computation methods. Symbolic computation software~\cite{sympy} offers a promising solution. With minor modifications to the original algorithm code, it can automatically compute the expression for each $x_t$. For more detailed information, please refer to the accompanying code.

	For higher-order sampling methods, their iteration rules are relatively complex, but the expression for $x_t$ can also be quickly calculated with the help of symbolic computation software. The calculation results indicate that DPMSOLVER, DPMSOLVER++, and DEIS yield results similar to those of first-order sampling methods (see Figures \ref{figure:deis_equiv_coeff}-\ref{figure:dpmsolverpp2s_equiv_coeff}). 
	
	\subsection{Advantages of the Natural Inference framework}
	
	Thus, we have used a completely new perspective to understanding the inference process. This new perspective has several advantages:
	\begin{itemize}
		\item The new perspective maintains \textbf{training-testing consistency}, where the goal during training is to predict $x_0$, and the goal during testing is also to predict $x_0$. 
		\item The new perspective divides the inference process into a series of operations for predicting $x_0$, each of which has clear input image signals and output image signals. This makes the inference process \textbf{more visual and interpretable}, providing significant help for debugging and problem analysis. Figures \ref{figure:sharp_process_0} and \ref{figure:sharp_process_1} provide a visualization of the complete inference process.
		\item As discussed in Section~\ref{section:new_target_view}, predicting $x_0$ can be regarded as an information enhancement operator. Similarly, Section~\ref{section:self_guide} shows that classifier-free guidance can also be viewed as an information enhancement operator. Therefore, the entire inference process can be understood as a progressive enhancement of information, a process that does not require any statistical knowledge.
		\item From this new perspective, existing sampling algorithms are merely specific parameter configurations within the Natural Inference framework. Within this framework, other more optimal parameter configurations can be found.
	\end{itemize}
	
	\section{Application of the Natural Inference framework} \label{section:application}
	
	\subsection{Reinterpreting why high-order sampling speedup sampling}
	
	In this subsection, we will explain, based on the Natural Inference framework, why algorithms like DEIS and DPMSolver perform better than DDPM and DDIM under low sampling steps.
	
	Tables \ref{table:ddpm_coeff_18} and \ref{table:ddim_coeff_18} show the coefficient matrices for DDPM and DDIM under the Natural Inference framework, while Tables \ref{table:deis_coeff_18} and \ref{table:dpmsolver3s_coeff_18} for DEIS and DPMSolver show their coefficient matrices. A careful comparison reveals a significant difference: 
	The coefficients in DEIS and DPMSolver contain negative values, which, according to the analysis in section \ref{section:self_guide}, can be considered as Fore Self Guidance operations, which can enhance image quality. In contrast, the coefficients in DDPM and DDIM are all positive, corresponding to Mid Self Guidance, which can not enhance image quality.
	Therefore, the key to the better performance of higher-order sampling algorithms lies in the design of Self Guidance.

	\subsection{Reinterpreting why low order sampling can adapt to big CFG}
	
	We know that, with a larger CFG value, most higher-order sampling algorithms, including 3rd-order DEIS and DPMSolver, exhibit a similar over-exposure problem, while DDIM and 2nd-order DPMSolver++ can alleviate this issue\cite{lu2022dpmpp}. So why do they work, while others do not? By comparing their coefficient matrices, we can identify the core of the problem. Tables \ref{table:ddim_coeff_18} and \ref{table:dpmsolverpp2s_coeff_18} show the coefficient matrices for DDIM and 2nd-order DPMSolver++, while Tables \ref{table:deis_coeff_18} and \ref{table:dpmsolver3s_coeff_18} show the coefficient matrices for 3rd-order DEIS and DPMSolver. A comparison reveals that the coefficients of 2nd-order DPMSolver++ and DDIM are all positive, representing a composition of Mid Self Guidances, while 3rd-order DEIS and DPMSolver contain negative values, indicating the presence of Fore Self Guidances. Therefore, it can be inferred that Mid Self Guidance (positive coefficients) is key to handling large CFGs, which can be further supported by the simple experiment in Appendix \ref{section:large_cfg}.
	
	Furthermore, we know that, compared to DDIM, 2nd-order DPMSolver++ not only adapts better to larger CFGs but also provides relatively better image quality. This can also be explained by the coefficient matrixs: compared to DDIM, the coefficient matrix (Table \ref{table:dpmsolverpp2s_coeff_18}) of 2nd-order DPMSolver++ has more zero elements, with more weight placed on the diagonal elements, i.e., the most recent outputs of the current step, leading to better image quality.
	
	\subsection{A way to control image sharpness}	

	As mentioned in section \ref{section:self_guide}, different Self Guidance operations affect the sharpness of the generated image. Therefore, by reasonably designing the weight distribution within the coefficient matrix, we can control the sharpness of the generated image. This principle is demonstrated with a SD3 example. Table \ref{table:euler_28} shows the coefficient matrix for the original Euler method, and Table \ref{table:adjusted_euler_28} shows the adjusted coefficient matrix. A key observation is that the Self Guidances in the Euler method primarily consist of composite Mid Self Guidance, with a significant amount of weight placed on earlier outputs. Conversely, the adjusted coefficient matrix eliminates weights for earlier outputs and assigns greater importance to more recent outputs, resulting in the generation of sharper images, as shown in Figure \ref{figure:adjusted coeff matrix}. 
	
	Note that although the adjusted coefficient matrix minimizes the weight of earlier outputs, each row still retains at least three non-zero elements. This is primarily to accommodate the influence of larger CFGs, especially for the earlier steps (larger $t$).
	
	\subsection{Better coefficient matrix} \label{section:better_matrix}
	
	In this subsection, we will take the pre-trained model on the CIFAR10 dataset\cite{krizhevsky2009learning} as an example to demonstrate that there exist better coefficient matrices that can achieve better FID scores under the same number of sampling steps. The pre-trained model uses the ScoreSDE \cite{song2020score} released cifar10\_ddpmpp\_continuous(vp).
	
	Table \ref{table:opt_coeff_5} shows an optimized coefficient matrix for 5 steps (NFE=5), Table \ref{table:opt_coeff_10} shows an optimized coefficient matrix for 10 steps, and Table \ref{table:opt_coeff_15} shows an optimized coefficient matrix for 15 steps. As shown in Table \ref{table:fid_result}, the optimized coefficient matrices achieve better FID scores than DEIS, DPMSolver, and DPMSolver++ when using the same number of steps. All algorithms use the same time discretization schedule(quadratic), and for DEIS, DPMSolver, and DPMSolver++, all hyperparameters are tested to select the best FID as the result.
	
	The optimized coefficient matrices are designed according to the following principles:
	\begin{itemize}[]
		\item The weight distribution tends to prioritize outputs closer to the current time step, reducing the weight of earlier outputs. In other words, \textit{reducing the tail length}. Generally, when the number of steps is larger, the \textit{tail} should be extended appropriately to avoid the \textbf{over-enhancement phenomenon}(see Appendix \ref{section:over_enhance}); when the number of steps is smaller, the \textit{tail} can be shortened.
		\item When the number of steps is small, Fore Self Guidance needs to be appropriately increased, which means adding more or larger negative coefficients before the diagonal elements.
		\item Conduct hyperparameter tuning to find more optimal coefficient matrices.
	\end{itemize}	
	\begin{table}[!ht]
		\centering
		\caption{Comparison of different algorithm (measured with FID)}
		\begin{tabular}{ccccc}
			\hline
			\textbf{step$\backslash$method} & \textbf{DEIS} & \textbf{DPMSolver} & \textbf{DPMSolver++} & \textbf{optimized matrix} \\ \hline
			\textbf{5 step} & 15.60 & 13.16 & 11.50 & \textbf{8.66} \\ 
			\textbf{10 step} & 3.94 & 5.06 & 4.64 & \textbf{3.58} \\ 
			\textbf{15 step} & 3.55 & 4.15 & 3.92 & \textbf{2.93} \\ 
			\hline
		\end{tabular}
		\label{table:fid_result}
	\end{table}

	\subsection{Limitations and directions for improvement}
	
	Although there are more optimal coefficient matrices than the existing sampling methods, the optimal coefficient matrix does not have a fixed form. When the model, number of steps, and time discretization strategy change, the coefficient matrix may also need to be adjusted accordingly. When the number of steps is large, the number of adjustable parameters become large, which poses a challenge for manual adjustments. One automated solution is to use hyperparameter optimization methods to automatically search for the best configuration \cite{yang2020hyperparameter}; another solution is to replace the Self Guidance operation (linear weighted sum) with a neural network, which can be optimized separately or jointly with the original model parameters. For the joint optimization approach, the model will take on an \textbf{autoregressive} form, whether during the training phase or the inference phase.
	
	\section{Conclusion}
	
	This paper investigates the operational principles of high-dimensional diffusion models. We first analyze the objective function and explore the impact of data sparsity in high-dimensional settings, demonstrating that, due to such sparsity, these models cannot effectively learn the underlying probability distributions or their key statistical quantities. Building on this insight, we propose a novel perspective for interpreting the objective function. In addition, we introduce a new inference framework that not only unifies most inference methods but also aligns with the degraded objective function. This framework offers an intuitive understanding of the inference process without relying on any statistical concepts. We hope that this work will encourage the community to rethink the operational principles of high-dimensional diffusion models and further enhance their training and inference methodologies.
	
	\FloatBarrier	
	
	\nocite{*}
	\bibliographystyle{plain}
	\bibliography{NaturalDiffusionEn.bib}
	
	\newpage

	\appendix
	
	\section{Additional proofs}
	
	\subsection{Predicting posterior mean is equivalent to predicting $X_0$}	\label{section:mean_eq_x0}
	
	In the following, we will prove that the following two objective functions are equivalent:
	\begin{align*}
		\min_\theta \int p(x_t) \left\|f_\theta(x_t) - \int p(x_0|x_t)\ x_0\ dx_0\right\|^2 dx_t \iff \min_\theta \iint p(x_0, x_t) \left\|f_\theta(x_t) - \ x_0\right\|^2 dx_0 dx_t 
	\end{align*}
	
	Proof:
	
	For $\left\|f_\theta(x_t) - \int p(x_0|x_t)x_0dx_0\right\|^2$, the following relation holds:
	\begin{align}
		&\left\|f_\theta(x_t) - \int p(x_0|x_t)x_0 dx_0\right\|^2 \\
		&= f^2_\theta(x_t) - 2f_\theta(x_t)\int p(x_0|x_t)x_0dx_0 +\left\|\int p(x_0|x_t)x_0dx_0\right\|^2 \\
		&= \int p(x_0|x_t) f^2_\theta(x_t) dx_0 - 2f_\theta(x_t)\int p(x_0|x_t)x_0dx_0 + C_1 \label{eq_full_prob} \\
		&= \int p(x_0|x_t)\left(f^2_\theta(x_t)-2f_\theta(x_t)x_0 + x^2_0\right) dx_0 - \int p(x_0|x_t) x^2_0 dx_0 + C_1 \\
		&= \int p(x_0|x_t)\left\|f^2_\theta(x_t) - x_0 \right\|^2 dx_0 - C_2 + C_1
	\end{align}
	Where $C_1$ and $C_2$ are constants that do not depend on $\theta$. In Equation \eqref{eq_full_prob}, we apply $f^2_\theta(x_t) = f^2_\theta(x_t) \int p(x_0|x_t) dx_0= \int p(x_0|x_t) f^2_\theta(x_t) dx_0$.
	
	Substituting the above relation into the objective function for predicting posterior mean, we get:
	\begin{align} \label{eq_pred_eps}
		&\int p(x_t) \left\|f_\theta(x_t) - \int p(x_0|x_t)x_0 dx_0\right\|^2 dx_t \\
		&= \int p(x_t) \left(\int p(x_0|x_t)\left\|f^2_\theta(x_t) - x_0 \right\|^2 dx_0 - C_2 + C_1\right) dx_t \\
		&= \iint p(x_0, x_t) \left\|f_\theta(x_t) - \ x_0\right\|^2 dx_0 dx_t + \int p(x_t)\left(C_1 - C_2\right)dx_t \\
		&= \iint p(x_0, x_t) \left\|f_\theta(x_t) - \ x_0\right\|^2 dx_0 dx_t + C_3
	\end{align}
	
	That is, the two objective functions differ only by a constant that does not depend on the optimization parameters. Therefore, the two objective functions are equivalent.

	\subsection{Conditional score} \label{section:conditional_score}
	
	Below is the proof of the following relation:
	\begin{alignat}{2}
		\frac{\partial \log p(x_t)}{\partial x_t} = \int p(x_0|x_t)\ \frac{\partial \log p(x_t|x_0)}{\partial x_t} dx_0
	\end{alignat}
	
	Proof:
	
	\begin{alignat}{2}
		\frac{\partial \log p(x_t)}{\partial x_t} &= \frac{1}{p(x_t)}\frac{\partial p(x_t)}{\partial x_t} \\
		 &= \frac{1}{p(x_t)}\frac{\partial \left(\int p(x_0)p(x_t|x_0)dx_0\right)}{\partial x_t} \\
		&= \int \frac{p(x_0)}{p(x_t)}\frac{\partial p(x_t|x_0)}{\partial x_t} dx_0 \\
		&= \int \frac{p(x_0,x_t)/p(x_t)}{p(x_0,x_t)/p(x_0)}\ \frac{\partial p(x_t|x_0)}{\partial x_t} dx_0 \\
		&= \int \frac{p(x_0|x_t)}{p(x_t|x_0)}\ \frac{\partial p(x_t|x_0)}{\partial x_t} dx_0 \\
		 &= \int p(x_0|x_t)\ \frac{\partial \log p(x_t|x_0)}{\partial x_t} dx_0
	\end{alignat}
	
	\subsection{Form of the posterior probability} \label{section:posterior_form}
	
	The following derivation is based on \cite{zheng2023art} and \cite{zheng2024interactive}.
	
	Assume that $x_t$ has the following form:
	\begin{align}
		x_t = c_0 \cdot x_0 + c_1 \cdot \epsilon \qquad	\text{where $c_0$ and $c_1$ is constant}
	\end{align}
	
	Then we have:
	
	\begin{align}	\label{eq_fore_norm}
		p(x_t|x_0) \sim \mathcal{N}(x_t; c_0x_0, c^2_1)
	\end{align}
	
	According to Bayes' theorem, we have
	\begin{align}
		p(x_0|x_t) &= \frac{p(x_t|x_0)p(x_0)}{p(x_t)} 	\\
		&= \frac{p(x_t|x_0)p(x_0)}{\int p(x_t|x_0)p(x_0) dx_0} \\
		&= \text{Normalize}\big(p(x_t|x_0)p(x_0)\big)
	\end{align}
	
	where \textit{Normalize} represents the normalization operator, and the normalization divisor is $\int p(x_t|x_0)p(x_0) dx_0$.

	Substituting Equation \eqref{eq_fore_norm} into this, we get:
	
	\begin{align}
		p(x_0|x_t) &= \text{Normalize}\left(\frac{1}{\sqrt{2\pi c^2_1}}\exp{\frac{-(x_t-c_0x_0)^2}{2c^2_1}}\ p(x_0)\right) \\
		&= \text{Normalize}\left(\frac{1}{\sqrt{2\pi c^2_1}}\exp{\frac{-(x_0 - \frac{x_t}{c_0})^2}{2\frac{c^2_1}{c^2_0}}}\ p(x_0)\right) \\
		&= \text{Normalize}\left(\exp{\frac{-(x_0 - \mu)^2}{2\sigma^2}}\ p(x_0)\right) \\
		&\qquad \text{where}\ \mu= \frac{x_t}{c_0} \qquad \sigma = \frac{c_1}{c_0}
	\end{align}
	
	In the above derivation, due to the presence of the normalization operator, we can ignore the factor $\frac{1}{\sqrt{2\pi c^2_1}}$.

	\FloatBarrier

	\section{Self Guidance and its composition} \label{section:self_guide_appendix}
	
	In this section, we show that the linear combination of any two model outputs(prediting $x_0$) can be viewed as a Self Guidance operation.
	
	As described in Section 3.2, the self guidance is defined as follows:
	\begin{align}
		I_{out} = I_{bad} + \lambda \cdot (I_{good} - I_{bad})
	\end{align}
	This equation can be further written as
	\begin{align} \label{eq_guide_second}
		I_{out} &= \lambda \cdot I_{good} + (1-\lambda) \cdot I_{bad} \\ &= \mlg{\eta}_{good} \cdot I_{good} + \mlg{\eta}_{bad} \cdot I_{bad}\\ \text{where}\quad &\mlg{\eta}_{good}, \mlg{\eta}_{bad} \in real \quad \mlg{\eta}_{good} + \mlg{\eta}_{bad} = 1 
	\end{align}
	As shown above, the coefficients of $I_{bad}$ and $I_{good}$ can take any value, but the sum of $I_{bad}$ and $I_{good}$ must equal 1. For \textbf{Fore Self Guidence}, $\mlg{\eta}_{good} > 0$, $\mlg{\eta}_{bad} < 0$; for \textbf{Mid Self Guidence}, $\mlg{\eta}_{good} > 0$, $\mlg{\eta}_{bad} > 0$; for \textbf{Back Self Guidence}, $\mlg{\eta}_{good} < 0$, $\mlg{\eta}_{bad} > 0$.
	
	For the linear combination of any two model outputs, it can be written as:
	\begin{align} \label{eq_guide_third}
		I_{out} = a \cdot I_{good} + b \cdot I_{bad} = (a+b)\cdot(\frac{a}{a+b} \cdot I_{good} + \frac{b}{a+b} \cdot I_{bad})
	\end{align}
	 Since the sum of the two coefficients equals 1, the operation inside the parentheses is a Self Guidence operation.
	 
	 Thus, \textbf{the linear combination of any two $I_{bad}$ and $I_{good}$ can be represented as Self Guidence with a scaling factor}. 

	 For the linear combination of multiple model outputs, it can be written as:
	 \begin{align} \label{eq_guide_four}
		 I_{out} &= a \cdot I_a + b \cdot I_b + c\cdot I_c \\
		         &= (a+b)\cdot(\frac{a}{a+b} \cdot I_a + \frac{b}{a+b} \cdot I_b) + c\cdot I_c \\
				 &= (a+b+c)\cdot \big (\frac{a+b}{a+b+c}\cdot(\frac{a}{a+b}I_a + \frac{b}{a+b}I_b) + \frac{c}{a+b+c}\cdot I_c\big )
	 \end{align}

	 Thus, \textbf{the linear combination of multiple model outputs can be viewed as a composition of Self Guidences}.

	\FloatBarrier

	\section{Represent sampling methods with Natural Inference framework} \label{section:represent}

	\subsection{Represent DDPM Ancestral Sampling with Natural Inference framework}
	
	This subsection will demonstrate that the DDPM Ancestor Sampling can be reformulated within the Natural Inference framework. The iterative process of the Ancestor Sampling is as follows:
	
	\begin{equation} 
		\begin{aligned} \label{eq_ancetral_sampling}
			\mlg{y}_t &= \mlg{f}_t(\mlg{x}_t)	\\
			\mlg{x}_{t-1} &= \mlg{d}_{t-1}\cdot \mlg{x}_{t} + \mlg{e}_{t-1}\cdot \mlg{y}_{t} + \mlg{g}_{t-1}\cdot \mlg{\eps_{t-1}}	\\
			\text{where}\ d_{t-1} = \frac{\sqrt{\alpha_t}(1-\bah_{t-1})}{1-\bah_t}& \quad e_{t-1} = \frac{\sqrt{\bah_{t-1}}\beta_t}{1-\bah_t} \quad g_{t-1} = \sqrt{\frac{1-\bah_{t-1}}{1-\bah_t}\beta_t}
		\end{aligned} 
	\end{equation}
	
	Here, $f_t$ is the model function at the $t$-th step. In this case, we assume the model predicts $x_0$, but other forms of prediction models (such as predict $\varepsilon$ or predict $\mlg{v}$) can be transformed into the form of predicting $x_0$. $\mlg{y}_t$ is the output of $f_t$, which is the predicted $x_0$ at the $t$-th step.
	
	According to the above iterative algorithm, $x_{T-1}$ can be expressed as
	\begin{equation} 
	\begin{aligned} \label{eq_ancetral_sampling_2}
			\mlg{x}_T &= \mlg{g}_{T}\cdot \mlgg{\eps_{T}}	 \quad \text{where}\ \mlg{g}_{T} = 1\\
			\mlg{x}_{T-1} &= \mlg{d}_{T-1}\cdot \mlg{x}_{T} + \mlg{e}_{T-1}\cdot \mlgr{y_{T}} + \mlg{g}_{T-1}\cdot \mlgg{\eps_{T-1}} \\
			&=\mlg{e}_{T-1}\mlgr{y_{T}} + (\mlg{d}_{T-1}\mlg{g}_{T}\mlgg{\eps_{T}} + \mlg{g}_{T-1}\mlgg{\eps_{T-1}}) 
	\end{aligned} 
	\end{equation}
	
	Based on the expression of $x_{T-1}$, the expression of $x_{T-2}$ can be written as
	\begin{equation} 
	\begin{aligned} \label{eq_ancetral_sampling_3}
		\mlg{x}_{T-2} &= \mlg{d}_{T-2}\cdot \mlg{x}_{T-1} + \mlg{e}_{T-2}\cdot \mlgr{y_{T-1}} + \mlg{g}_{T-2}\cdot \mlgg{\eps_{T-2}} \\
		&= (\mlg{d}_{T-2}\mlg{e}_{T-1}\mlgr{y_{T}} + \mlg{e}_{T-2}\mlgr{y_{T-1}}) + (\mlg{d}_{T-2}\mlg{d}_{T-1}\mlg{g}_{T}\mlgg{\eps_{T}} + \mlg{d}_{T-2}\mlg{g}_{T-1}\mlgg{\eps_{T-1}} + \mlg{g}_{T-2}\mlgg{\eps_{T-2}})
	\end{aligned} 
	\end{equation}
	
	Based on the expression of $\mlg{x}_{T-2}$, the expression of $\mlg{x}_{T-3}$ can be further written as
	\begin{equation} 
		\begin{aligned} \label{eq_ancetral_sampling_4}
			\mlg{x}_{T-3} &= \mlg{d}_{T-3}\cdot \mlg{x}_{T-2} + \mlg{e}_{T-3}\cdot \mlgr{y_{T-2}} + \mlg{g}_{T-3}\cdot \mlgg{\eps_{T-3}} \\
			&= (\mlg{d}_{T-3}\mlg{d}_{T-2}\mlg{e}_{T-1}\mlgr{y_{T}} + \mlg{d}_{T-3}\mlg{e}_{T-2}\mlgr{y_{T-1}} + \mlg{e}_{T-3}\mlgr{y_{T-2}}) \\
			 &\ + (\mlg{d}_{T-3}\mlg{d}_{T-2}\mlg{d}_{T-1}\mlg{g}_{T}\mlgg{\eps_{T}} + \mlg{d}_{T-3}\mlg{d}_{T-2}\mlg{g}_{T-1}\mlgg{\eps_{T-1}} + \mlg{d}_{T-3}\mlg{g}_{T-2}\mlgg{\eps_{T-2}} + \mlg{g}_{T-3} \mlgg{\eps_{T-3}})
		\end{aligned} 
	\end{equation}
	
	Similarly, each $x_t$ can be recursively written in a similar form. It can be observed that each $x_t$ can be decomposed into two parts: one part is a weighted sum of past predictions of $x_0$ (i.e., $y_t$), and the other part is a weighted sum of past noise and newly added noise. Since $d_t$, $e_t$, and $g_t$ are all known constants, the equivalent signal coefficient and equivalent noise coefficient for each $x_t$ can be accurately computed.
	
\begin{figure}
		\centering
		\begin{subfigure}[]{\includegraphics[width=0.32\textwidth]{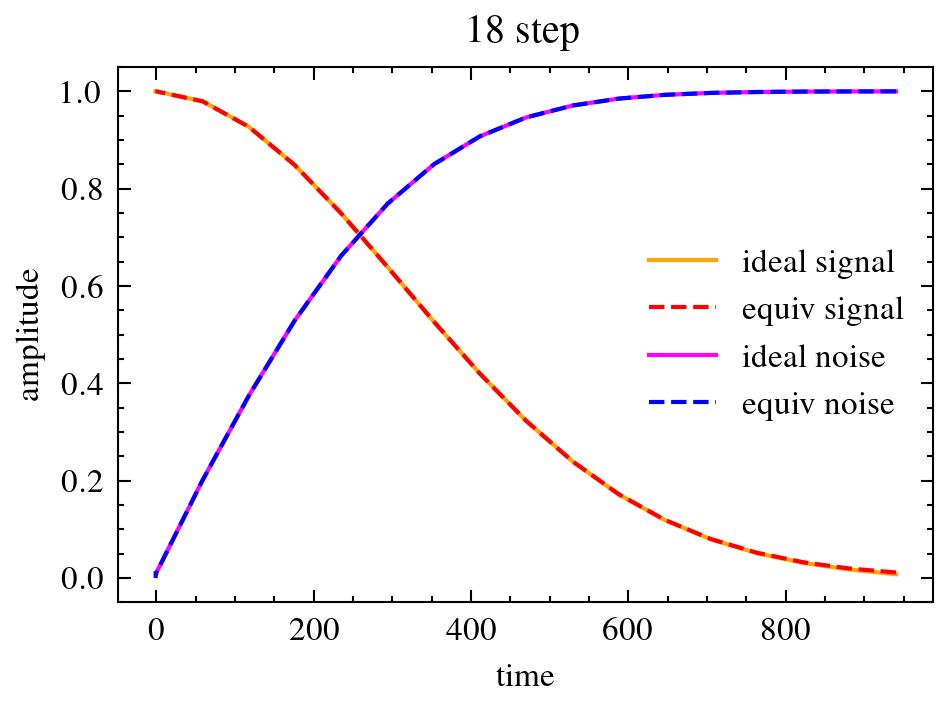}} \end{subfigure}
		\begin{subfigure}[]{\includegraphics[width=0.32\textwidth]{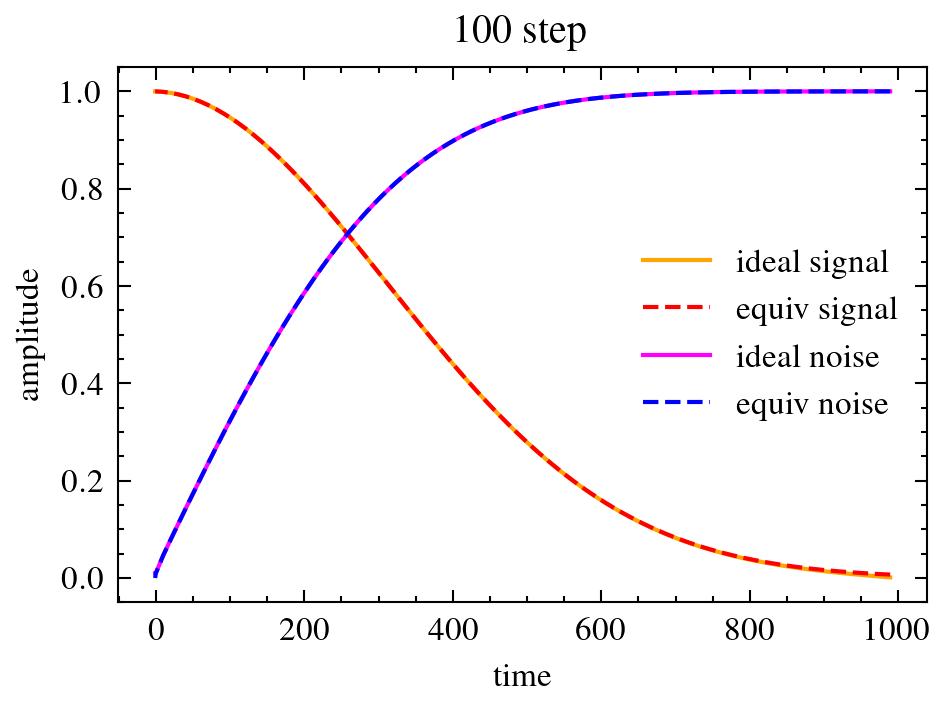}} \end{subfigure}
		\begin{subfigure}[]{\includegraphics[width=0.32\textwidth]{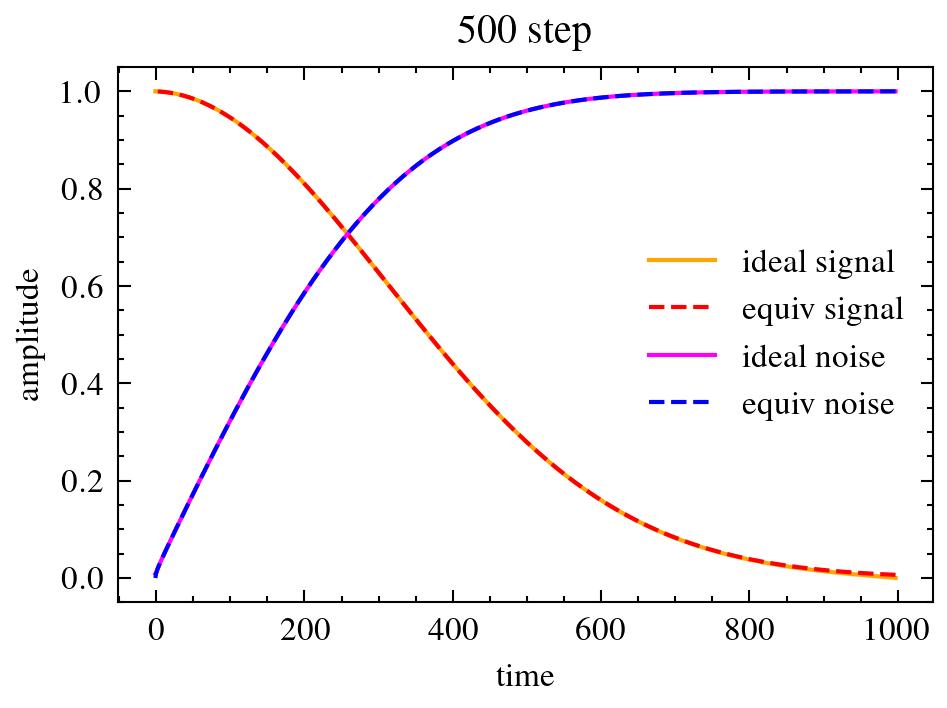}} \end{subfigure}
		\caption{DDPM equivalent marginal coefficients and ideal margingal coefficients \quad (a) 18 step \quad  (b) 100 step \quad (c) 500 step} 
		\label{figure:ddpm_equiv_coeff}
	\end{figure}

	The computation results show that the equivalent signal coefficient of each $x_t$ is almost equal to $\sqrt{\bah_t}$, and the equivalent noise coefficient is approximately $\sqrt{1-\bah_t}$. Moreover, the slight error diminishes as the number of sampling steps $T$ increases. Specifically, Figure \ref{figure:ddpm_equiv_coeff} illustrates the results for 18 steps, 100 steps, and 500 steps.
	
	Table \ref{table:ddpm_coeff_18} presents the complete coefficients of each $x_t$ with respect to $y_t$ in matrix form, where each row corresponds to an $x_t$. Table \ref{table:ddpm_coeff_eps_18} provides the complete coefficients of each $x_t$ with respect to $\mlg{\eps}_t$ in matrix form. It can be seen that the noise coefficient matrix differs slightly from the signal coefficient matrix, with an additional nonzero coefficient appearing to the right of the diagonal elements. This indicates that a small amount of new noise is introduced at each step, causing the overall noise \textit{pattern} to change at a slow rate.
	
	At this point, we have successfully demonstrated that the DDPM Ancestral Sampling process can be represented using the Natural Inference framework.
	
	\subsection{Represent DDIM with Natural Inference framework}
	
	The iterative rule of the DDIM can be expressed in the following form:
	\begin{equation} 
		\begin{aligned} \label{eq_ddim_sampling}
			\mlg{y}_t &= \mlg{f}_t(\mlg{x}_t)	\\
			\mlg{x}_{t-1} &= \sqrt{\bah_{t-1}}\cdot \mlg{x}_t + \sqrt{1-\bah_{t-1}}\cdot \frac{x_t - \sqrt{\bah_t}\mlg{y}_t}{\sqrt{1-\bah_t}} \cdot \mlg{y}_{t}\\
			 &= \mlg{d}_{t-1}\cdot \mlg{x}_{t} + \mlg{e}_{t-1}\cdot \mlg{y}_{t}  \\
			 \text{where}\ \mlg{d}_{t-1} &= \frac{\sqrt{1-\bah_{t-1}}}{\sqrt{1-\bah_{t}}} \quad \mlg{e}_{t-1} = (\sqrt{\bah_{t-1}}-\frac{\sqrt{1-\bah_{t-1}}}{\sqrt{1-\bah_{t}}\sqrt{\bah_t}})
		\end{aligned} 
	\end{equation}
	
	It can be seen that the iterative rule of DDIM is similar to those of DDPM Ancestral Sampling, except that the term $\mlg{g}_{t-1}\cdot \mlg{\eps_{t-1}}$ is missing, meaning that no new noise is added at each step. Following the recursive way of DDPM Ancestral Sampling, each $x_t$ corresponding to $t$ can also be written in a similar form, as follows:
	
	\begin{equation} 
	\begin{aligned} \label{eq_ddim_sampling_0}
		\mlg{x}_T &= \mlg{g}_{T}\cdot \mlgg{\eps_{T}}	 \qquad \text{where}\ \mlg{g}_{T} = 1 \\
		\mlg{x}_{T-1} &=\mlg{e}_{T-1}\mlgr{y_{T}} + \mlg{d}_{T-1}\mlg{g}_{T}\mlgg{\eps_{T}} \\
		\mlg{x}_{T-2} &= (\mlg{d}_{T-2}\mlg{e}_{T-1}\mlgr{y_{T}} + \mlg{e}_{T-2}\mlgr{y_{T-1}}) + \mlg{d}_{T-2}\mlg{d}_{T-1}\mlg{g}_{T}\mlgg{\eps_{T}} \\
		\mlg{x}_{T-3} &= (\mlg{d}_{T-3}\mlg{d}_{T-2}\mlg{e}_{T-1}\mlgr{y_{T}} + \mlg{d}_{T-3}\mlg{e}_{T-2}\mlgr{y_{T-1}} + \mlg{e}_{T-3}\mlgr{y_{T-2}}) \\
		&\ + \mlg{d}_{T-3}\mlg{d}_{T-2}\mlg{d}_{T-1}\mlg{g}_{T}\mlgg{\eps_{T}}
	\end{aligned} 
	\end{equation}
	
	It can be seen that the form of DDIM is slightly different from DDPM. Since DDIM does not introduce new noise at each step, there is only one noise term.
	
\begin{figure}
		\centering
		\begin{subfigure}[]{\includegraphics[width=0.32\textwidth]{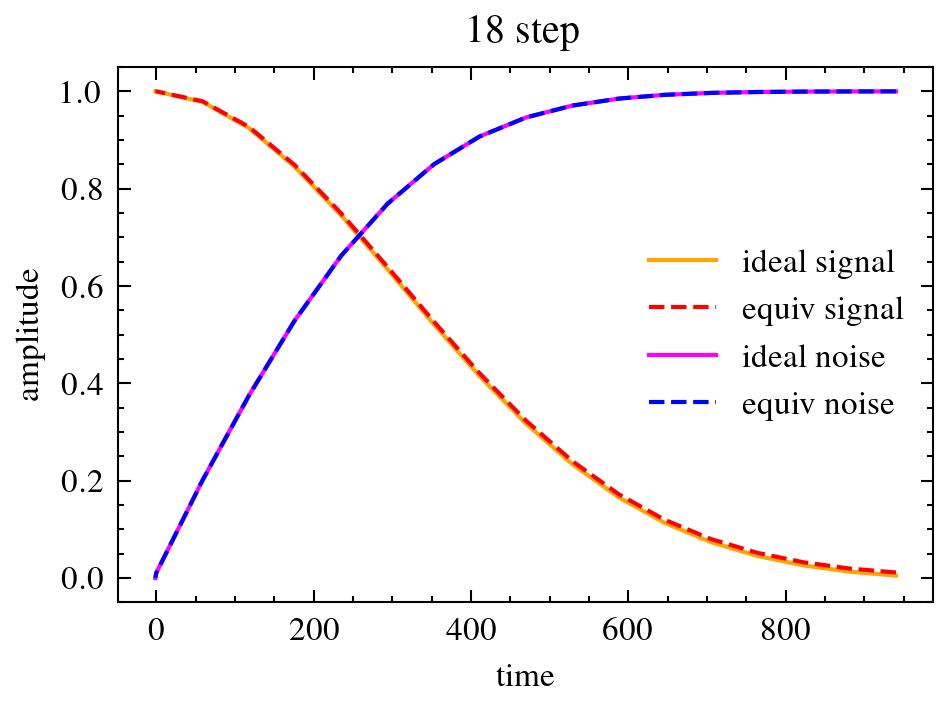}} \end{subfigure}
		\begin{subfigure}[]{\includegraphics[width=0.32\textwidth]{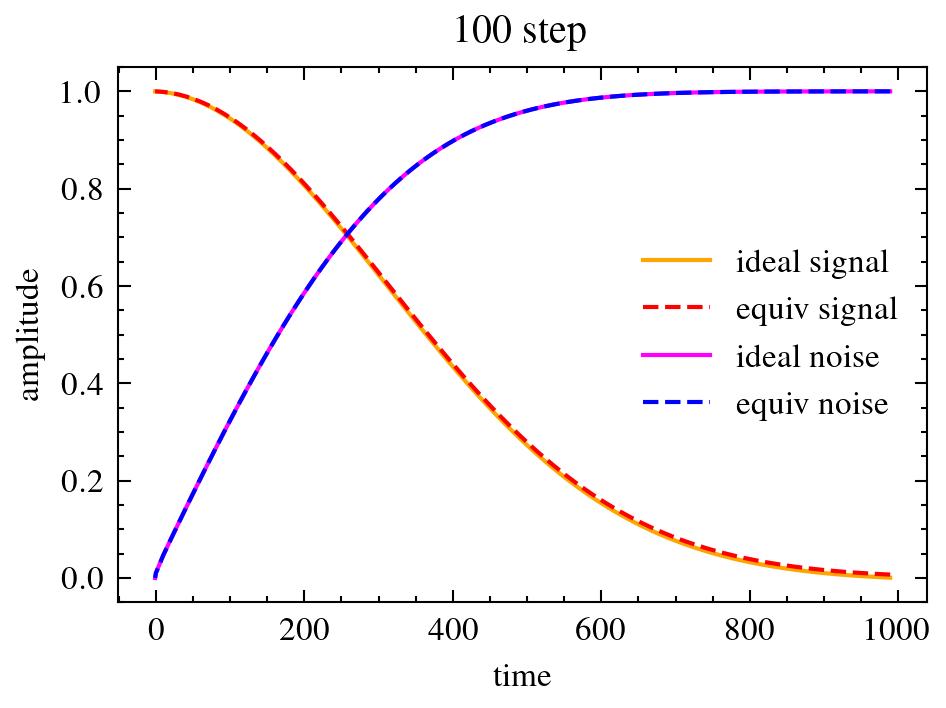}} \end{subfigure}
		\begin{subfigure}[]{\includegraphics[width=0.32\textwidth]{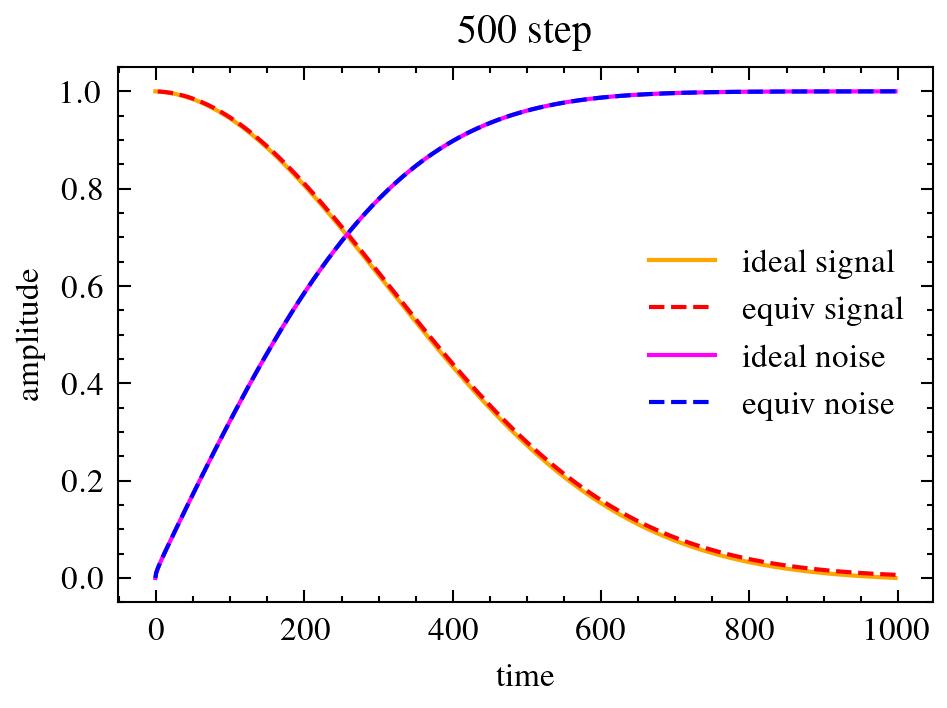}} \end{subfigure}
		\caption{DDIM equivalent marginal coefficients and ideal marginal coefficients \quad (a) 18 step \quad (b) 100 step \quad (c) 500 step} 
		\label{figure:ddim_equiv_coeff}
	\end{figure}

	The computation results show that the equivalent signal coefficients of each $x_t$ are approximately equal to $\sqrt{\bah_t}$, and the equivalent noise coefficient contains only the term related to $\eps_T$, whose coefficient is almost equal to $\sqrt{1-\bah_t}$. Figure \ref{figure:ddim_equiv_coeff} illustrates the results for 18 steps, 100 steps, and 500 steps, respectively. It can be observed that the errors in the equivalent coefficients are minimal and almost indistinguishable. Table \ref{table:ddim_coeff_18} presents the complete signal coefficient matrix for 18 steps.
	
	Therefore, the sampling process of DDIM can also be represented using the Natural Inference framework.
	
	\subsection{Represent Flow Matching Euler Sampling with Natural Inference Framework}
	
	The noise mixing method of Flow Matching is shown in Equation \eqref{eq_flow_merge}. When using Euler discretized integral sampling, its iterative rule can be expressed as follows:
	
	\begin{equation} 
	\begin{aligned} \label{eq_fm_0}
		\mlg{y}_i &= \mlg{f}_i(\mlg{x}_i) \\
		\mlg{x}_{i-1} &= \mlg{x}_{i} + (\mlg{t}_{i-1}-\mlg{t}_i)(-\mlg{y}_i + \epsilon)	\\
			  &= \mlg{x}_{i} + (\mlg{t}_{i-1}-\mlg{t}_i)\frac{\mlg{x}_i - \mlg{y}_i}{t_i} \\
			  &= \mlg{d}_{i-1}\cdot \mlg{x}_{i} + \mlg{e}_{i-1}\cdot \mlg{y_i} \\
			  \text{where}\ &\mlg{d}_{i-1} = \frac{\mlg{t}_{i-1}}{\mlg{t}_{t_i}} \quad \mlg{e}_{i-1} = (1-\frac{\mlg{t}_{i-1}}{\mlg{t}_{t_i}})
	\end{aligned} 
	\end{equation}
	
	where $\mlg{f}_i$ is the model predicting $x_0$, and $\mlg{y}_i$ is the output of the model $\mlg{f}_i$ corresponding to the discrete time point $\mlg{t}_i$.
	
	It can be observed that the iterative rule of the Euler algorithm in Flow Matching is similar to that of DDIM, so each $x_i$ can also be expressed in a similar form.
	
\begin{figure}
		\centering
		\begin{subfigure}[]{\includegraphics[width=0.32\textwidth]{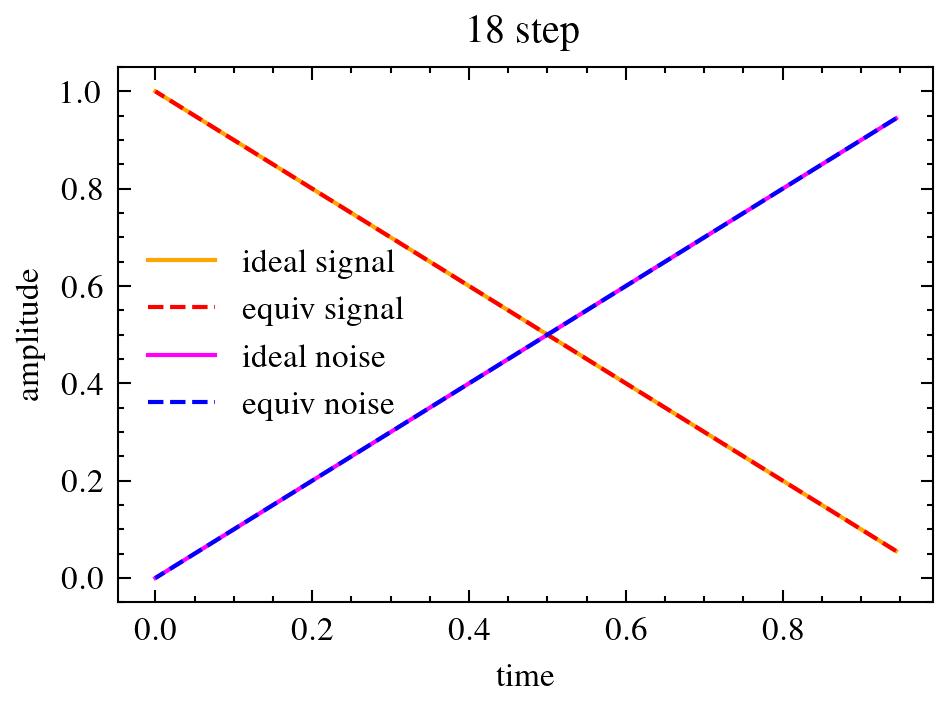}} \end{subfigure}
		\begin{subfigure}[]{\includegraphics[width=0.32\textwidth]{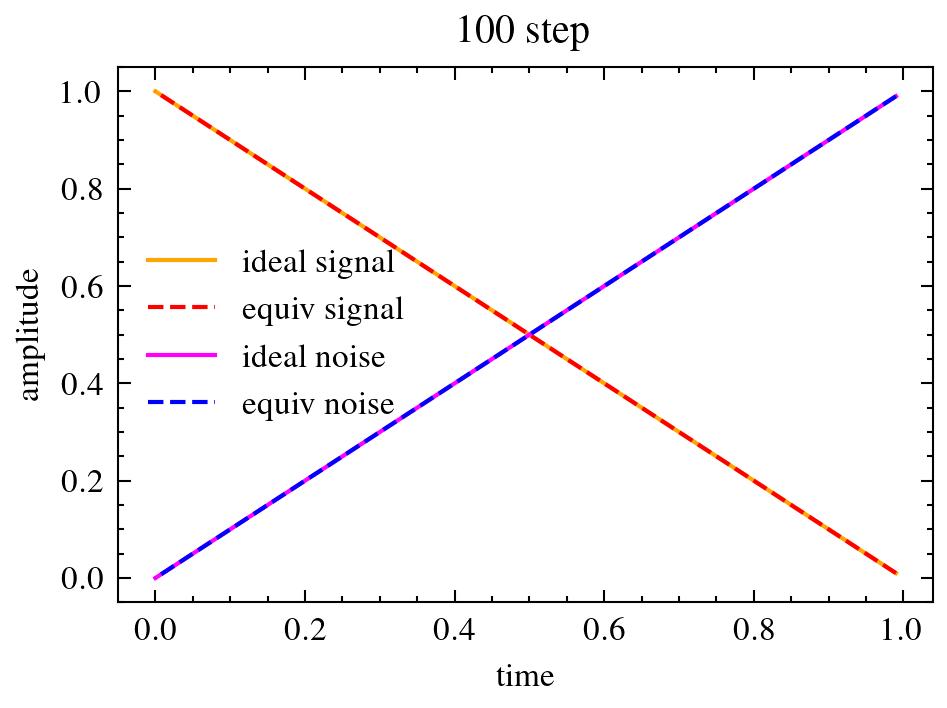}} \end{subfigure}
		\begin{subfigure}[]{\includegraphics[width=0.32\textwidth]{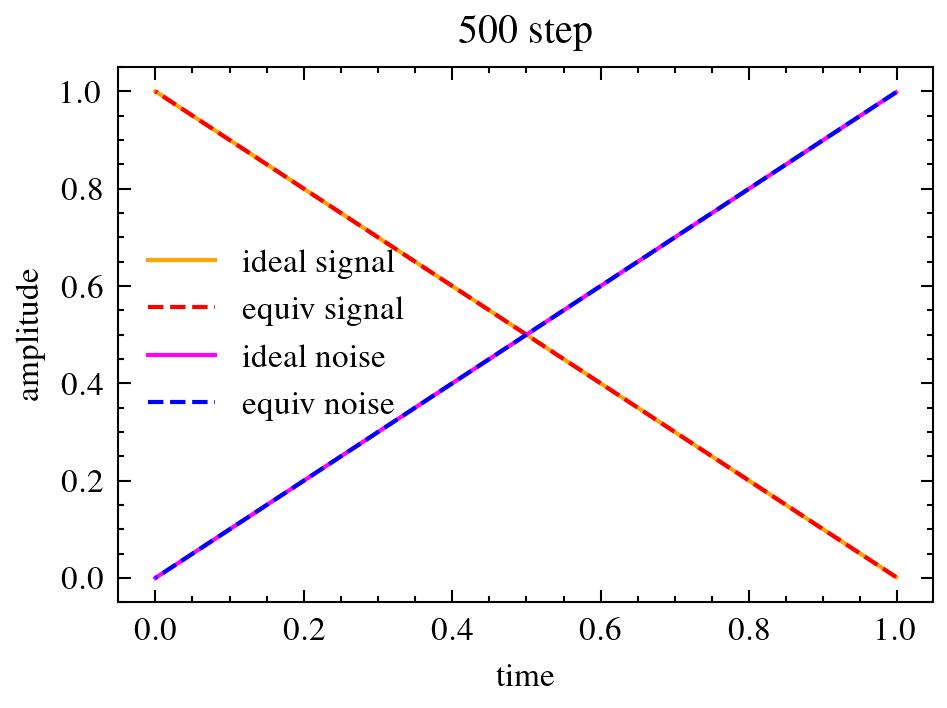}} \end{subfigure}
		\caption{Flow matching euler sampler equivalent marginal coefficients and ideal marginal coefficients \quad  (a) 18 step  \quad (b) 100 step \quad (c) 500 step} 
		\label{figure:flow_euler_equiv_coeff}
	\end{figure}

	The computation results show that for each discrete point $\mlg{t}_i$, the equivalent signal coefficient of $x_i$ is \textbf{exactly equal to} $1-\mlg{t}_i$, and the equivalent noise coefficient has only the $\eps_N$ term, whose coefficient is \textbf{exactly equal to} $\mlg{t}_i$. The specific results can be seen in Figure \ref{figure:flow_euler_equiv_coeff}, which shows the results for 18 steps, 200 steps, and 500 steps, respectively. Table \ref{table:flow_euler_coeff_18} presents the signal coefficient matrix for 18 steps.
	
	\subsection{Represent high order samplers with Natural Inference framework}
	
	In the previous sections, most first-order sampling algorithms have already been represented using the Natural Inference framework. For second-order and higher-order sampling algorithms, since the update rules of $x_i$ are more complex, it is quite difficult to directly compute the expression of each $x_i$. Therefore, alternative solutions must be sought. Symbolic computation tools provide a suitable solution to this challenge, as they can automatically analyze complex mathematical expressions. With slight modifications to the original algorithm code, they can automatically compute the coefficients of each $y_i$ term and $\eps_i$ term. The toolkit used in this paper is SymPy\cite{sympy}, and For specific details, please refer to the code attached in this paper.
	
	The compuation results show that DEIS, DPMSolver, and DPMSolver++ yield the same conclusion as DDIM: each $x_i$ can be decomposed into two parts, with its equivalent signal coefficient approximately equal to $\sqrt{\bah_i}$ and its equivalent noise coefficient approximately equal to $\sqrt{1-\bah_i}$.
	
\begin{figure}
		\centering
		\begin{subfigure}[]{\includegraphics[width=0.32\textwidth]{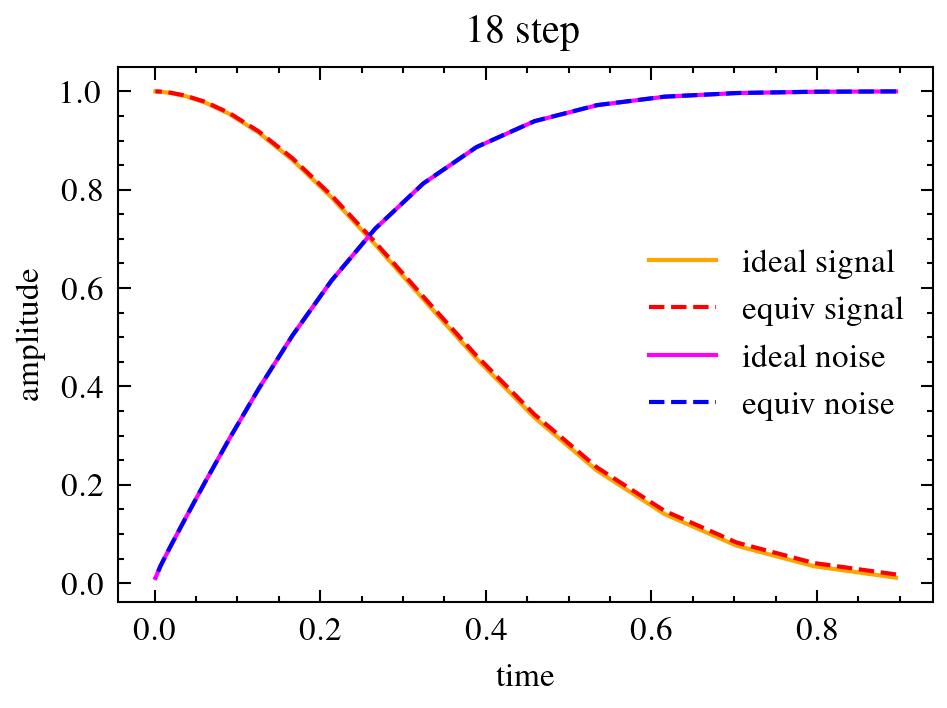}} \end{subfigure}
		\begin{subfigure}[]{\includegraphics[width=0.32\textwidth]{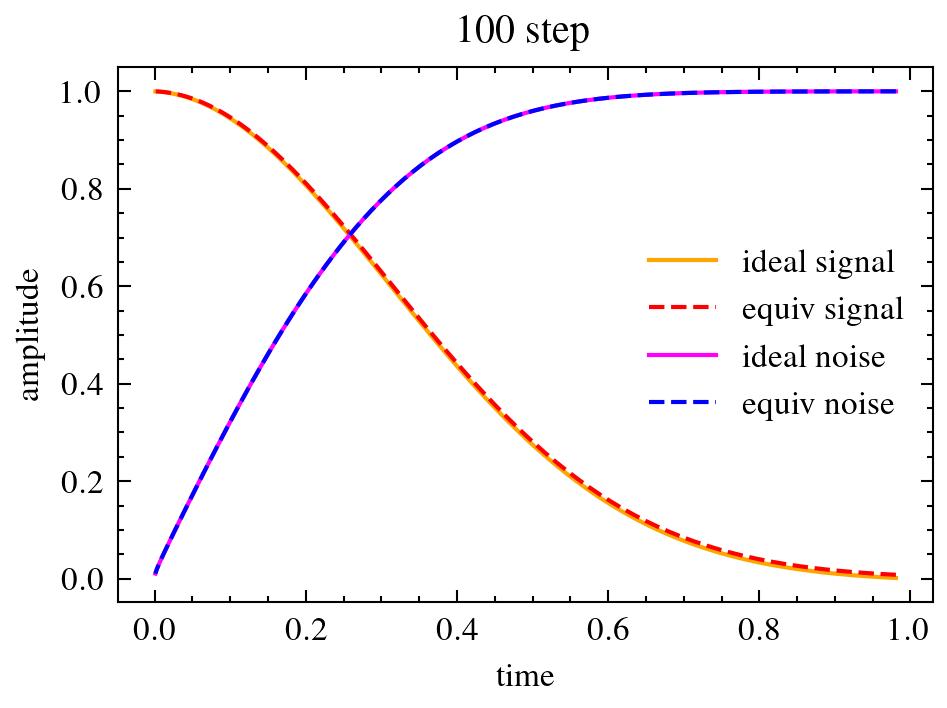}} \end{subfigure}
		\begin{subfigure}[]{\includegraphics[width=0.32\textwidth]{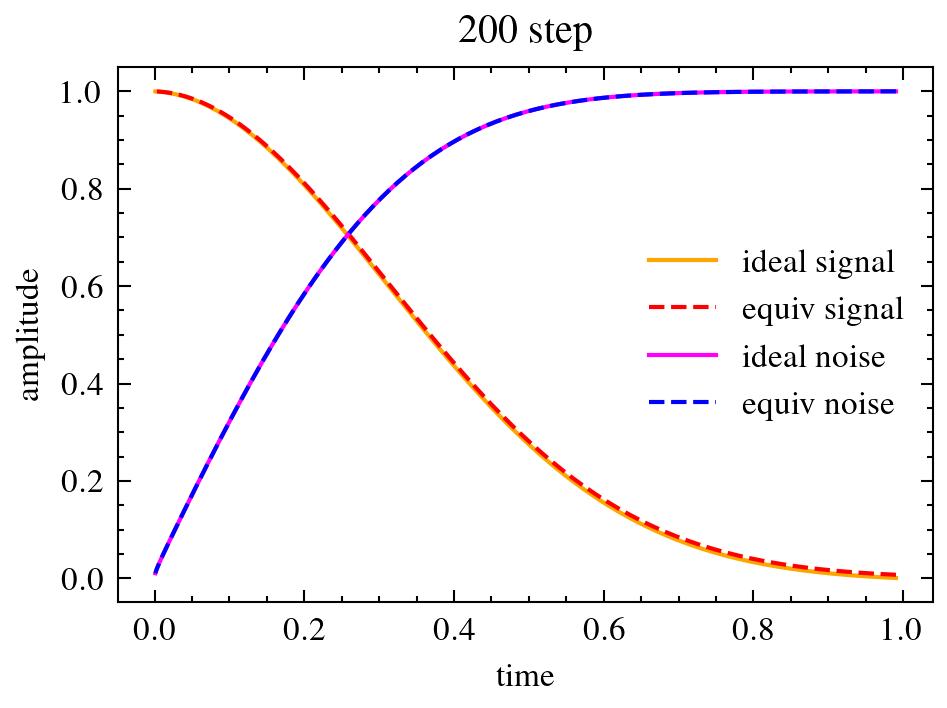}} \end{subfigure}
		\caption{DEIS equivalent marginal coefficients and ideal marginal coefficients \quad (a) 18 step \quad (b) 100 step \quad (c) 500 step} 
		\label{figure:deis_equiv_coeff}
	\end{figure}

\begin{figure}
		\centering
		\begin{subfigure}[]{\includegraphics[width=0.32\textwidth]{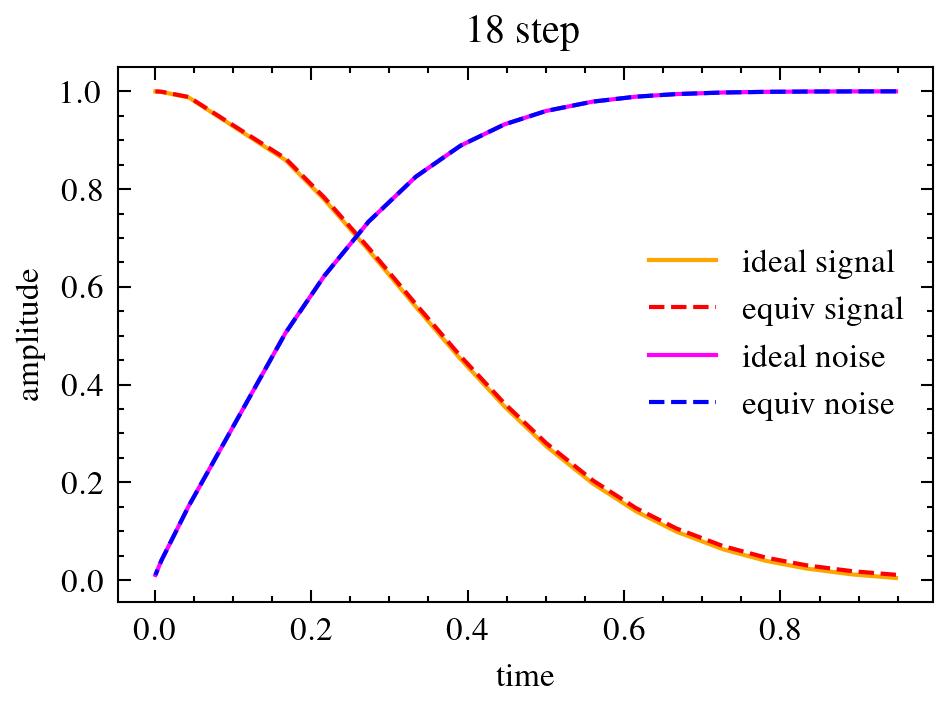}} \end{subfigure}
		\begin{subfigure}[]{\includegraphics[width=0.32\textwidth]{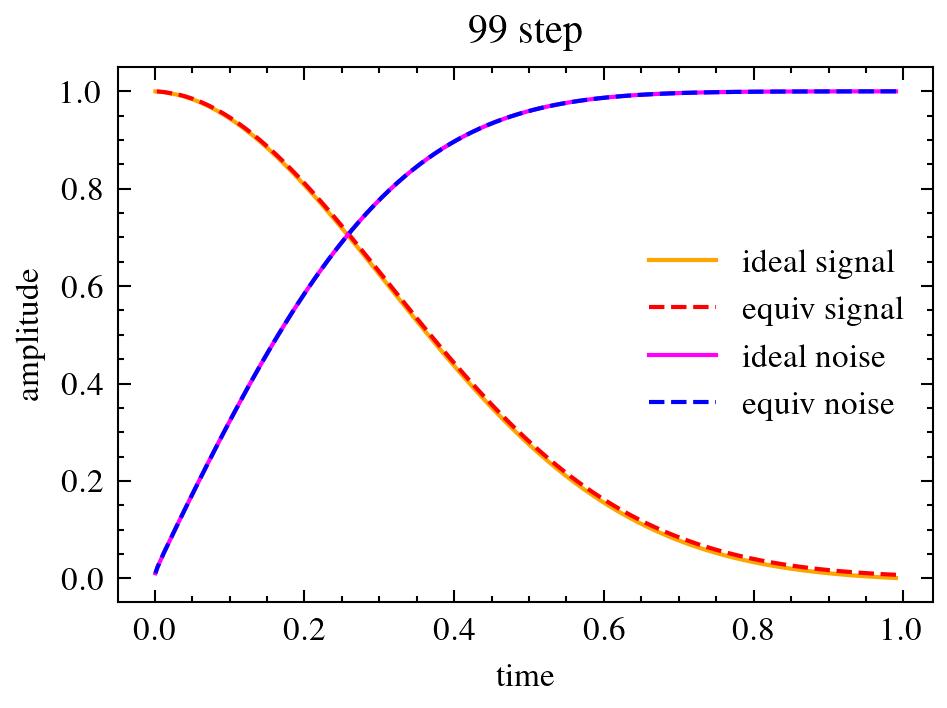}} \end{subfigure}
		\begin{subfigure}[]{\includegraphics[width=0.32\textwidth]{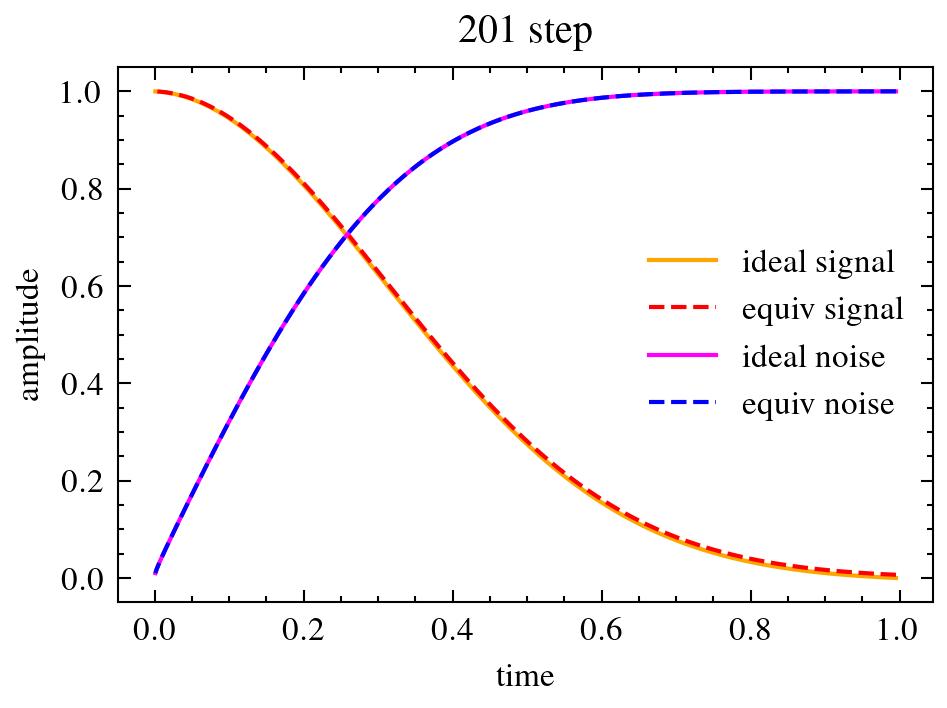}} \end{subfigure}
		\caption{dpmsolver3s equivalent marginal coefficients and ideal marginal coefficients \quad (a) 18 step \quad (b) 99 step \quad (c) 201 step} 
		\label{figure:dpmsolver3s_equiv_coeff}
	\end{figure}

\begin{figure}
		\centering
		\begin{subfigure}[]{\includegraphics[width=0.32\textwidth]{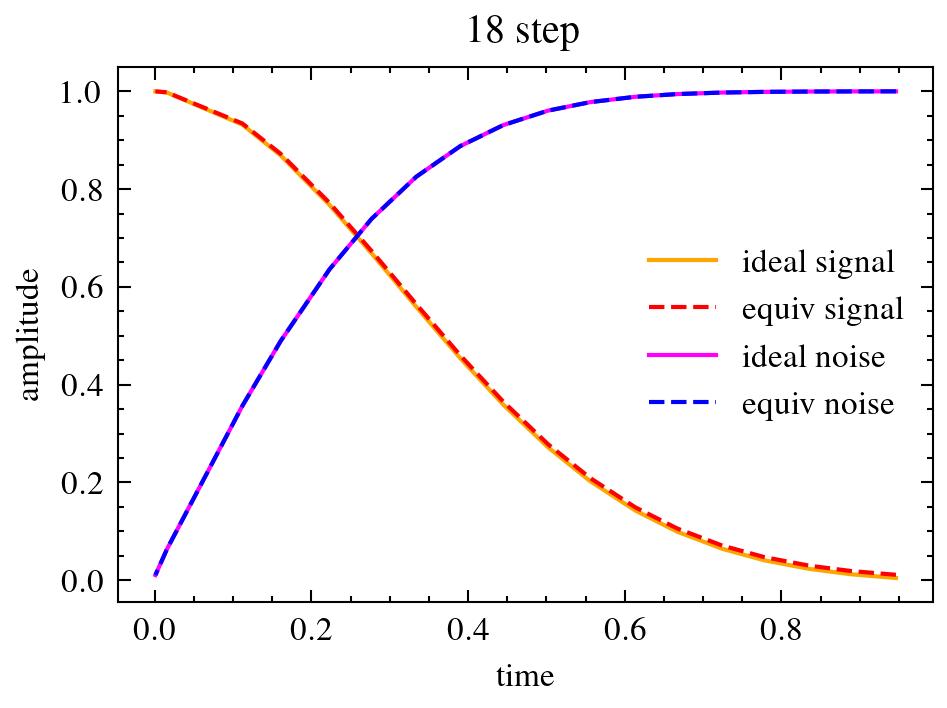}} \end{subfigure}
		\begin{subfigure}[]{\includegraphics[width=0.32\textwidth]{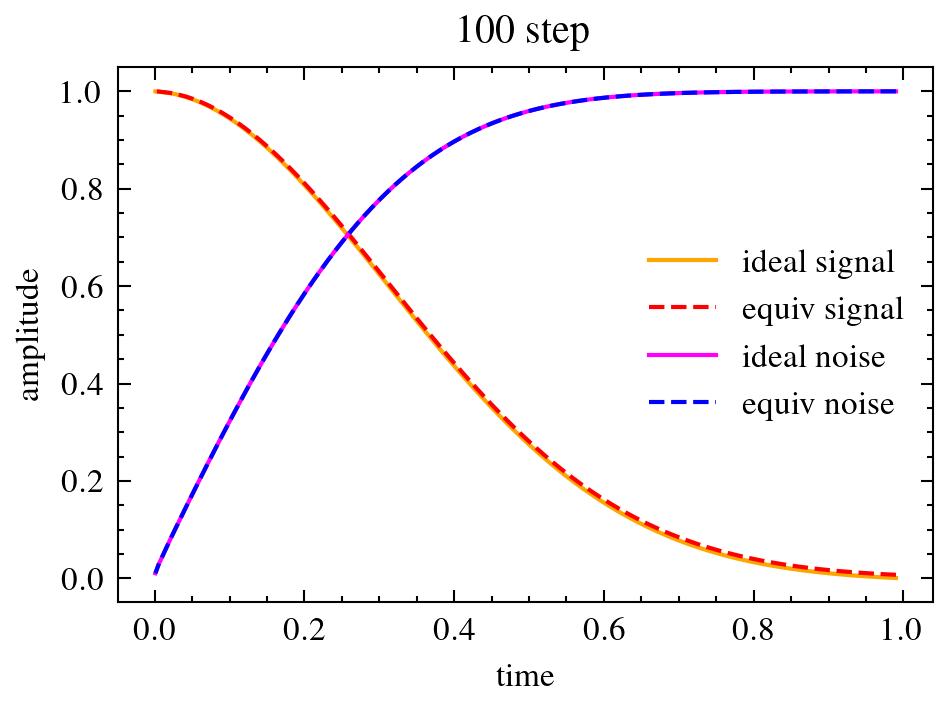}} \end{subfigure}
		\begin{subfigure}[]{\includegraphics[width=0.32\textwidth]{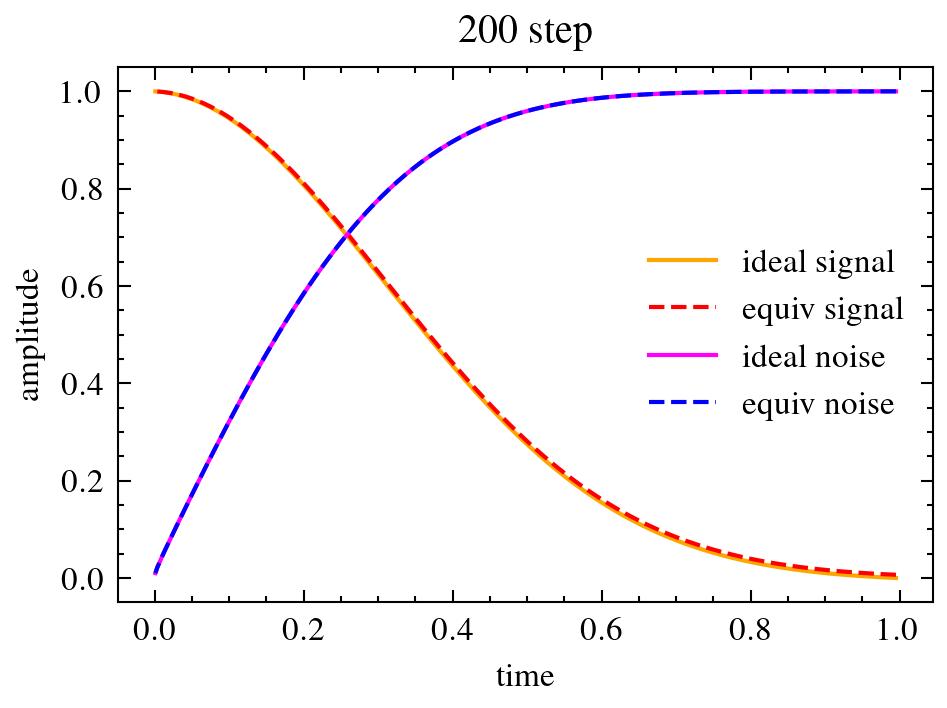}} \end{subfigure}
		\caption{dpmsolver++2s equivalent marginal coefficients and ideal marginal coefficients \quad  (a) 18 step \quad (b) 99 step  \quad (c) 201 step} 
		\label{figure:dpmsolverpp2s_equiv_coeff}
	\end{figure}

	Figure \ref{figure:deis_equiv_coeff} shows the results for the DEIS(tab3) algorithm, Figure \ref{figure:dpmsolver3s_equiv_coeff} presents the results for the third-order DPMSolver, and Figure \ref{figure:dpmsolverpp2s_equiv_coeff} illustrates the results for the second-order DPMSolver++. It can be observed that these higher-order sampling algorithms exhibit the same properties and can also be represented using the Natural Inference framework. 
	
	Table \ref{table:deis_coeff_18} provides the coefficient matrix for the third-order DEIS algorithm (18 steps). Table \ref{table:dpmsolver2s_coeff_18} and Table \ref{table:dpmsolver3s_coeff_18} present the coefficient matrices for the second-order and third-order DPMSolver algorithms (18 steps). Table \ref{table:dpmsolverpp2s_coeff_18} and Table \ref{table:dpmsolverpp3s_coeff_18} provide the coefficient matrices for the second-order and third-order DPMSolver++ algorithms (18 steps).

	\subsection{Represent SDE Euler and ODE Euler with Natural Inference framework}

\begin{figure}
		\centering
		\begin{subfigure}[]{\includegraphics[width=0.32\textwidth]{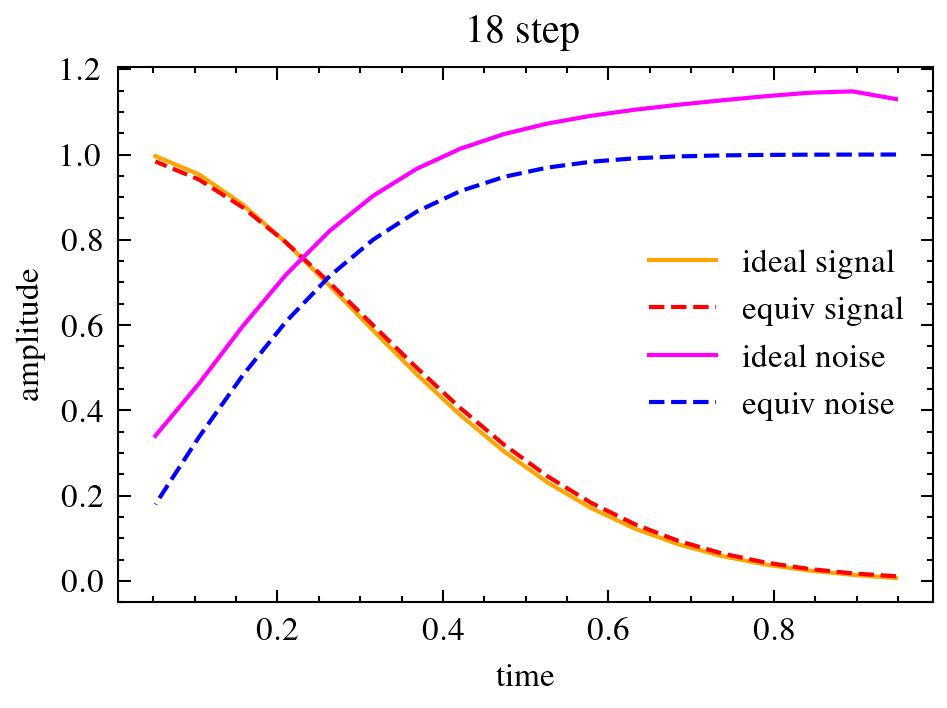}} \end{subfigure}
		\begin{subfigure}[]{\includegraphics[width=0.32\textwidth]{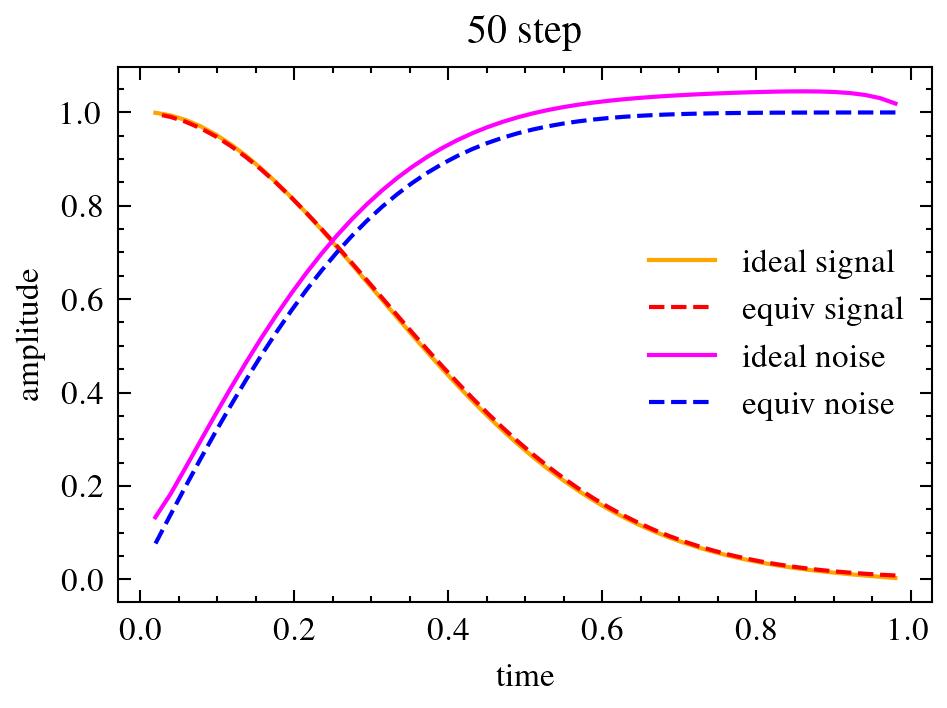}} \end{subfigure}
		\begin{subfigure}[]{\includegraphics[width=0.32\textwidth]{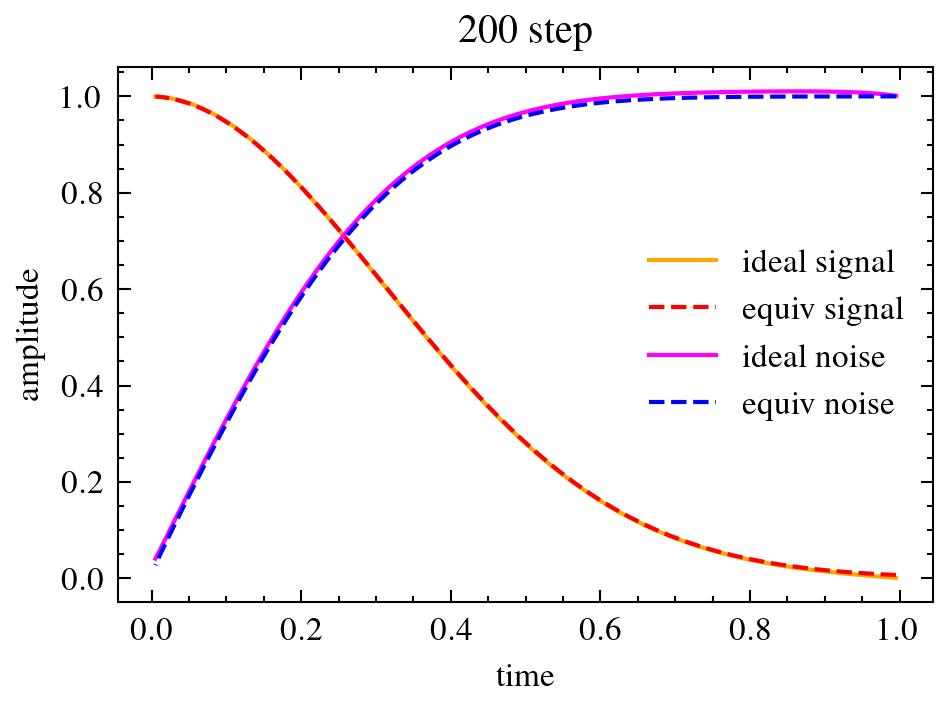}} \end{subfigure}
		\caption{SDE Euler equivalent marginal coefficients and ideal marginal coefficients \quad  (a) 18 step \quad (b) 50 step \quad (c) 200 step} 
		\label{figure:sde_euler_equiv_coeff}
	\end{figure}

\begin{figure}
		\centering
		\begin{subfigure}[]{\includegraphics[width=0.32\textwidth]{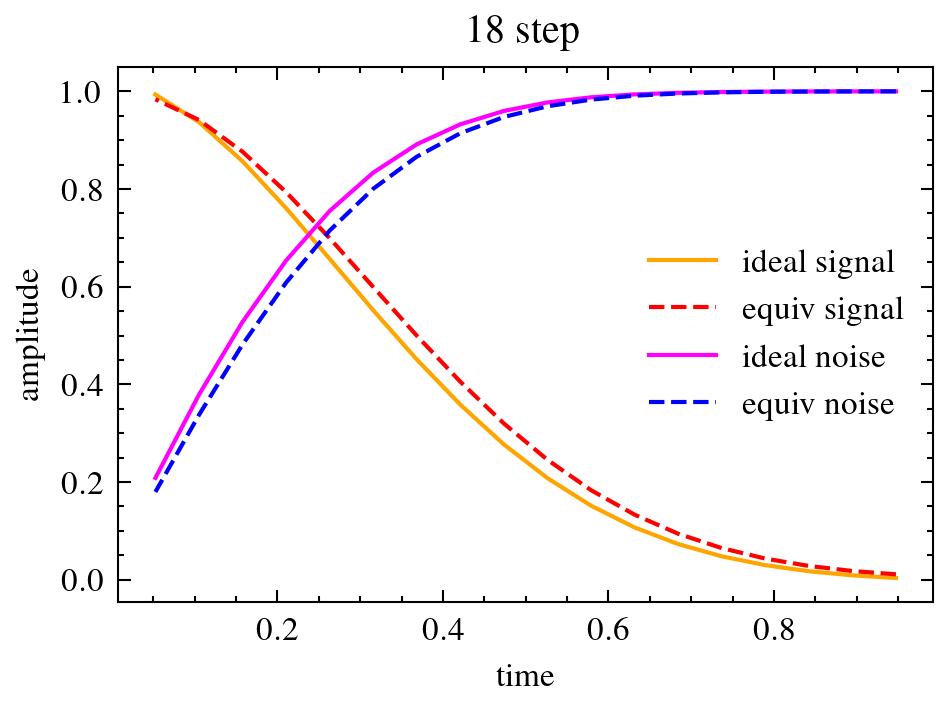}} \end{subfigure}
		\begin{subfigure}[]{\includegraphics[width=0.32\textwidth]{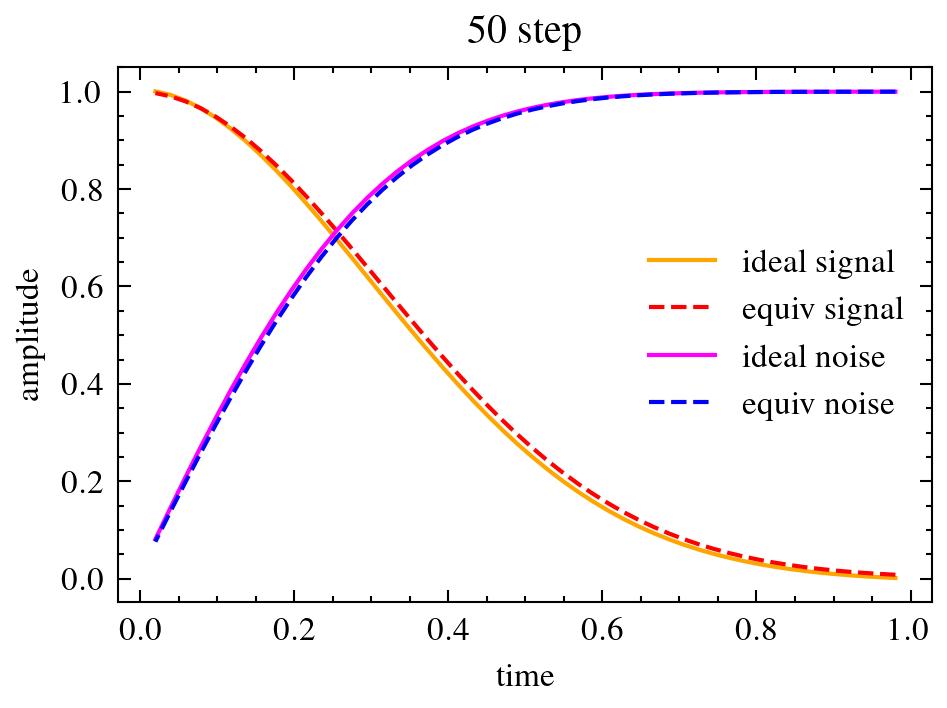}} \end{subfigure}
		\begin{subfigure}[]{\includegraphics[width=0.32\textwidth]{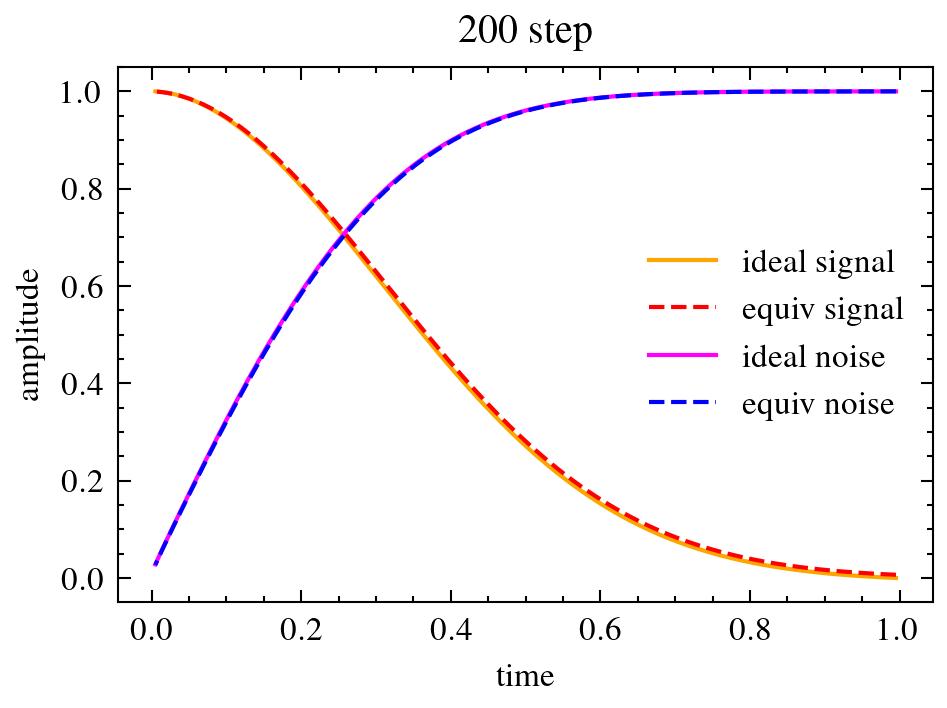}} \end{subfigure}
		\caption{ODE Euler equivalent marginal coefficients and ideal marginal coefficients \quad  (a) 18 step \quad (b) 50 step \quad (c) 200 step} 
		\label{figure:ode_euler_equiv_coeff}
	\end{figure}

	For SDE Euler and ODE Euler, the expressions for each $x_t$ can also be automatically computed using SymPy. The compuation results indicate that these two algorithms yield results similar to previous algorithms, but they suffer from relatively larger errors, especially when the number of steps is small. For details, see Figure \ref{figure:sde_euler_equiv_coeff} and Figure \ref{figure:ode_euler_equiv_coeff}.

	\FloatBarrier
	
	\section{Over Enhancement phenomenon} \label{section:over_enhance}
	
	This subsection introduces a phenomenon called \textbf{over-enhancement}, which influences the design of the coefficient matrix.
	
	\paragraph{Over-enhancement phenomenon} As established in Section \ref{section:new_target_view}, fitting $x_0$ essentially compensates for higher-frequency components that are drowned out by noise, which in turn enriching the details in the input image. Consequently, the model can be regarded as an image quality enhancement operator. Furthermore, the preceding analysis has also demonstrated that Self Guidance operations are also capable of improving image quality. It follows that when \textbf{a multitude of enhancement operations} are employed, the propensity for the \textbf{over-enhancement phenomenon} is heightened. This can occur, for instance, with frequent and dense model enhancements—where sampling time points are proximate and time intervals are minimal—or through the application of potent Self Guidance operations.

	This phenomenon can be intuitively illustrated through the following procedure: Given a standard input image, by fixing the model time point $t$ and the noise level, and then iteratively applying model enhancements, the \textit{over-enhancement} phenomenon progressively becomes apparent. With a fixed $t$ and a zero time interval between applications, this setup fulfills the condition of dense enhancement, as previously discussed. In this process, the output image from one iteration serves directly as the input for the next. This can be interpreted as a form of Self Guidance operating without a \textit{tail} component and with $\lambda=1$ at a high intensity level.
	
\begin{figure}
		\centering
		\begin{subfigure}[]{\includegraphics[width=0.32\textwidth]{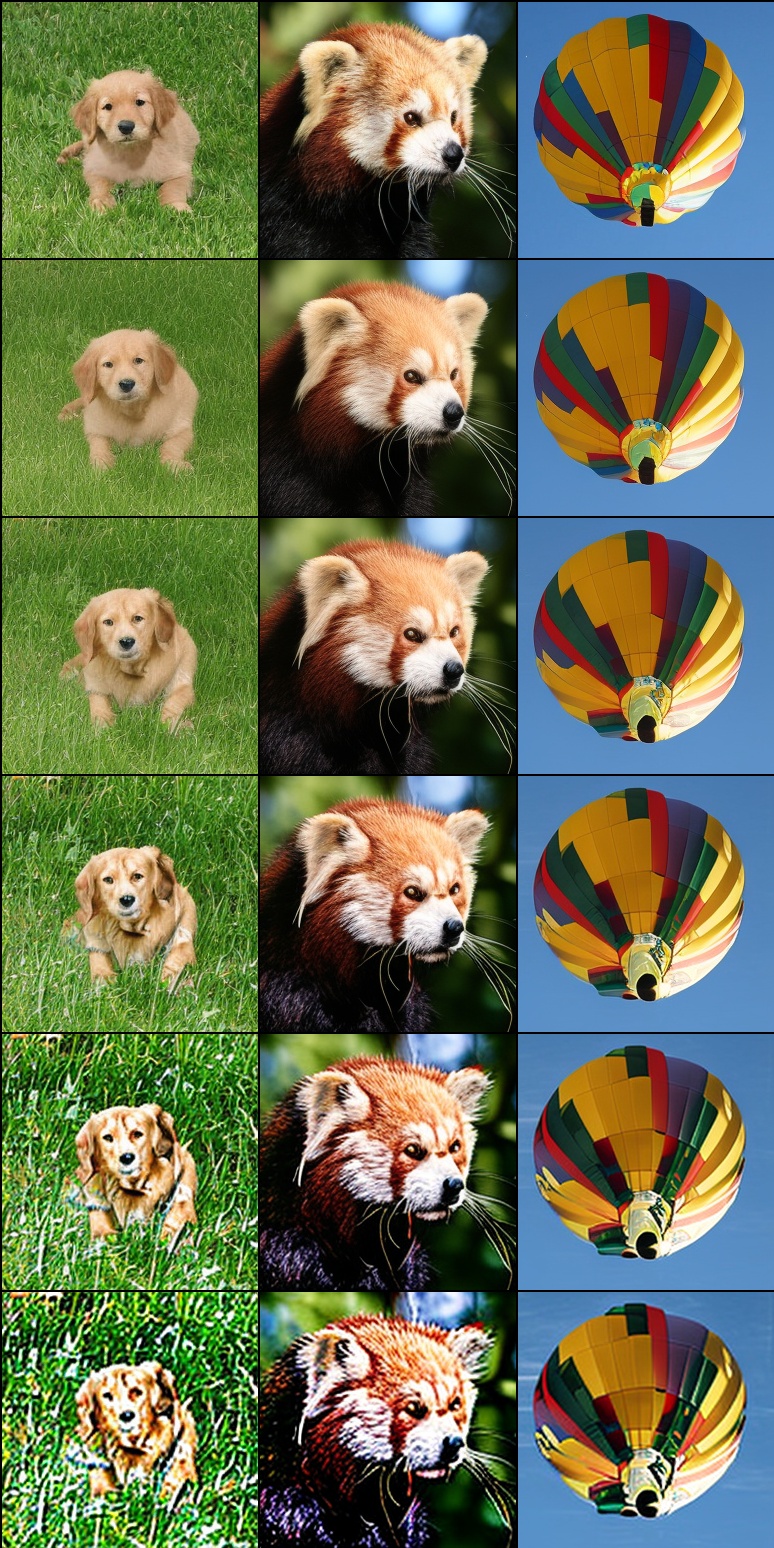}} \end{subfigure}
		\begin{subfigure}[]{\includegraphics[width=0.32\textwidth]{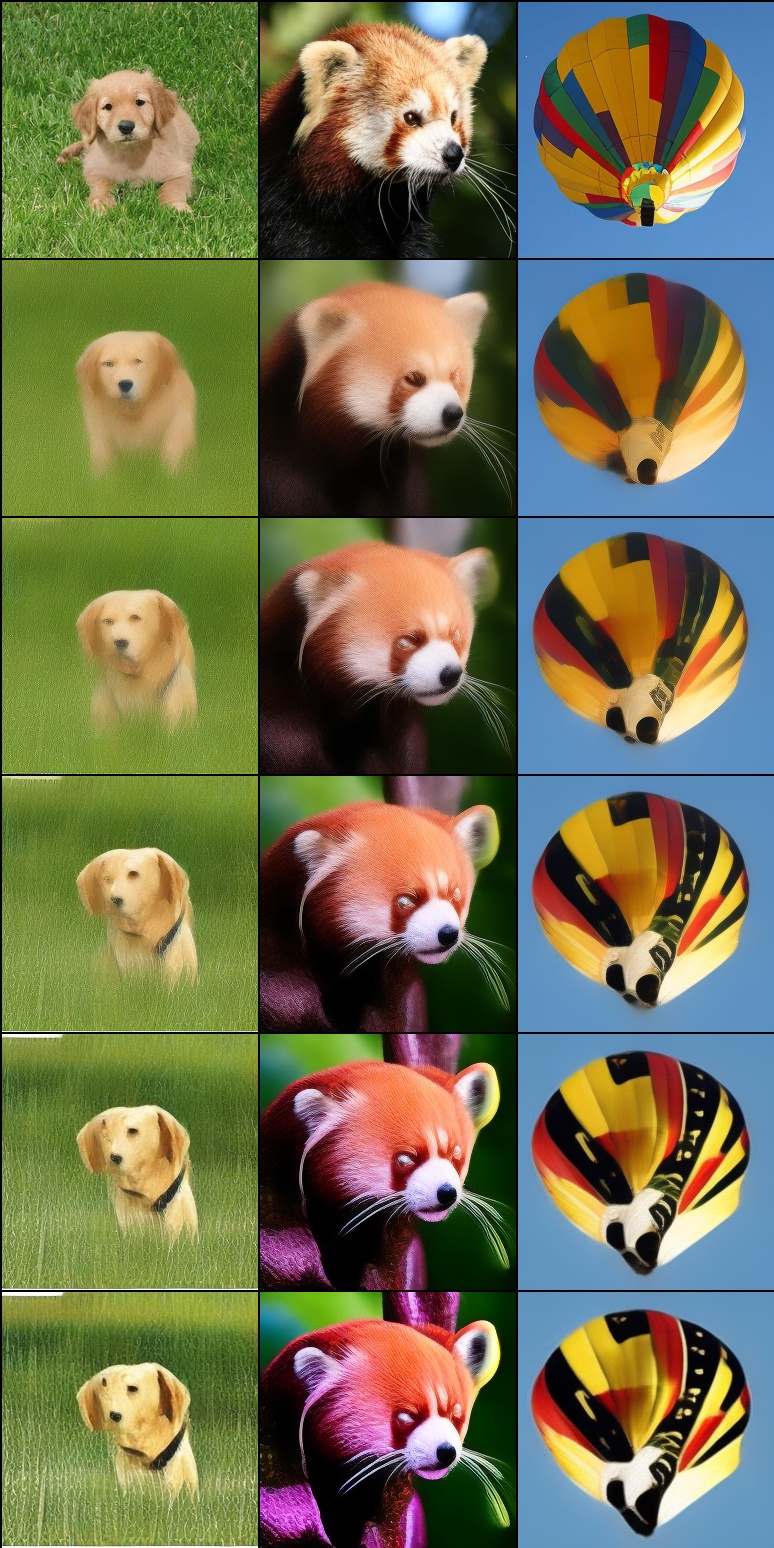}} \end{subfigure}
		\begin{subfigure}[]{\includegraphics[width=0.32\textwidth]{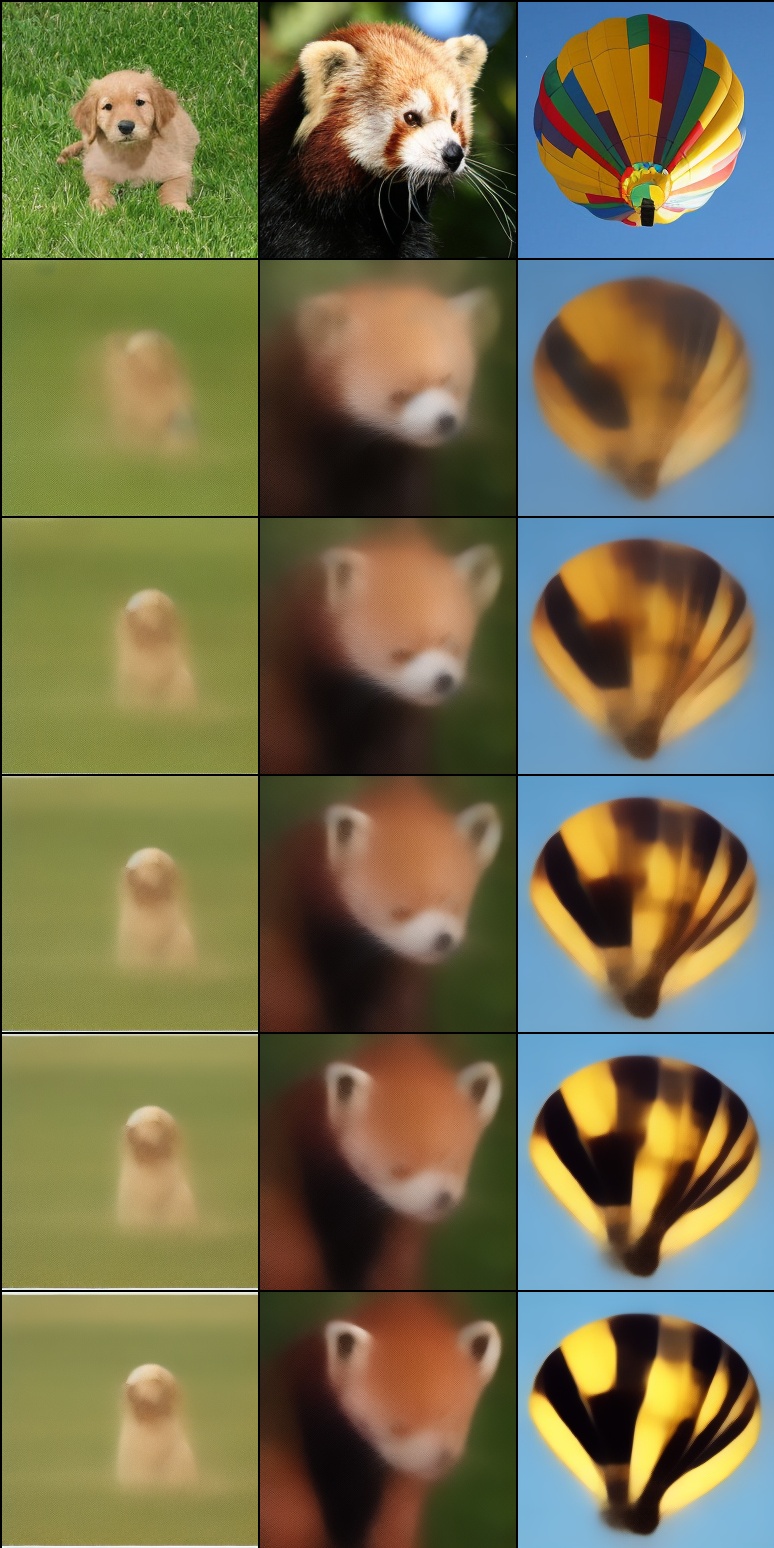}} \end{subfigure}
		\caption{over enhancing phenomenon on ddpm latent model trained on imageget-256 \quad (a) t=100 \quad (b) t=300 \quad (c) t=500} 
		\label{figure:over enhancing}
	\end{figure}

	The specific visual outcomes are depicted in Figure \ref{figure:over enhancing}. The first row displays the original image, while the second row presents the result after a single application of the model. A decrease in image quality is observable at this stage, primarily because some high-frequency components are drowned out by the noise, leading to a low signal-to-noise ratio, which makes it difficult for the model to predict. From the fourth row onwards, the \textit{over-enhancement} phenomenon progressively emerges and intensifies.

	Furthermore, it is evident that the characteristics of the \textit{over-enhancement} phenomenon vary with different time points $t$. For smaller values of $t$, the image exhibits a pronounced grainy texture and retains more high-frequency information. Conversely, at larger values, the image tends to become overly simplified and deficient in detail.	
	
	\paragraph{Train-test mismatch in input domain} This phenomenon may be related to the \textbf{train-test mismatch} in the model input domain. During training, the input images are natural, containing complete frequency components, while during inference, the input images are not complete frequency components. Early in the inference process, the model usually only has low-frequency components, and as $t$ increases, higher-frequency components gradually appear. Therefore, there is a certain difference between these two data domains.
	
\begin{figure}
		\centering
		\begin{subfigure}[]{\includegraphics[width=0.32\textwidth]{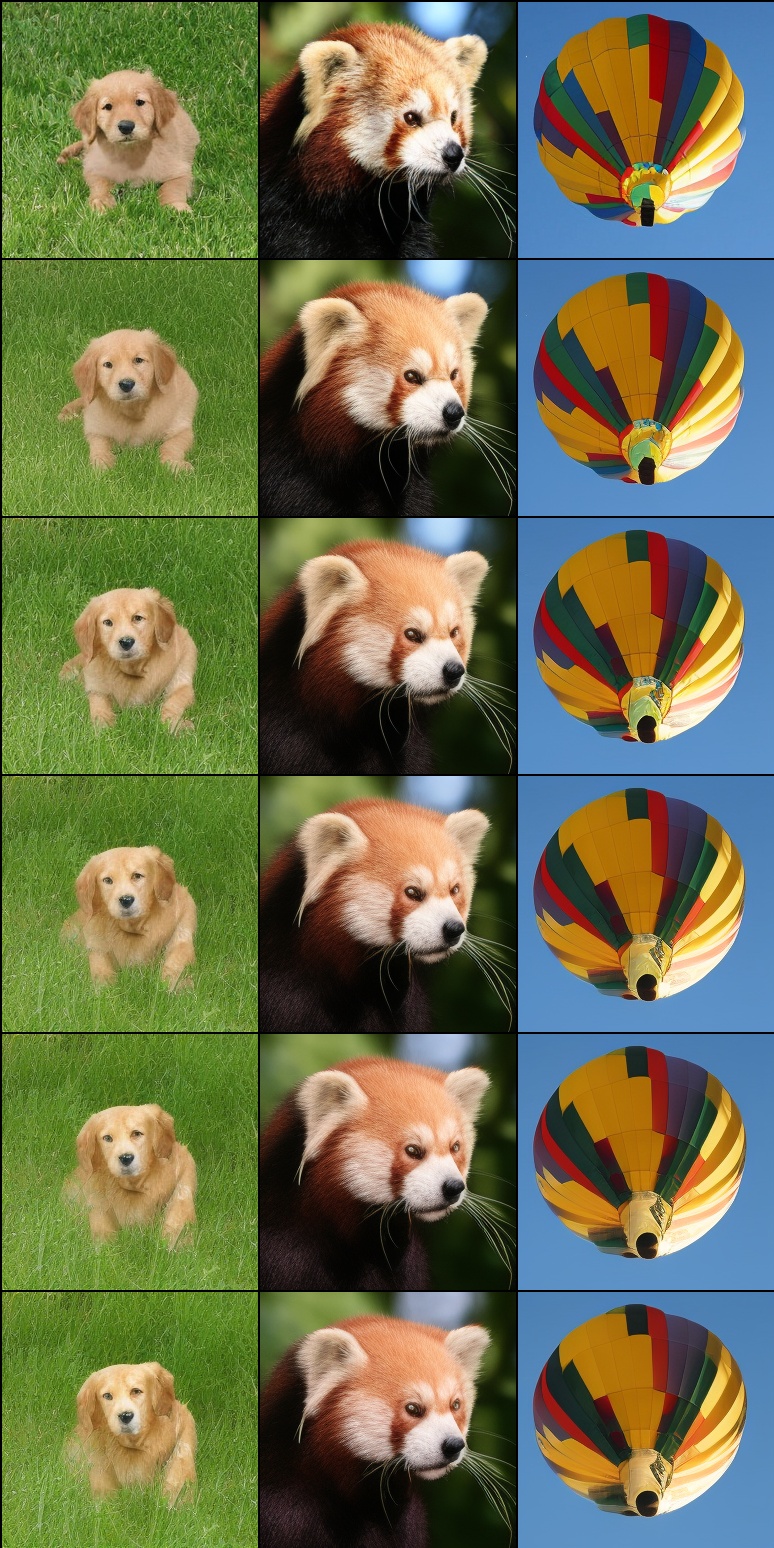}} \end{subfigure}
		\begin{subfigure}[]{\includegraphics[width=0.32\textwidth]{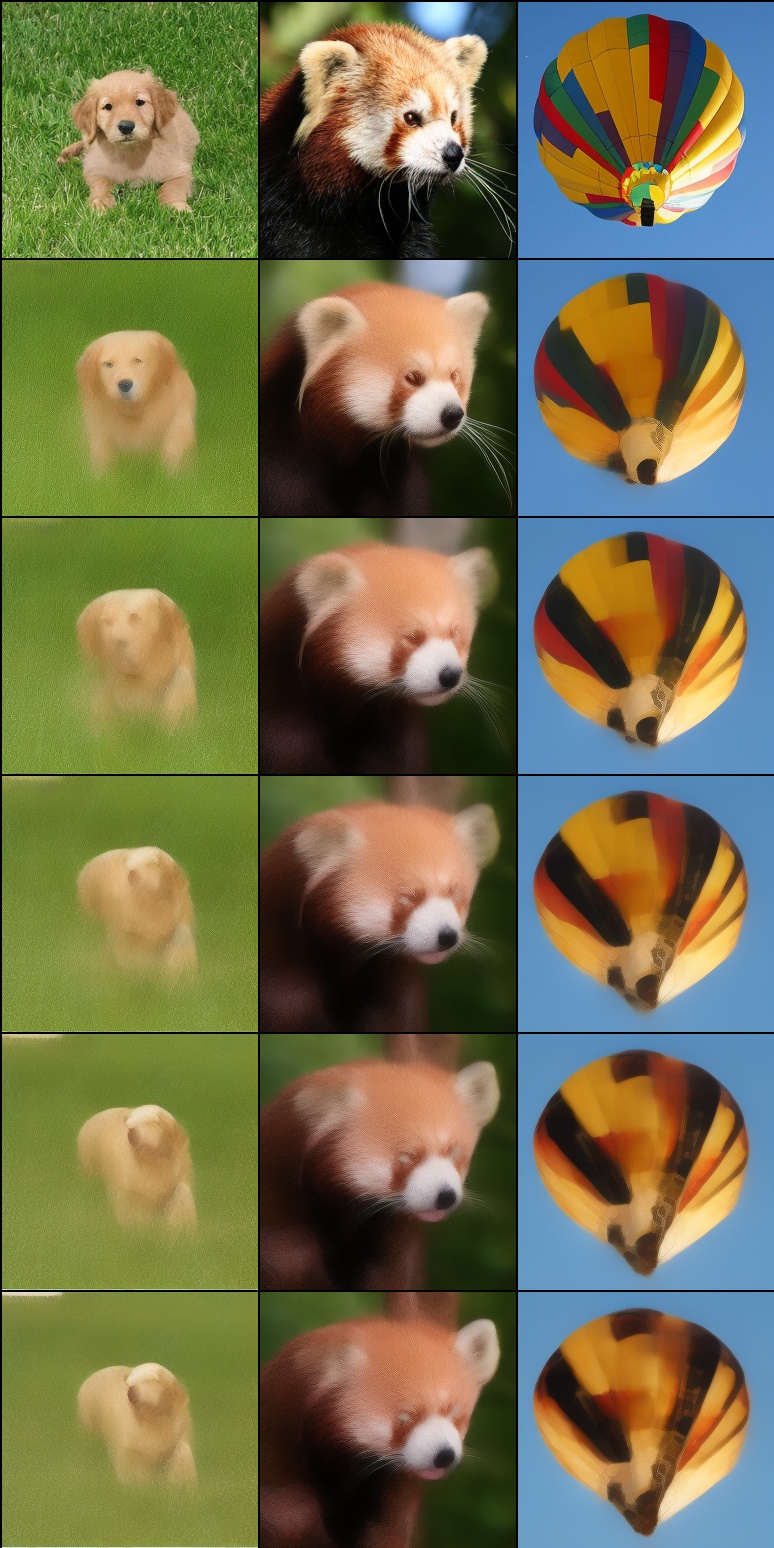}} \end{subfigure}
		\begin{subfigure}[]{\includegraphics[width=0.32\textwidth]{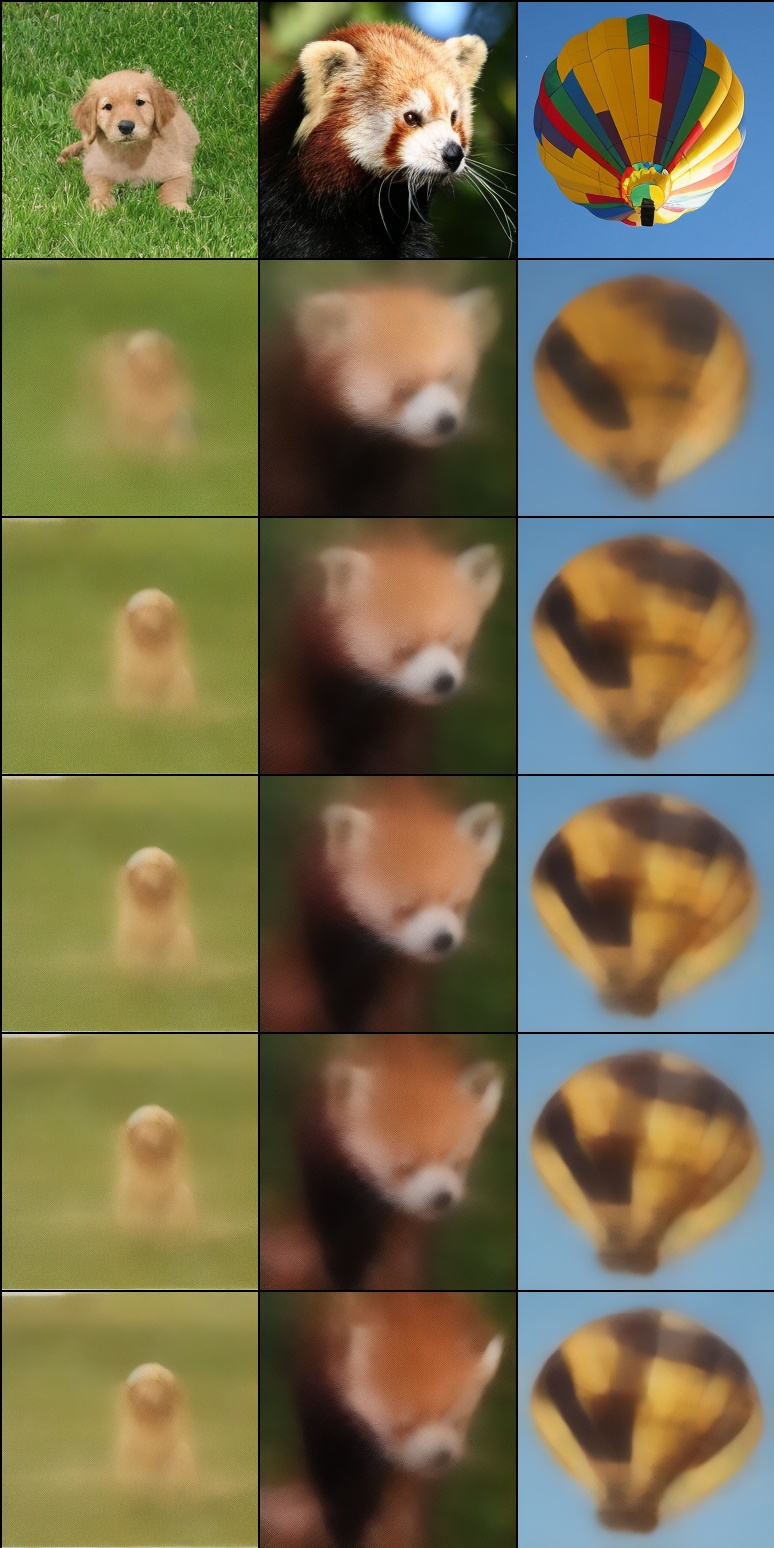}} \end{subfigure}
		\caption{improved over enhancing phenomenon after finetuning with model output images \quad (a) t=100 \quad (b) t=300 \quad (c) t=500} 
		\label{figure:improved_over enhancing}
	\end{figure}

	To validate this analysis, a straightforward experiment can be conducted. For a model demonstrably exhibiting the \textit{over-enhancement} phenomenon, its training data is augmented by incorporating samples akin to those depicted in the second and third rows of Figure \ref{figure:over enhancing}. Subsequent to fine-tuning this modified model, over-enhanced images are generated using the identical procedure. The results of this experiment are illustrated in Figure \ref{figure:improved_over enhancing}. It is evident that the image quality has significantly improved: for smaller $t$, graininess is notably diminished, while for larger $t$, a greater degree of detail is preserved.

	\FloatBarrier

	\section{Coefficient matrixes}
	
	\subsection{DDPM coefficient matrix}
	
	\begin{table}[!ht]
		\centering
		\caption{DDPM's signal coefficient matrix on Natural Inference framework} 
		\begin{adjustbox}{width=1\textwidth}
		\begin{tabular}{cccccccccccccccccccc}
			\toprule
			\textbf{time} & \textbf{999} & \textbf{940} & \textbf{881} & \textbf{823} & \textbf{764} & \textbf{705} & \textbf{646} & \textbf{588} & \textbf{529} & \textbf{470} & \textbf{411} & \textbf{353} & \textbf{294} & \textbf{235} & \textbf{176} & \textbf{118} & \textbf{059} & \textbf{000} & \textbf{sum} \\
			\midrule
			\textbf{940} & 0.008 & 0.0 & 0.0 & 0.0 & 0.0 & 0.0 & 0.0 & 0.0 & 0.0 & 0.0 & 0.0 & 0.0 & 0.0 & 0.0 & 0.0 & 0.0 & 0.0 & 0.0 & 0.008 \\ 
			\textbf{881} & 0.005 & 0.013 & 0.0 & 0.0 & 0.0 & 0.0 & 0.0 & 0.0 & 0.0 & 0.0 & 0.0 & 0.0 & 0.0 & 0.0 & 0.0 & 0.0 & 0.0 & 0.0 & 0.017 \\ 
			\textbf{823} & 0.003 & 0.008 & 0.02 & 0.0 & 0.0 & 0.0 & 0.0 & 0.0 & 0.0 & 0.0 & 0.0 & 0.0 & 0.0 & 0.0 & 0.0 & 0.0 & 0.0 & 0.0 & 0.031 \\ 
			\textbf{764} & 0.002 & 0.005 & 0.013 & 0.032 & 0.0 & 0.0 & 0.0 & 0.0 & 0.0 & 0.0 & 0.0 & 0.0 & 0.0 & 0.0 & 0.0 & 0.0 & 0.0 & 0.0 & 0.051 \\ 
			\textbf{705} & 0.001 & 0.003 & 0.008 & 0.02 & 0.047 & 0.0 & 0.0 & 0.0 & 0.0 & 0.0 & 0.0 & 0.0 & 0.0 & 0.0 & 0.0 & 0.0 & 0.0 & 0.0 & 0.079 \\ 
			\textbf{646} & 0.001 & 0.002 & 0.005 & 0.013 & 0.031 & 0.067 & 0.0 & 0.0 & 0.0 & 0.0 & 0.0 & 0.0 & 0.0 & 0.0 & 0.0 & 0.0 & 0.0 & 0.0 & 0.119 \\ 
			\textbf{588} & 0.0 & 0.001 & 0.004 & 0.009 & 0.021 & 0.046 & 0.09 & 0.0 & 0.0 & 0.0 & 0.0 & 0.0 & 0.0 & 0.0 & 0.0 & 0.0 & 0.0 & 0.0 & 0.172 \\ 
			\textbf{529} & 0.0 & 0.001 & 0.003 & 0.006 & 0.015 & 0.032 & 0.062 & 0.12 & 0.0 & 0.0 & 0.0 & 0.0 & 0.0 & 0.0 & 0.0 & 0.0 & 0.0 & 0.0 & 0.24 \\ 
			\textbf{470} & 0.0 & 0.001 & 0.002 & 0.005 & 0.01 & 0.022 & 0.044 & 0.085 & 0.154 & 0.0 & 0.0 & 0.0 & 0.0 & 0.0 & 0.0 & 0.0 & 0.0 & 0.0 & 0.323 \\ 
			\textbf{411} & 0.0 & 0.0 & 0.001 & 0.003 & 0.007 & 0.016 & 0.031 & 0.06 & 0.109 & 0.192 & 0.0 & 0.0 & 0.0 & 0.0 & 0.0 & 0.0 & 0.0 & 0.0 & 0.42 \\ 
			\textbf{353} & 0.0 & 0.0 & 0.001 & 0.002 & 0.005 & 0.011 & 0.022 & 0.042 & 0.076 & 0.135 & 0.232 & 0.0 & 0.0 & 0.0 & 0.0 & 0.0 & 0.0 & 0.0 & 0.526 \\ 
			\textbf{294} & 0.0 & 0.0 & 0.001 & 0.002 & 0.003 & 0.007 & 0.015 & 0.028 & 0.051 & 0.091 & 0.156 & 0.284 & 0.0 & 0.0 & 0.0 & 0.0 & 0.0 & 0.0 & 0.639 \\ 
			\textbf{235} & 0.0 & 0.0 & 0.0 & 0.001 & 0.002 & 0.005 & 0.009 & 0.018 & 0.033 & 0.057 & 0.099 & 0.18 & 0.345 & 0.0 & 0.0 & 0.0 & 0.0 & 0.0 & 0.749 \\ 
			\textbf{176} & 0.0 & 0.0 & 0.0 & 0.001 & 0.001 & 0.003 & 0.005 & 0.01 & 0.018 & 0.032 & 0.056 & 0.101 & 0.195 & 0.426 & 0.0 & 0.0 & 0.0 & 0.0 & 0.849 \\ 
			\textbf{118} & 0.0 & 0.0 & 0.0 & 0.0 & 0.001 & 0.001 & 0.002 & 0.005 & 0.008 & 0.015 & 0.026 & 0.047 & 0.09 & 0.196 & 0.536 & 0.0 & 0.0 & 0.0 & 0.927 \\ 
			\textbf{059} & 0.0 & 0.0 & 0.0 & 0.0 & 0.0 & 0.0 & 0.001 & 0.001 & 0.002 & 0.004 & 0.007 & 0.013 & 0.024 & 0.053 & 0.145 & 0.728 & 0.0 & 0.0 & 0.98 \\ 
			\textbf{000} & 0.0 & 0.0 & 0.0 & 0.0 & 0.0 & 0.0 & 0.0 & 0.0 & 0.0 & 0.0 & 0.0 & 0.0 & 0.0 & 0.0 & 0.0 & 0.002 & 0.998 & 0.0 & 1.0 \\ 
			\textbf{-01} & 0.0 & 0.0 & 0.0 & 0.0 & 0.0 & 0.0 & 0.0 & 0.0 & 0.0 & 0.0 & 0.0 & 0.0 & 0.0 & 0.0 & 0.0 & 0.0 & 0.0 & 1.0 & 1.0 \\ 
			\bottomrule
		\end{tabular}
	\end{adjustbox}
	\label{table:ddpm_coeff_18}
	\end{table}

	\begin{table}[!ht]
		\centering
		\caption{DDPM's noise coefficient matrix on Natural Inference framework} 
		\begin{adjustbox}{width=1\textwidth}
		\begin{tabular}{ccccccccccccccccccccc}
			\toprule
			\textbf{time} & \textbf{999} & \textbf{940} & \textbf{881} & \textbf{823} & \textbf{764} & \textbf{705} & \textbf{646} & \textbf{588} & \textbf{529} & \textbf{470} & \textbf{411} & \textbf{353} & \textbf{294} & \textbf{235} & \textbf{176} & \textbf{118} & \textbf{059} & \textbf{000} & \textbf{-01} & \textbf{norm} \\
			\midrule
			\textbf{940} & 0.561 & 0.828 & 0.0 & 0.0 & 0.0 & 0.0 & 0.0 & 0.0 & 0.0 & 0.0 & 0.0 & 0.0 & 0.0 & 0.0 & 0.0 & 0.0 & 0.0 & 0.0 & 0.0 & 1.0 \\ 
			\textbf{881} & 0.326 & 0.481 & 0.814 & 0.0 & 0.0 & 0.0 & 0.0 & 0.0 & 0.0 & 0.0 & 0.0 & 0.0 & 0.0 & 0.0 & 0.0 & 0.0 & 0.0 & 0.0 & 0.0 & 1.0 \\ 
			\textbf{823} & 0.197 & 0.292 & 0.494 & 0.795 & 0.0 & 0.0 & 0.0 & 0.0 & 0.0 & 0.0 & 0.0 & 0.0 & 0.0 & 0.0 & 0.0 & 0.0 & 0.0 & 0.0 & 0.0 & 0.999 \\ 
			\textbf{764} & 0.123 & 0.181 & 0.307 & 0.494 & 0.782 & 0.0 & 0.0 & 0.0 & 0.0 & 0.0 & 0.0 & 0.0 & 0.0 & 0.0 & 0.0 & 0.0 & 0.0 & 0.0 & 0.0 & 0.999 \\ 
			\textbf{705} & 0.079 & 0.117 & 0.197 & 0.318 & 0.502 & 0.763 & 0.0 & 0.0 & 0.0 & 0.0 & 0.0 & 0.0 & 0.0 & 0.0 & 0.0 & 0.0 & 0.0 & 0.0 & 0.0 & 0.997 \\ 
			\textbf{646} & 0.052 & 0.077 & 0.131 & 0.211 & 0.333 & 0.506 & 0.741 & 0.0 & 0.0 & 0.0 & 0.0 & 0.0 & 0.0 & 0.0 & 0.0 & 0.0 & 0.0 & 0.0 & 0.0 & 0.993 \\ 
			\textbf{588} & 0.036 & 0.053 & 0.09 & 0.144 & 0.228 & 0.347 & 0.508 & 0.712 & 0.0 & 0.0 & 0.0 & 0.0 & 0.0 & 0.0 & 0.0 & 0.0 & 0.0 & 0.0 & 0.0 & 0.985 \\ 
			\textbf{529} & 0.025 & 0.037 & 0.062 & 0.1 & 0.159 & 0.241 & 0.353 & 0.496 & 0.687 & 0.0 & 0.0 & 0.0 & 0.0 & 0.0 & 0.0 & 0.0 & 0.0 & 0.0 & 0.0 & 0.971 \\ 
			\textbf{470} & 0.018 & 0.026 & 0.044 & 0.071 & 0.112 & 0.17 & 0.249 & 0.349 & 0.485 & 0.653 & 0.0 & 0.0 & 0.0 & 0.0 & 0.0 & 0.0 & 0.0 & 0.0 & 0.0 & 0.946 \\ 
			\textbf{411} & 0.012 & 0.018 & 0.031 & 0.05 & 0.079 & 0.12 & 0.176 & 0.247 & 0.342 & 0.462 & 0.613 & 0.0 & 0.0 & 0.0 & 0.0 & 0.0 & 0.0 & 0.0 & 0.0 & 0.907 \\ 
			\textbf{353} & 0.009 & 0.013 & 0.022 & 0.035 & 0.056 & 0.084 & 0.123 & 0.173 & 0.24 & 0.324 & 0.43 & 0.564 & 0.0 & 0.0 & 0.0 & 0.0 & 0.0 & 0.0 & 0.0 & 0.85 \\ 
			\textbf{294} & 0.006 & 0.009 & 0.015 & 0.024 & 0.037 & 0.057 & 0.083 & 0.117 & 0.162 & 0.218 & 0.29 & 0.38 & 0.513 & 0.0 & 0.0 & 0.0 & 0.0 & 0.0 & 0.0 & 0.769 \\ 
			\textbf{235} & 0.004 & 0.005 & 0.009 & 0.015 & 0.024 & 0.036 & 0.053 & 0.074 & 0.102 & 0.138 & 0.183 & 0.24 & 0.324 & 0.449 & 0.0 & 0.0 & 0.0 & 0.0 & 0.0 & 0.662 \\ 
			\textbf{176} & 0.002 & 0.003 & 0.005 & 0.008 & 0.013 & 0.02 & 0.03 & 0.042 & 0.058 & 0.078 & 0.103 & 0.135 & 0.183 & 0.253 & 0.375 & 0.0 & 0.0 & 0.0 & 0.0 & 0.529 \\ 
			\textbf{118} & 0.001 & 0.001 & 0.002 & 0.004 & 0.006 & 0.009 & 0.014 & 0.019 & 0.027 & 0.036 & 0.048 & 0.062 & 0.084 & 0.117 & 0.173 & 0.285 & 0.0 & 0.0 & 0.0 & 0.375 \\ 
			\textbf{059} & 0.0 & 0.0 & 0.001 & 0.001 & 0.002 & 0.003 & 0.004 & 0.005 & 0.007 & 0.01 & 0.013 & 0.017 & 0.023 & 0.032 & 0.047 & 0.077 & 0.173 & 0.0 & 0.0 & 0.201 \\ 
			\textbf{000} & 0.0 & 0.0 & 0.0 & 0.0 & 0.0 & 0.0 & 0.0 & 0.0 & 0.0 & 0.0 & 0.0 & 0.0 & 0.0 & 0.0 & 0.0 & 0.0 & 0.0 & 0.01 & 0.0 & 0.01 \\ 
			\textbf{-01} & 0.0 & 0.0 & 0.0 & 0.0 & 0.0 & 0.0 & 0.0 & 0.0 & 0.0 & 0.0 & 0.0 & 0.0 & 0.0 & 0.0 & 0.0 & 0.0 & 0.0 & 0.0 & 0.0 & 0.00 \\ 
			\bottomrule
		\end{tabular}
		\end{adjustbox}
		\label{table:ddpm_coeff_eps_18}
	\end{table}
	
	\FloatBarrier
	\subsection{DDIM coefficient matrix}
	
	\begin{table}[!ht]
		\centering
		\caption{DDIM's signal coefficient matrix on the Natural Inference framework} 
		\begin{adjustbox}{width=1\textwidth}
		\begin{tabular}{cccccccccccccccccccc}
			\toprule
			\textbf{time} & \textbf{999} & \textbf{940} & \textbf{881} & \textbf{823} & \textbf{764} & \textbf{705} & \textbf{646} & \textbf{588} & \textbf{529} & \textbf{470} & \textbf{411} & \textbf{353} & \textbf{294} & \textbf{235} & \textbf{176} & \textbf{118} & \textbf{059} & \textbf{000} & \textbf{sum} \\
			\midrule
			\textbf{940} & 0.005 & 0.0 & 0.0 & 0.0 & 0.0 & 0.0 & 0.0 & 0.0 & 0.0 & 0.0 & 0.0 & 0.0 & 0.0 & 0.0 & 0.0 & 0.0 & 0.0 & 0.0 & 0.005 \\ 
			\textbf{881} & 0.005 & 0.008 & 0.0 & 0.0 & 0.0 & 0.0 & 0.0 & 0.0 & 0.0 & 0.0 & 0.0 & 0.0 & 0.0 & 0.0 & 0.0 & 0.0 & 0.0 & 0.0 & 0.013 \\ 
			\textbf{823} & 0.005 & 0.008 & 0.013 & 0.0 & 0.0 & 0.0 & 0.0 & 0.0 & 0.0 & 0.0 & 0.0 & 0.0 & 0.0 & 0.0 & 0.0 & 0.0 & 0.0 & 0.0 & 0.026 \\ 
			\textbf{764} & 0.005 & 0.008 & 0.013 & 0.019 & 0.0 & 0.0 & 0.0 & 0.0 & 0.0 & 0.0 & 0.0 & 0.0 & 0.0 & 0.0 & 0.0 & 0.0 & 0.0 & 0.0 & 0.045 \\ 
			\textbf{705} & 0.005 & 0.008 & 0.013 & 0.019 & 0.028 & 0.0 & 0.0 & 0.0 & 0.0 & 0.0 & 0.0 & 0.0 & 0.0 & 0.0 & 0.0 & 0.0 & 0.0 & 0.0 & 0.074 \\ 
			\textbf{646} & 0.005 & 0.008 & 0.013 & 0.019 & 0.028 & 0.04 & 0.0 & 0.0 & 0.0 & 0.0 & 0.0 & 0.0 & 0.0 & 0.0 & 0.0 & 0.0 & 0.0 & 0.0 & 0.113 \\ 
			\textbf{588} & 0.005 & 0.008 & 0.012 & 0.019 & 0.028 & 0.04 & 0.053 & 0.0 & 0.0 & 0.0 & 0.0 & 0.0 & 0.0 & 0.0 & 0.0 & 0.0 & 0.0 & 0.0 & 0.166 \\ 
			\textbf{529} & 0.005 & 0.008 & 0.012 & 0.019 & 0.028 & 0.039 & 0.052 & 0.07 & 0.0 & 0.0 & 0.0 & 0.0 & 0.0 & 0.0 & 0.0 & 0.0 & 0.0 & 0.0 & 0.234 \\ 
			\textbf{470} & 0.005 & 0.008 & 0.012 & 0.018 & 0.027 & 0.038 & 0.051 & 0.069 & 0.089 & 0.0 & 0.0 & 0.0 & 0.0 & 0.0 & 0.0 & 0.0 & 0.0 & 0.0 & 0.317 \\ 
			\textbf{411} & 0.005 & 0.007 & 0.011 & 0.018 & 0.026 & 0.037 & 0.049 & 0.066 & 0.086 & 0.111 & 0.0 & 0.0 & 0.0 & 0.0 & 0.0 & 0.0 & 0.0 & 0.0 & 0.415 \\ 
			\textbf{353} & 0.004 & 0.007 & 0.011 & 0.017 & 0.024 & 0.034 & 0.046 & 0.062 & 0.08 & 0.104 & 0.132 & 0.0 & 0.0 & 0.0 & 0.0 & 0.0 & 0.0 & 0.0 & 0.521 \\ 
			\textbf{294} & 0.004 & 0.006 & 0.01 & 0.015 & 0.022 & 0.031 & 0.041 & 0.056 & 0.073 & 0.094 & 0.12 & 0.163 & 0.0 & 0.0 & 0.0 & 0.0 & 0.0 & 0.0 & 0.634 \\ 
			\textbf{235} & 0.003 & 0.005 & 0.008 & 0.013 & 0.019 & 0.027 & 0.036 & 0.048 & 0.063 & 0.081 & 0.103 & 0.14 & 0.199 & 0.0 & 0.0 & 0.0 & 0.0 & 0.0 & 0.745 \\ 
			\textbf{176} & 0.003 & 0.004 & 0.007 & 0.01 & 0.015 & 0.021 & 0.029 & 0.038 & 0.05 & 0.065 & 0.082 & 0.112 & 0.159 & 0.25 & 0.0 & 0.0 & 0.0 & 0.0 & 0.845 \\ 
			\textbf{118} & 0.002 & 0.003 & 0.005 & 0.007 & 0.011 & 0.015 & 0.02 & 0.027 & 0.035 & 0.046 & 0.058 & 0.08 & 0.113 & 0.177 & 0.325 & 0.0 & 0.0 & 0.0 & 0.924 \\ 
			\textbf{059} & 0.001 & 0.002 & 0.003 & 0.004 & 0.006 & 0.008 & 0.011 & 0.015 & 0.019 & 0.025 & 0.031 & 0.043 & 0.06 & 0.095 & 0.174 & 0.483 & 0.0 & 0.0 & 0.978 \\ 
			\textbf{000} & 0.0 & 0.0 & 0.0 & 0.0 & 0.0 & 0.0 & 0.001 & 0.001 & 0.001 & 0.001 & 0.002 & 0.002 & 0.003 & 0.005 & 0.009 & 0.024 & 0.951 & 0.0 & 1.0 \\ 
			\textbf{-01} & 0.0 & 0.0 & 0.0 & 0.0 & 0.0 & 0.0 & 0.0 & 0.0 & 0.0 & 0.0 & 0.0 & 0.0 & 0.0 & 0.0 & 0.0 & 0.0 & 0.0 & 1.0 & 1.0 \\ 
			\bottomrule
		\end{tabular}
		\end{adjustbox}
		\label{table:ddim_coeff_18}
	\end{table}

	\FloatBarrier
	\subsection{Flow Matching Coefficient Matrix}
	
	\begin{table}[!ht]
		\centering
		\caption{Flow Matching Euler sampler's signal coefficient matrix on Natural Inference framework} 
		\begin{adjustbox}{width=1\textwidth}
		\begin{tabular}{cccccccccccccccccccc}
			\toprule
			\textbf{time} & \textbf{1.000} & \textbf{0.944} & \textbf{0.889} & \textbf{0.833} & \textbf{0.778} & \textbf{0.722} & \textbf{0.667} & \textbf{0.611} & \textbf{0.556} & \textbf{0.500} & \textbf{0.444} & \textbf{0.389} & \textbf{0.333} & \textbf{0.278} & \textbf{0.222} & \textbf{0.167} & \textbf{0.111} & \textbf{0.056} & \textbf{sum} \\
			\midrule
			\textbf{0.944} & 0.056 & 0.0 & 0.0 & 0.0 & 0.0 & 0.0 & 0.0 & 0.0 & 0.0 & 0.0 & 0.0 & 0.0 & 0.0 & 0.0 & 0.0 & 0.0 & 0.0 & 0.0 & 0.056 \\ 
			\textbf{0.889} & 0.052 & 0.059 & 0.0 & 0.0 & 0.0 & 0.0 & 0.0 & 0.0 & 0.0 & 0.0 & 0.0 & 0.0 & 0.0 & 0.0 & 0.0 & 0.0 & 0.0 & 0.0 & 0.111 \\ 
			\textbf{0.833} & 0.049 & 0.055 & 0.062 & 0.0 & 0.0 & 0.0 & 0.0 & 0.0 & 0.0 & 0.0 & 0.0 & 0.0 & 0.0 & 0.0 & 0.0 & 0.0 & 0.0 & 0.0 & 0.167 \\ 
			\textbf{0.778} & 0.046 & 0.051 & 0.058 & 0.067 & 0.0 & 0.0 & 0.0 & 0.0 & 0.0 & 0.0 & 0.0 & 0.0 & 0.0 & 0.0 & 0.0 & 0.0 & 0.0 & 0.0 & 0.222 \\ 
			\textbf{0.722} & 0.042 & 0.048 & 0.054 & 0.062 & 0.071 & 0.0 & 0.0 & 0.0 & 0.0 & 0.0 & 0.0 & 0.0 & 0.0 & 0.0 & 0.0 & 0.0 & 0.0 & 0.0 & 0.278 \\ 
			\textbf{0.667} & 0.039 & 0.044 & 0.05 & 0.057 & 0.066 & 0.077 & 0.0 & 0.0 & 0.0 & 0.0 & 0.0 & 0.0 & 0.0 & 0.0 & 0.0 & 0.0 & 0.0 & 0.0 & 0.333 \\ 
			\textbf{0.611} & 0.036 & 0.04 & 0.046 & 0.052 & 0.06 & 0.071 & 0.083 & 0.0 & 0.0 & 0.0 & 0.0 & 0.0 & 0.0 & 0.0 & 0.0 & 0.0 & 0.0 & 0.0 & 0.389 \\ 
			\textbf{0.556} & 0.033 & 0.037 & 0.042 & 0.048 & 0.055 & 0.064 & 0.076 & 0.091 & 0.0 & 0.0 & 0.0 & 0.0 & 0.0 & 0.0 & 0.0 & 0.0 & 0.0 & 0.0 & 0.444 \\ 
			\textbf{0.500} & 0.029 & 0.033 & 0.038 & 0.043 & 0.049 & 0.058 & 0.068 & 0.082 & 0.1 & 0.0 & 0.0 & 0.0 & 0.0 & 0.0 & 0.0 & 0.0 & 0.0 & 0.0 & 0.5 \\ 
			\textbf{0.444} & 0.026 & 0.029 & 0.033 & 0.038 & 0.044 & 0.051 & 0.061 & 0.073 & 0.089 & 0.111 & 0.0 & 0.0 & 0.0 & 0.0 & 0.0 & 0.0 & 0.0 & 0.0 & 0.556 \\ 
			\textbf{0.389} & 0.023 & 0.026 & 0.029 & 0.033 & 0.038 & 0.045 & 0.053 & 0.064 & 0.078 & 0.097 & 0.125 & 0.0 & 0.0 & 0.0 & 0.0 & 0.0 & 0.0 & 0.0 & 0.611 \\ 
			\textbf{0.333} & 0.02 & 0.022 & 0.025 & 0.029 & 0.033 & 0.038 & 0.045 & 0.055 & 0.067 & 0.083 & 0.107 & 0.143 & 0.0 & 0.0 & 0.0 & 0.0 & 0.0 & 0.0 & 0.667 \\ 
			\textbf{0.278} & 0.016 & 0.018 & 0.021 & 0.024 & 0.027 & 0.032 & 0.038 & 0.045 & 0.056 & 0.069 & 0.089 & 0.119 & 0.167 & 0.0 & 0.0 & 0.0 & 0.0 & 0.0 & 0.722 \\ 
			\textbf{0.222} & 0.013 & 0.015 & 0.017 & 0.019 & 0.022 & 0.026 & 0.03 & 0.036 & 0.044 & 0.056 & 0.071 & 0.095 & 0.133 & 0.2 & 0.0 & 0.0 & 0.0 & 0.0 & 0.778 \\ 
			\textbf{0.167} & 0.01 & 0.011 & 0.012 & 0.014 & 0.016 & 0.019 & 0.023 & 0.027 & 0.033 & 0.042 & 0.054 & 0.071 & 0.1 & 0.15 & 0.25 & 0.0 & 0.0 & 0.0 & 0.833 \\ 
			\textbf{0.111} & 0.007 & 0.007 & 0.008 & 0.01 & 0.011 & 0.013 & 0.015 & 0.018 & 0.022 & 0.028 & 0.036 & 0.048 & 0.067 & 0.1 & 0.167 & 0.333 & 0.0 & 0.0 & 0.889 \\ 
			\textbf{0.056} & 0.003 & 0.004 & 0.004 & 0.005 & 0.005 & 0.006 & 0.008 & 0.009 & 0.011 & 0.014 & 0.018 & 0.024 & 0.033 & 0.05 & 0.083 & 0.167 & 0.5 & 0.0 & 0.944 \\ 
			\textbf{0.000} & 0.0 & 0.0 & 0.0 & 0.0 & 0.0 & 0.0 & 0.0 & 0.0 & 0.0 & 0.0 & 0.0 & 0.0 & 0.0 & 0.0 & 0.0 & 0.0 & 0.0 & 1.0 & 1.0 \\ 
			\bottomrule
		\end{tabular}
		\end{adjustbox}
		\label{table:flow_euler_coeff_18}
	\end{table}
	
	\FloatBarrier
	\subsection{DEIS coefficient matrix}
	
	\begin{table}[!ht]
		\centering
		\caption{DEIS sampler's signal coefficient matrix on Natural Inference framework} 
		\begin{adjustbox}{width=1\textwidth}
			\begin{tabular}{cccccccccccccccccccc}
				\toprule
				\textbf{time} & \textbf{1.000} & \textbf{0.895} & \textbf{0.796} & \textbf{0.703} & \textbf{0.616} & \textbf{0.534} & \textbf{0.459} & \textbf{0.389} & \textbf{0.324} & \textbf{0.266} & \textbf{0.213} & \textbf{0.167} & \textbf{0.126} & \textbf{0.090} & \textbf{0.061} & \textbf{0.037} & \textbf{0.019} & \textbf{0.007} & \textbf{sum} \\
				\midrule
				\textbf{0.895} & 0.011 & 0.0 & 0.0 & 0.0 & 0.0 & 0.0 & 0.0 & 0.0 & 0.0 & 0.0 & 0.0 & 0.0 & 0.0 & 0.0 & 0.0 & 0.0 & 0.0 & 0.0 & 0.011 \\ 
				\textbf{0.796} & 0.002 & 0.033 & 0.0 & 0.0 & 0.0 & 0.0 & 0.0 & 0.0 & 0.0 & 0.0 & 0.0 & 0.0 & 0.0 & 0.0 & 0.0 & 0.0 & 0.0 & 0.0 & 0.034 \\ 
				\textbf{0.703} & 0.014 & -0.01 & 0.072 & 0.0 & 0.0 & 0.0 & 0.0 & 0.0 & 0.0 & 0.0 & 0.0 & 0.0 & 0.0 & 0.0 & 0.0 & 0.0 & 0.0 & 0.0 & 0.076 \\ 
				\textbf{0.616} & -0.005 & 0.058 & -0.043 & 0.13 & 0.0 & 0.0 & 0.0 & 0.0 & 0.0 & 0.0 & 0.0 & 0.0 & 0.0 & 0.0 & 0.0 & 0.0 & 0.0 & 0.0 & 0.14 \\ 
				\textbf{0.534} & 0.014 & -0.013 & 0.09 & -0.046 & 0.183 & 0.0 & 0.0 & 0.0 & 0.0 & 0.0 & 0.0 & 0.0 & 0.0 & 0.0 & 0.0 & 0.0 & 0.0 & 0.0 & 0.229 \\ 
				\textbf{0.459} & -0.004 & 0.054 & -0.037 & 0.135 & -0.046 & 0.235 & 0.0 & 0.0 & 0.0 & 0.0 & 0.0 & 0.0 & 0.0 & 0.0 & 0.0 & 0.0 & 0.0 & 0.0 & 0.337 \\ 
				\textbf{0.389} & 0.011 & -0.005 & 0.069 & -0.02 & 0.165 & -0.046 & 0.283 & 0.0 & 0.0 & 0.0 & 0.0 & 0.0 & 0.0 & 0.0 & 0.0 & 0.0 & 0.0 & 0.0 & 0.457 \\ 
				\textbf{0.324} & -0.001 & 0.038 & -0.015 & 0.093 & -0.004 & 0.19 & -0.047 & 0.324 & 0.0 & 0.0 & 0.0 & 0.0 & 0.0 & 0.0 & 0.0 & 0.0 & 0.0 & 0.0 & 0.577 \\ 
				\textbf{0.266} & 0.007 & 0.004 & 0.041 & 0.004 & 0.105 & 0.009 & 0.209 & -0.053 & 0.363 & 0.0 & 0.0 & 0.0 & 0.0 & 0.0 & 0.0 & 0.0 & 0.0 & 0.0 & 0.689 \\ 
				\textbf{0.213} & 0.001 & 0.023 & -0.001 & 0.055 & 0.017 & 0.113 & 0.016 & 0.223 & -0.063 & 0.401 & 0.0 & 0.0 & 0.0 & 0.0 & 0.0 & 0.0 & 0.0 & 0.0 & 0.785 \\ 
				\textbf{0.167} & 0.004 & 0.006 & 0.022 & 0.012 & 0.06 & 0.025 & 0.116 & 0.015 & 0.234 & -0.076 & 0.441 & 0.0 & 0.0 & 0.0 & 0.0 & 0.0 & 0.0 & 0.0 & 0.86 \\ 
				\textbf{0.126} & 0.001 & 0.013 & 0.003 & 0.03 & 0.018 & 0.062 & 0.026 & 0.117 & 0.009 & 0.245 & -0.094 & 0.487 & 0.0 & 0.0 & 0.0 & 0.0 & 0.0 & 0.0 & 0.916 \\ 
				\textbf{0.090} & 0.002 & 0.005 & 0.011 & 0.01 & 0.032 & 0.02 & 0.06 & 0.021 & 0.115 & -0.003 & 0.257 & -0.115 & 0.541 & 0.0 & 0.0 & 0.0 & 0.0 & 0.0 & 0.954 \\ 
				\textbf{0.061} & 0.001 & 0.006 & 0.002 & 0.015 & 0.011 & 0.03 & 0.016 & 0.056 & 0.012 & 0.114 & -0.02 & 0.271 & -0.141 & 0.606 & 0.0 & 0.0 & 0.0 & 0.0 & 0.977 \\ 
				\textbf{0.037} & 0.001 & 0.002 & 0.005 & 0.004 & 0.014 & 0.009 & 0.027 & 0.01 & 0.051 & -0.0 & 0.112 & -0.042 & 0.284 & -0.173 & 0.687 & 0.0 & 0.0 & 0.0 & 0.99 \\ 
				\textbf{0.019} & 0.0 & 0.002 & 0.001 & 0.006 & 0.004 & 0.012 & 0.005 & 0.022 & 0.002 & 0.045 & -0.014 & 0.11 & -0.066 & 0.292 & -0.208 & 0.785 & 0.0 & 0.0 & 0.997 \\ 
				\textbf{0.007} & 0.0 & 0.0 & 0.002 & 0.001 & 0.004 & 0.002 & 0.008 & 0.001 & 0.017 & -0.005 & 0.039 & -0.027 & 0.103 & -0.088 & 0.285 & -0.244 & 0.902 & 0.0 & 0.999 \\ 
				\textbf{0.001} & -0.0 & 0.0 & -0.0 & 0.001 & -0.0 & 0.002 & -0.001 & 0.005 & -0.003 & 0.012 & -0.012 & 0.033 & -0.039 & 0.09 & -0.111 & 0.262 & -0.319 & 1.078 & 1.0 \\ 
				\bottomrule
			\end{tabular}
		\end{adjustbox}
		\label{table:deis_coeff_18}
	\end{table}

	\FloatBarrier
	\subsection{DPMSolver coefficient matrix}
	
	\begin{table}[!ht]
		\centering
		\caption{DPMSolver2S's signal coefficient matrix on Natural Inference framework} 
		\begin{adjustbox}{width=1\textwidth}
		\begin{tabular}{cccccccccccccccccccc}
			\toprule
			\textbf{time} & \textbf{1.000} & \textbf{0.946} & \textbf{0.889} & \textbf{0.835} & \textbf{0.778} & \textbf{0.724} & \textbf{0.667} & \textbf{0.614} & \textbf{0.556} & \textbf{0.502} & \textbf{0.445} & \textbf{0.390} & \textbf{0.334} & \textbf{0.277} & \textbf{0.223} & \textbf{0.161} & \textbf{0.112} & \textbf{0.016} & \textbf{sum} \\
			\midrule
			\textbf{0.946} & 0.005 & 0.0 & 0.0 & 0.0 & 0.0 & 0.0 & 0.0 & 0.0 & 0.0 & 0.0 & 0.0 & 0.0 & 0.0 & 0.0 & 0.0 & 0.0 & 0.0 & 0.0 & 0.005 \\ 
			\textbf{0.889} & -0.008 & 0.021 & 0.0 & 0.0 & 0.0 & 0.0 & 0.0 & 0.0 & 0.0 & 0.0 & 0.0 & 0.0 & 0.0 & 0.0 & 0.0 & 0.0 & 0.0 & 0.0 & 0.012 \\ 
			\textbf{0.835} & -0.008 & 0.021 & 0.011 & 0.0 & 0.0 & 0.0 & 0.0 & 0.0 & 0.0 & 0.0 & 0.0 & 0.0 & 0.0 & 0.0 & 0.0 & 0.0 & 0.0 & 0.0 & 0.023 \\ 
			\textbf{0.778} & -0.008 & 0.021 & -0.017 & 0.045 & 0.0 & 0.0 & 0.0 & 0.0 & 0.0 & 0.0 & 0.0 & 0.0 & 0.0 & 0.0 & 0.0 & 0.0 & 0.0 & 0.0 & 0.041 \\ 
			\textbf{0.724} & -0.008 & 0.021 & -0.017 & 0.045 & 0.024 & 0.0 & 0.0 & 0.0 & 0.0 & 0.0 & 0.0 & 0.0 & 0.0 & 0.0 & 0.0 & 0.0 & 0.0 & 0.0 & 0.064 \\ 
			\textbf{0.667} & -0.008 & 0.02 & -0.017 & 0.045 & -0.029 & 0.088 & 0.0 & 0.0 & 0.0 & 0.0 & 0.0 & 0.0 & 0.0 & 0.0 & 0.0 & 0.0 & 0.0 & 0.0 & 0.099 \\ 
			\textbf{0.614} & -0.008 & 0.02 & -0.017 & 0.045 & -0.029 & 0.087 & 0.044 & 0.0 & 0.0 & 0.0 & 0.0 & 0.0 & 0.0 & 0.0 & 0.0 & 0.0 & 0.0 & 0.0 & 0.143 \\ 
			\textbf{0.556} & -0.008 & 0.02 & -0.017 & 0.045 & -0.029 & 0.086 & -0.044 & 0.149 & 0.0 & 0.0 & 0.0 & 0.0 & 0.0 & 0.0 & 0.0 & 0.0 & 0.0 & 0.0 & 0.203 \\ 
			\textbf{0.502} & -0.008 & 0.02 & -0.016 & 0.044 & -0.028 & 0.085 & -0.043 & 0.146 & 0.073 & 0.0 & 0.0 & 0.0 & 0.0 & 0.0 & 0.0 & 0.0 & 0.0 & 0.0 & 0.272 \\ 
			\textbf{0.445} & -0.008 & 0.019 & -0.016 & 0.042 & -0.027 & 0.082 & -0.042 & 0.142 & -0.059 & 0.225 & 0.0 & 0.0 & 0.0 & 0.0 & 0.0 & 0.0 & 0.0 & 0.0 & 0.359 \\ 
			\textbf{0.390} & -0.007 & 0.018 & -0.015 & 0.04 & -0.026 & 0.078 & -0.04 & 0.135 & -0.056 & 0.215 & 0.112 & 0.0 & 0.0 & 0.0 & 0.0 & 0.0 & 0.0 & 0.0 & 0.454 \\ 
			\textbf{0.334} & -0.007 & 0.017 & -0.014 & 0.038 & -0.024 & 0.073 & -0.037 & 0.126 & -0.052 & 0.2 & -0.077 & 0.318 & 0.0 & 0.0 & 0.0 & 0.0 & 0.0 & 0.0 & 0.559 \\ 
			\textbf{0.277} & -0.006 & 0.015 & -0.012 & 0.034 & -0.022 & 0.065 & -0.033 & 0.112 & -0.047 & 0.179 & -0.069 & 0.285 & 0.168 & 0.0 & 0.0 & 0.0 & 0.0 & 0.0 & 0.669 \\ 
			\textbf{0.223} & -0.005 & 0.013 & -0.011 & 0.029 & -0.019 & 0.056 & -0.028 & 0.097 & -0.04 & 0.154 & -0.059 & 0.245 & -0.112 & 0.45 & 0.0 & 0.0 & 0.0 & 0.0 & 0.768 \\ 
			\textbf{0.161} & -0.004 & 0.01 & -0.008 & 0.022 & -0.014 & 0.043 & -0.022 & 0.074 & -0.031 & 0.118 & -0.046 & 0.188 & -0.086 & 0.346 & 0.278 & 0.0 & 0.0 & 0.0 & 0.869 \\ 
			\textbf{0.112} & -0.003 & 0.007 & -0.006 & 0.016 & -0.01 & 0.031 & -0.016 & 0.054 & -0.023 & 0.086 & -0.033 & 0.137 & -0.063 & 0.252 & -0.235 & 0.735 & 0.0 & 0.0 & 0.932 \\ 
			\textbf{0.016} & -0.001 & 0.001 & -0.001 & 0.003 & -0.002 & 0.006 & -0.003 & 0.01 & -0.004 & 0.015 & -0.006 & 0.024 & -0.011 & 0.045 & -0.042 & 0.13 & 0.833 & 0.0 & 0.998 \\ 
			\textbf{0.001} & -0.0 & 0.0 & -0.0 & 0.0 & -0.0 & 0.001 & -0.0 & 0.002 & -0.001 & 0.003 & -0.001 & 0.004 & -0.002 & 0.007 & -0.007 & 0.022 & -4.895 & 5.867 & 1.0 \\ 
			\bottomrule
		\end{tabular}
		\end{adjustbox}
		\label{table:dpmsolver2s_coeff_18}
	\end{table}

	\begin{table}[!ht]
		\centering
		\caption{DPMSolver3S's signal coefficient matrix on Natural Inference framework} 
		\begin{adjustbox}{width=1\textwidth}
		\begin{tabular}{cccccccccccccccccccc}
			\toprule
			\textbf{time} & \textbf{1.000} & \textbf{0.948} & \textbf{0.892} & \textbf{0.834} & \textbf{0.782} & \textbf{0.727} & \textbf{0.667} & \textbf{0.615} & \textbf{0.560} & \textbf{0.500} & \textbf{0.447} & \textbf{0.391} & \textbf{0.334} & \textbf{0.273} & \textbf{0.217} & \textbf{0.167} & \textbf{0.044} & \textbf{0.009} & \textbf{sum} \\
			\midrule
			\textbf{0.948} & 0.004 & 0.0 & 0.0 & 0.0 & 0.0 & 0.0 & 0.0 & 0.0 & 0.0 & 0.0 & 0.0 & 0.0 & 0.0 & 0.0 & 0.0 & 0.0 & 0.0 & 0.0 & 0.004 \\ 
			\textbf{0.892} & -0.004 & 0.016 & 0.0 & 0.0 & 0.0 & 0.0 & 0.0 & 0.0 & 0.0 & 0.0 & 0.0 & 0.0 & 0.0 & 0.0 & 0.0 & 0.0 & 0.0 & 0.0 & 0.012 \\ 
			\textbf{0.834} & 0.019 & -0.033 & 0.037 & 0.0 & 0.0 & 0.0 & 0.0 & 0.0 & 0.0 & 0.0 & 0.0 & 0.0 & 0.0 & 0.0 & 0.0 & 0.0 & 0.0 & 0.0 & 0.024 \\ 
			\textbf{0.782} & 0.019 & -0.033 & 0.037 & 0.016 & 0.0 & 0.0 & 0.0 & 0.0 & 0.0 & 0.0 & 0.0 & 0.0 & 0.0 & 0.0 & 0.0 & 0.0 & 0.0 & 0.0 & 0.039 \\ 
			\textbf{0.727} & 0.019 & -0.033 & 0.037 & -0.012 & 0.052 & 0.0 & 0.0 & 0.0 & 0.0 & 0.0 & 0.0 & 0.0 & 0.0 & 0.0 & 0.0 & 0.0 & 0.0 & 0.0 & 0.063 \\ 
			\textbf{0.667} & 0.019 & -0.033 & 0.037 & 0.049 & -0.078 & 0.104 & 0.0 & 0.0 & 0.0 & 0.0 & 0.0 & 0.0 & 0.0 & 0.0 & 0.0 & 0.0 & 0.0 & 0.0 & 0.099 \\ 
			\textbf{0.615} & 0.019 & -0.033 & 0.037 & 0.049 & -0.077 & 0.104 & 0.042 & 0.0 & 0.0 & 0.0 & 0.0 & 0.0 & 0.0 & 0.0 & 0.0 & 0.0 & 0.0 & 0.0 & 0.141 \\ 
			\textbf{0.560} & 0.019 & -0.032 & 0.036 & 0.048 & -0.076 & 0.103 & -0.024 & 0.125 & 0.0 & 0.0 & 0.0 & 0.0 & 0.0 & 0.0 & 0.0 & 0.0 & 0.0 & 0.0 & 0.198 \\ 
			\textbf{0.500} & 0.019 & -0.032 & 0.036 & 0.047 & -0.075 & 0.101 & 0.093 & -0.134 & 0.219 & 0.0 & 0.0 & 0.0 & 0.0 & 0.0 & 0.0 & 0.0 & 0.0 & 0.0 & 0.274 \\ 
			\textbf{0.447} & 0.018 & -0.031 & 0.035 & 0.046 & -0.073 & 0.098 & 0.09 & -0.13 & 0.213 & 0.089 & 0.0 & 0.0 & 0.0 & 0.0 & 0.0 & 0.0 & 0.0 & 0.0 & 0.356 \\ 
			\textbf{0.391} & 0.017 & -0.029 & 0.033 & 0.044 & -0.069 & 0.093 & 0.086 & -0.124 & 0.203 & -0.04 & 0.238 & 0.0 & 0.0 & 0.0 & 0.0 & 0.0 & 0.0 & 0.0 & 0.452 \\ 
			\textbf{0.334} & 0.016 & -0.027 & 0.031 & 0.041 & -0.064 & 0.087 & 0.08 & -0.115 & 0.188 & 0.147 & -0.191 & 0.368 & 0.0 & 0.0 & 0.0 & 0.0 & 0.0 & 0.0 & 0.559 \\ 
			\textbf{0.273} & 0.014 & -0.024 & 0.027 & 0.036 & -0.057 & 0.077 & 0.071 & -0.102 & 0.167 & 0.131 & -0.17 & 0.327 & 0.178 & 0.0 & 0.0 & 0.0 & 0.0 & 0.0 & 0.675 \\ 
			\textbf{0.217} & 0.012 & -0.02 & 0.023 & 0.031 & -0.049 & 0.065 & 0.06 & -0.087 & 0.142 & 0.111 & -0.144 & 0.277 & -0.078 & 0.435 & 0.0 & 0.0 & 0.0 & 0.0 & 0.778 \\ 
			\textbf{0.167} & 0.01 & -0.017 & 0.019 & 0.025 & -0.039 & 0.053 & 0.049 & -0.07 & 0.115 & 0.09 & -0.117 & 0.225 & 0.248 & -0.336 & 0.605 & 0.0 & 0.0 & 0.0 & 0.859 \\ 
			\textbf{0.044} & 0.003 & -0.005 & 0.006 & 0.007 & -0.012 & 0.016 & 0.015 & -0.021 & 0.035 & 0.027 & -0.035 & 0.068 & 0.074 & -0.101 & 0.181 & 0.73 & 0.0 & 0.0 & 0.987 \\ 
			\textbf{0.009} & 0.001 & -0.001 & 0.001 & 0.002 & -0.003 & 0.004 & 0.004 & -0.006 & 0.009 & 0.007 & -0.009 & 0.018 & 0.02 & -0.027 & 0.048 & -1.201 & 2.132 & 0.0 & 0.999 \\ 
			\textbf{0.001} & 0.0 & -0.0 & 0.0 & 0.001 & -0.001 & 0.001 & 0.001 & -0.001 & 0.002 & 0.002 & -0.002 & 0.005 & 0.005 & -0.007 & 0.013 & 6.607 & -10.588 & 4.963 & 1.0 \\ 
			\bottomrule
		\end{tabular}
		\end{adjustbox}
		\label{table:dpmsolver3s_coeff_18}
	\end{table}

	\FloatBarrier
	\subsection{DPMSolver++ coefficient matrix}
	
	\begin{table}[!ht]
		\centering
		\caption{DPMSolverpp2S's signal coefficient matrix on Natural Inference framework} 
		\begin{adjustbox}{width=1\textwidth}
		\begin{tabular}{cccccccccccccccccccc}
			\toprule
			\textbf{time} & \textbf{1.000} & \textbf{0.946} & \textbf{0.889} & \textbf{0.835} & \textbf{0.778} & \textbf{0.724} & \textbf{0.667} & \textbf{0.614} & \textbf{0.556} & \textbf{0.502} & \textbf{0.445} & \textbf{0.390} & \textbf{0.334} & \textbf{0.277} & \textbf{0.223} & \textbf{0.161} & \textbf{0.112} & \textbf{0.016} & \textbf{sum} \\
			\midrule
			\textbf{0.946} & 0.005 & 0.0 & 0.0 & 0.0 & 0.0 & 0.0 & 0.0 & 0.0 & 0.0 & 0.0 & 0.0 & 0.0 & 0.0 & 0.0 & 0.0 & 0.0 & 0.0 & 0.0 & 0.005 \\ 
			\textbf{0.889} & 0.0 & 0.012 & 0.0 & 0.0 & 0.0 & 0.0 & 0.0 & 0.0 & 0.0 & 0.0 & 0.0 & 0.0 & 0.0 & 0.0 & 0.0 & 0.0 & 0.0 & 0.0 & 0.012 \\ 
			\textbf{0.835} & 0.0 & 0.012 & 0.011 & 0.0 & 0.0 & 0.0 & 0.0 & 0.0 & 0.0 & 0.0 & 0.0 & 0.0 & 0.0 & 0.0 & 0.0 & 0.0 & 0.0 & 0.0 & 0.023 \\ 
			\textbf{0.778} & 0.0 & 0.012 & 0.0 & 0.029 & 0.0 & 0.0 & 0.0 & 0.0 & 0.0 & 0.0 & 0.0 & 0.0 & 0.0 & 0.0 & 0.0 & 0.0 & 0.0 & 0.0 & 0.041 \\ 
			\textbf{0.724} & 0.0 & 0.012 & 0.0 & 0.029 & 0.024 & 0.0 & 0.0 & 0.0 & 0.0 & 0.0 & 0.0 & 0.0 & 0.0 & 0.0 & 0.0 & 0.0 & 0.0 & 0.0 & 0.064 \\ 
			\textbf{0.667} & 0.0 & 0.012 & 0.0 & 0.028 & 0.0 & 0.059 & 0.0 & 0.0 & 0.0 & 0.0 & 0.0 & 0.0 & 0.0 & 0.0 & 0.0 & 0.0 & 0.0 & 0.0 & 0.099 \\ 
			\textbf{0.614} & 0.0 & 0.012 & 0.0 & 0.028 & 0.0 & 0.058 & 0.044 & 0.0 & 0.0 & 0.0 & 0.0 & 0.0 & 0.0 & 0.0 & 0.0 & 0.0 & 0.0 & 0.0 & 0.143 \\ 
			\textbf{0.556} & 0.0 & 0.012 & 0.0 & 0.028 & 0.0 & 0.058 & 0.0 & 0.105 & 0.0 & 0.0 & 0.0 & 0.0 & 0.0 & 0.0 & 0.0 & 0.0 & 0.0 & 0.0 & 0.203 \\ 
			\textbf{0.502} & 0.0 & 0.012 & 0.0 & 0.028 & 0.0 & 0.057 & 0.0 & 0.103 & 0.073 & 0.0 & 0.0 & 0.0 & 0.0 & 0.0 & 0.0 & 0.0 & 0.0 & 0.0 & 0.272 \\ 
			\textbf{0.445} & 0.0 & 0.011 & 0.0 & 0.027 & 0.0 & 0.055 & 0.0 & 0.1 & 0.0 & 0.166 & 0.0 & 0.0 & 0.0 & 0.0 & 0.0 & 0.0 & 0.0 & 0.0 & 0.359 \\ 
			\textbf{0.390} & 0.0 & 0.011 & 0.0 & 0.025 & 0.0 & 0.052 & 0.0 & 0.095 & 0.0 & 0.159 & 0.112 & 0.0 & 0.0 & 0.0 & 0.0 & 0.0 & 0.0 & 0.0 & 0.454 \\ 
			\textbf{0.334} & 0.0 & 0.01 & 0.0 & 0.024 & 0.0 & 0.049 & 0.0 & 0.089 & 0.0 & 0.147 & 0.0 & 0.241 & 0.0 & 0.0 & 0.0 & 0.0 & 0.0 & 0.0 & 0.559 \\ 
			\textbf{0.277} & 0.0 & 0.009 & 0.0 & 0.021 & 0.0 & 0.044 & 0.0 & 0.079 & 0.0 & 0.132 & 0.0 & 0.216 & 0.168 & 0.0 & 0.0 & 0.0 & 0.0 & 0.0 & 0.669 \\ 
			\textbf{0.223} & 0.0 & 0.008 & 0.0 & 0.018 & 0.0 & 0.037 & 0.0 & 0.068 & 0.0 & 0.113 & 0.0 & 0.185 & 0.0 & 0.338 & 0.0 & 0.0 & 0.0 & 0.0 & 0.768 \\ 
			\textbf{0.161} & 0.0 & 0.006 & 0.0 & 0.014 & 0.0 & 0.029 & 0.0 & 0.052 & 0.0 & 0.087 & 0.0 & 0.143 & 0.0 & 0.26 & 0.278 & 0.0 & 0.0 & 0.0 & 0.869 \\ 
			\textbf{0.112} & 0.0 & 0.004 & 0.0 & 0.01 & 0.0 & 0.021 & 0.0 & 0.038 & 0.0 & 0.064 & 0.0 & 0.104 & 0.0 & 0.189 & 0.0 & 0.501 & 0.0 & 0.0 & 0.932 \\ 
			\textbf{0.016} & 0.0 & 0.001 & 0.0 & 0.002 & 0.0 & 0.004 & 0.0 & 0.007 & 0.0 & 0.011 & 0.0 & 0.018 & 0.0 & 0.034 & 0.0 & 0.089 & 0.833 & 0.0 & 0.998 \\ 
			\textbf{0.001} & 0.0 & 0.0 & 0.0 & 0.0 & 0.0 & 0.001 & 0.0 & 0.001 & 0.0 & 0.002 & 0.0 & 0.003 & 0.0 & 0.006 & 0.0 & 0.015 & 0.0 & 0.972 & 1.0 \\ 
			\bottomrule
		\end{tabular}
		\end{adjustbox}
		\label{table:dpmsolverpp2s_coeff_18}
	\end{table}

	\begin{table}[!ht]
		\centering
		\caption{DPMSolverpp3S's signal coefficient matrix on Natural Inference framework} 
		\begin{adjustbox}{width=1\textwidth}
		\begin{tabular}{cccccccccccccccccccc}
			\toprule
			\textbf{time} & \textbf{1.000} & \textbf{0.948} & \textbf{0.892} & \textbf{0.834} & \textbf{0.782} & \textbf{0.727} & \textbf{0.667} & \textbf{0.615} & \textbf{0.560} & \textbf{0.500} & \textbf{0.447} & \textbf{0.391} & \textbf{0.334} & \textbf{0.273} & \textbf{0.217} & \textbf{0.167} & \textbf{0.044} & \textbf{0.009} & \textbf{sum} \\
			\midrule
			\textbf{0.948} & 0.004 & 0.0 & 0.0 & 0.0 & 0.0 & 0.0 & 0.0 & 0.0 & 0.0 & 0.0 & 0.0 & 0.0 & 0.0 & 0.0 & 0.0 & 0.0 & 0.0 & 0.0 & 0.004 \\ 
			\textbf{0.892} & 0.025 & -0.014 & 0.0 & 0.0 & 0.0 & 0.0 & 0.0 & 0.0 & 0.0 & 0.0 & 0.0 & 0.0 & 0.0 & 0.0 & 0.0 & 0.0 & 0.0 & 0.0 & 0.012 \\ 
			\textbf{0.834} & 0.046 & 0.0 & -0.022 & 0.0 & 0.0 & 0.0 & 0.0 & 0.0 & 0.0 & 0.0 & 0.0 & 0.0 & 0.0 & 0.0 & 0.0 & 0.0 & 0.0 & 0.0 & 0.024 \\ 
			\textbf{0.782} & 0.046 & 0.0 & -0.022 & 0.016 & 0.0 & 0.0 & 0.0 & 0.0 & 0.0 & 0.0 & 0.0 & 0.0 & 0.0 & 0.0 & 0.0 & 0.0 & 0.0 & 0.0 & 0.039 \\ 
			\textbf{0.727} & 0.046 & 0.0 & -0.022 & 0.085 & -0.045 & 0.0 & 0.0 & 0.0 & 0.0 & 0.0 & 0.0 & 0.0 & 0.0 & 0.0 & 0.0 & 0.0 & 0.0 & 0.0 & 0.063 \\ 
			\textbf{0.667} & 0.046 & 0.0 & -0.022 & 0.144 & 0.0 & -0.068 & 0.0 & 0.0 & 0.0 & 0.0 & 0.0 & 0.0 & 0.0 & 0.0 & 0.0 & 0.0 & 0.0 & 0.0 & 0.099 \\ 
			\textbf{0.615} & 0.045 & 0.0 & -0.022 & 0.143 & 0.0 & -0.068 & 0.042 & 0.0 & 0.0 & 0.0 & 0.0 & 0.0 & 0.0 & 0.0 & 0.0 & 0.0 & 0.0 & 0.0 & 0.141 \\ 
			\textbf{0.560} & 0.045 & 0.0 & -0.022 & 0.142 & 0.0 & -0.067 & 0.211 & -0.111 & 0.0 & 0.0 & 0.0 & 0.0 & 0.0 & 0.0 & 0.0 & 0.0 & 0.0 & 0.0 & 0.198 \\ 
			\textbf{0.500} & 0.044 & 0.0 & -0.021 & 0.139 & 0.0 & -0.066 & 0.334 & 0.0 & -0.156 & 0.0 & 0.0 & 0.0 & 0.0 & 0.0 & 0.0 & 0.0 & 0.0 & 0.0 & 0.274 \\ 
			\textbf{0.447} & 0.043 & 0.0 & -0.021 & 0.135 & 0.0 & -0.064 & 0.325 & 0.0 & -0.151 & 0.089 & 0.0 & 0.0 & 0.0 & 0.0 & 0.0 & 0.0 & 0.0 & 0.0 & 0.356 \\ 
			\textbf{0.391} & 0.041 & 0.0 & -0.02 & 0.129 & 0.0 & -0.061 & 0.31 & 0.0 & -0.144 & 0.415 & -0.217 & 0.0 & 0.0 & 0.0 & 0.0 & 0.0 & 0.0 & 0.0 & 0.452 \\ 
			\textbf{0.334} & 0.038 & 0.0 & -0.018 & 0.119 & 0.0 & -0.057 & 0.288 & 0.0 & -0.134 & 0.6 & 0.0 & -0.277 & 0.0 & 0.0 & 0.0 & 0.0 & 0.0 & 0.0 & 0.559 \\ 
			\textbf{0.273} & 0.034 & 0.0 & -0.016 & 0.106 & 0.0 & -0.05 & 0.255 & 0.0 & -0.119 & 0.533 & 0.0 & -0.246 & 0.178 & 0.0 & 0.0 & 0.0 & 0.0 & 0.0 & 0.675 \\ 
			\textbf{0.217} & 0.029 & 0.0 & -0.014 & 0.09 & 0.0 & -0.043 & 0.217 & 0.0 & -0.101 & 0.452 & 0.0 & -0.209 & 0.749 & -0.393 & 0.0 & 0.0 & 0.0 & 0.0 & 0.778 \\ 
			\textbf{0.167} & 0.023 & 0.0 & -0.011 & 0.073 & 0.0 & -0.035 & 0.176 & 0.0 & -0.082 & 0.368 & 0.0 & -0.17 & 0.962 & 0.0 & -0.445 & 0.0 & 0.0 & 0.0 & 0.859 \\ 
			\textbf{0.044} & 0.007 & 0.0 & -0.003 & 0.022 & 0.0 & -0.01 & 0.053 & 0.0 & -0.025 & 0.11 & 0.0 & -0.051 & 0.288 & 0.0 & -0.133 & 0.73 & 0.0 & 0.0 & 0.987 \\ 
			\textbf{0.009} & 0.002 & 0.0 & -0.001 & 0.006 & 0.0 & -0.003 & 0.014 & 0.0 & -0.007 & 0.029 & 0.0 & -0.013 & 0.076 & 0.0 & -0.035 & 2.235 & -1.304 & 0.0 & 0.999 \\ 
			\textbf{0.001} & 0.0 & 0.0 & -0.0 & 0.002 & 0.0 & -0.001 & 0.004 & 0.0 & -0.002 & 0.008 & 0.0 & -0.004 & 0.02 & 0.0 & -0.009 & 2.116 & 0.0 & -1.134 & 1.0 \\ 
			\bottomrule
		\end{tabular}
		\end{adjustbox}
		\label{table:dpmsolverpp3s_coeff_18}
	\end{table}

	\FloatBarrier
	\section{SD3's coefficient matrix and inference process visualization}
	
	\subsection{Coefficient matrix and its corresponding outputs}
	
	\begin{table}[!ht]
		\centering
		\caption{SD3's signal coefficient matrix for Flow Matching Euler sampling} 
		\begin{adjustbox}{width=1\textwidth}
		\begin{tabular}{ccccccccccccccccccccccccccccc}
			\toprule
			\textbf{time} & \textbf{1.00} & \textbf{0.99} & \textbf{0.97} & \textbf{0.96} & \textbf{0.95} & \textbf{0.93} & \textbf{0.91} & \textbf{0.90} & \textbf{0.88} & \textbf{0.86} & \textbf{0.84} & \textbf{0.81} & \textbf{0.79} & \textbf{0.76} & \textbf{0.74} & \textbf{0.71} & \textbf{0.68} & \textbf{0.64} & \textbf{0.60} & \textbf{0.56} & \textbf{0.52} & \textbf{0.46} & \textbf{0.41} & \textbf{0.35} & \textbf{0.28} & \textbf{0.20} & \textbf{0.11} & \textbf{0.01} \\
			\midrule
			\textbf{0.99} & 1.26 & 0.0 & 0.0 & 0.0 & 0.0 & 0.0 & 0.0 & 0.0 & 0.0 & 0.0 & 0.0 & 0.0 & 0.0 & 0.0 & 0.0 & 0.0 & 0.0 & 0.0 & 0.0 & 0.0 & 0.0 & 0.0 & 0.0 & 0.0 & 0.0 & 0.0 & 0.0 & 0.0 \\ 
			\textbf{0.97} & 1.26 & 1.33 & 0.0 & 0.0 & 0.0 & 0.0 & 0.0 & 0.0 & 0.0 & 0.0 & 0.0 & 0.0 & 0.0 & 0.0 & 0.0 & 0.0 & 0.0 & 0.0 & 0.0 & 0.0 & 0.0 & 0.0 & 0.0 & 0.0 & 0.0 & 0.0 & 0.0 & 0.0 \\ 
			\textbf{0.96} & 1.26 & 1.33 & 1.4 & 0.0 & 0.0 & 0.0 & 0.0 & 0.0 & 0.0 & 0.0 & 0.0 & 0.0 & 0.0 & 0.0 & 0.0 & 0.0 & 0.0 & 0.0 & 0.0 & 0.0 & 0.0 & 0.0 & 0.0 & 0.0 & 0.0 & 0.0 & 0.0 & 0.0 \\ 
			\textbf{0.95} & 1.26 & 1.33 & 1.4 & 1.47 & 0.0 & 0.0 & 0.0 & 0.0 & 0.0 & 0.0 & 0.0 & 0.0 & 0.0 & 0.0 & 0.0 & 0.0 & 0.0 & 0.0 & 0.0 & 0.0 & 0.0 & 0.0 & 0.0 & 0.0 & 0.0 & 0.0 & 0.0 & 0.0 \\ 
			\textbf{0.93} & 1.26 & 1.33 & 1.4 & 1.47 & 1.56 & 0.0 & 0.0 & 0.0 & 0.0 & 0.0 & 0.0 & 0.0 & 0.0 & 0.0 & 0.0 & 0.0 & 0.0 & 0.0 & 0.0 & 0.0 & 0.0 & 0.0 & 0.0 & 0.0 & 0.0 & 0.0 & 0.0 & 0.0 \\ 
			\textbf{0.91} & 1.26 & 1.33 & 1.4 & 1.47 & 1.56 & 1.65 & 0.0 & 0.0 & 0.0 & 0.0 & 0.0 & 0.0 & 0.0 & 0.0 & 0.0 & 0.0 & 0.0 & 0.0 & 0.0 & 0.0 & 0.0 & 0.0 & 0.0 & 0.0 & 0.0 & 0.0 & 0.0 & 0.0 \\ 
			\textbf{0.90} & 1.26 & 1.33 & 1.4 & 1.47 & 1.56 & 1.65 & 1.74 & 0.0 & 0.0 & 0.0 & 0.0 & 0.0 & 0.0 & 0.0 & 0.0 & 0.0 & 0.0 & 0.0 & 0.0 & 0.0 & 0.0 & 0.0 & 0.0 & 0.0 & 0.0 & 0.0 & 0.0 & 0.0 \\ 
			\textbf{0.88} & 1.26 & 1.33 & 1.4 & 1.47 & 1.56 & 1.65 & 1.74 & 1.85 & 0.0 & 0.0 & 0.0 & 0.0 & 0.0 & 0.0 & 0.0 & 0.0 & 0.0 & 0.0 & 0.0 & 0.0 & 0.0 & 0.0 & 0.0 & 0.0 & 0.0 & 0.0 & 0.0 & 0.0 \\ 
			\textbf{0.86} & 1.26 & 1.33 & 1.4 & 1.47 & 1.56 & 1.65 & 1.74 & 1.85 & 1.97 & 0.0 & 0.0 & 0.0 & 0.0 & 0.0 & 0.0 & 0.0 & 0.0 & 0.0 & 0.0 & 0.0 & 0.0 & 0.0 & 0.0 & 0.0 & 0.0 & 0.0 & 0.0 & 0.0 \\ 
			\textbf{0.84} & 1.26 & 1.33 & 1.4 & 1.47 & 1.56 & 1.65 & 1.74 & 1.85 & 1.97 & 2.1 & 0.0 & 0.0 & 0.0 & 0.0 & 0.0 & 0.0 & 0.0 & 0.0 & 0.0 & 0.0 & 0.0 & 0.0 & 0.0 & 0.0 & 0.0 & 0.0 & 0.0 & 0.0 \\ 
			\textbf{0.81} & 1.26 & 1.33 & 1.4 & 1.47 & 1.56 & 1.65 & 1.74 & 1.85 & 1.97 & 2.1 & 2.24 & 0.0 & 0.0 & 0.0 & 0.0 & 0.0 & 0.0 & 0.0 & 0.0 & 0.0 & 0.0 & 0.0 & 0.0 & 0.0 & 0.0 & 0.0 & 0.0 & 0.0 \\ 
			\textbf{0.79} & 1.26 & 1.33 & 1.4 & 1.47 & 1.56 & 1.65 & 1.74 & 1.85 & 1.97 & 2.1 & 2.24 & 2.4 & 0.0 & 0.0 & 0.0 & 0.0 & 0.0 & 0.0 & 0.0 & 0.0 & 0.0 & 0.0 & 0.0 & 0.0 & 0.0 & 0.0 & 0.0 & 0.0 \\ 
			\textbf{0.76} & 1.26 & 1.33 & 1.4 & 1.47 & 1.56 & 1.65 & 1.74 & 1.85 & 1.97 & 2.1 & 2.24 & 2.4 & 2.57 & 0.0 & 0.0 & 0.0 & 0.0 & 0.0 & 0.0 & 0.0 & 0.0 & 0.0 & 0.0 & 0.0 & 0.0 & 0.0 & 0.0 & 0.0 \\ 
			\textbf{0.74} & 1.26 & 1.33 & 1.4 & 1.47 & 1.56 & 1.65 & 1.74 & 1.85 & 1.97 & 2.1 & 2.24 & 2.4 & 2.57 & 2.76 & 0.0 & 0.0 & 0.0 & 0.0 & 0.0 & 0.0 & 0.0 & 0.0 & 0.0 & 0.0 & 0.0 & 0.0 & 0.0 & 0.0 \\ 
			\textbf{0.71} & 1.26 & 1.33 & 1.4 & 1.47 & 1.56 & 1.65 & 1.74 & 1.85 & 1.97 & 2.1 & 2.24 & 2.4 & 2.57 & 2.76 & 2.98 & 0.0 & 0.0 & 0.0 & 0.0 & 0.0 & 0.0 & 0.0 & 0.0 & 0.0 & 0.0 & 0.0 & 0.0 & 0.0 \\ 
			\textbf{0.68} & 1.26 & 1.33 & 1.4 & 1.47 & 1.56 & 1.65 & 1.74 & 1.85 & 1.97 & 2.1 & 2.24 & 2.4 & 2.57 & 2.76 & 2.98 & 3.22 & 0.0 & 0.0 & 0.0 & 0.0 & 0.0 & 0.0 & 0.0 & 0.0 & 0.0 & 0.0 & 0.0 & 0.0 \\ 
			\textbf{0.64} & 1.26 & 1.33 & 1.4 & 1.47 & 1.56 & 1.65 & 1.74 & 1.85 & 1.97 & 2.1 & 2.24 & 2.4 & 2.57 & 2.76 & 2.98 & 3.22 & 3.49 & 0.0 & 0.0 & 0.0 & 0.0 & 0.0 & 0.0 & 0.0 & 0.0 & 0.0 & 0.0 & 0.0 \\ 
			\textbf{0.60} & 1.26 & 1.33 & 1.4 & 1.47 & 1.56 & 1.65 & 1.74 & 1.85 & 1.97 & 2.1 & 2.24 & 2.4 & 2.57 & 2.76 & 2.98 & 3.22 & 3.49 & 3.8 & 0.0 & 0.0 & 0.0 & 0.0 & 0.0 & 0.0 & 0.0 & 0.0 & 0.0 & 0.0 \\ 
			\textbf{0.56} & 1.26 & 1.33 & 1.4 & 1.47 & 1.56 & 1.65 & 1.74 & 1.85 & 1.97 & 2.1 & 2.24 & 2.4 & 2.57 & 2.76 & 2.98 & 3.22 & 3.49 & 3.8 & 4.15 & 0.0 & 0.0 & 0.0 & 0.0 & 0.0 & 0.0 & 0.0 & 0.0 & 0.0 \\ 
			\textbf{0.52} & 1.26 & 1.33 & 1.4 & 1.47 & 1.56 & 1.65 & 1.74 & 1.85 & 1.97 & 2.1 & 2.24 & 2.4 & 2.57 & 2.76 & 2.98 & 3.22 & 3.49 & 3.8 & 4.15 & 4.56 & 0.0 & 0.0 & 0.0 & 0.0 & 0.0 & 0.0 & 0.0 & 0.0 \\ 
			\textbf{0.46} & 1.26 & 1.33 & 1.4 & 1.47 & 1.56 & 1.65 & 1.74 & 1.85 & 1.97 & 2.1 & 2.24 & 2.4 & 2.57 & 2.76 & 2.98 & 3.22 & 3.49 & 3.8 & 4.15 & 4.56 & 5.02 & 0.0 & 0.0 & 0.0 & 0.0 & 0.0 & 0.0 & 0.0 \\ 
			\textbf{0.41} & 1.26 & 1.33 & 1.4 & 1.47 & 1.56 & 1.65 & 1.74 & 1.85 & 1.97 & 2.1 & 2.24 & 2.4 & 2.57 & 2.76 & 2.98 & 3.22 & 3.49 & 3.8 & 4.15 & 4.56 & 5.02 & 5.56 & 0.0 & 0.0 & 0.0 & 0.0 & 0.0 & 0.0 \\ 
			\textbf{0.35} & 1.26 & 1.33 & 1.4 & 1.47 & 1.56 & 1.65 & 1.74 & 1.85 & 1.97 & 2.1 & 2.24 & 2.4 & 2.57 & 2.76 & 2.98 & 3.22 & 3.49 & 3.8 & 4.15 & 4.56 & 5.02 & 5.56 & 6.19 & 0.0 & 0.0 & 0.0 & 0.0 & 0.0 \\ 
			\textbf{0.28} & 1.26 & 1.33 & 1.4 & 1.47 & 1.56 & 1.65 & 1.74 & 1.85 & 1.97 & 2.1 & 2.24 & 2.4 & 2.57 & 2.76 & 2.98 & 3.22 & 3.49 & 3.8 & 4.15 & 4.56 & 5.02 & 5.56 & 6.19 & 6.93 & 0.0 & 0.0 & 0.0 & 0.0 \\ 
			\textbf{0.20} & 1.26 & 1.33 & 1.4 & 1.47 & 1.56 & 1.65 & 1.74 & 1.85 & 1.97 & 2.1 & 2.24 & 2.4 & 2.57 & 2.76 & 2.98 & 3.22 & 3.49 & 3.8 & 4.15 & 4.56 & 5.02 & 5.56 & 6.19 & 6.93 & 7.82 & 0.0 & 0.0 & 0.0 \\ 
			\textbf{0.11} & 1.26 & 1.33 & 1.4 & 1.47 & 1.56 & 1.65 & 1.74 & 1.85 & 1.97 & 2.1 & 2.24 & 2.4 & 2.57 & 2.76 & 2.98 & 3.22 & 3.49 & 3.8 & 4.15 & 4.56 & 5.02 & 5.56 & 6.19 & 6.93 & 7.82 & 8.89 & 0.0 & 0.0 \\ 
			\textbf{0.01} & 1.26 & 1.33 & 1.4 & 1.47 & 1.56 & 1.65 & 1.74 & 1.85 & 1.97 & 2.1 & 2.24 & 2.4 & 2.57 & 2.76 & 2.98 & 3.22 & 3.49 & 3.8 & 4.15 & 4.56 & 5.02 & 5.56 & 6.19 & 6.93 & 7.82 & 8.89 & 10.2 & 0.0 \\ 
			\textbf{0.00} & 1.26 & 1.33 & 1.4 & 1.47 & 1.56 & 1.65 & 1.74 & 1.85 & 1.97 & 2.1 & 2.24 & 2.4 & 2.57 & 2.76 & 2.98 & 3.22 & 3.49 & 3.8 & 4.15 & 4.56 & 5.02 & 5.56 & 6.19 & 6.93 & 7.82 & 8.89 & 10.2 & 0.89 \\
			\bottomrule
		\end{tabular}
		\end{adjustbox}
		\label{table:euler_28}
	\end{table}
	
	Note that, for readability, the coefficients in Table \ref{table:euler_28} are the original coefficients multiplied by 100. When using them, they should be normalized to the corresponding Marginal Coefficient for each step. For example, the first row should be normalized to 0.0126, and the second row should be normalized to 0.0259. The usage of Table \ref{table:adjusted_euler_28} is the same.
	
	\begin{table}[!ht]
		\centering
		\caption{SD3's signal coefficient matrix with more sharpness} 
		\begin{adjustbox}{width=1\textwidth}
		\begin{tabular}{ccccccccccccccccccccccccccccc}
			\toprule
			\textbf{time} & \textbf{1.00} & \textbf{0.99} & \textbf{0.97} & \textbf{0.96} & \textbf{0.95} & \textbf{0.93} & \textbf{0.91} & \textbf{0.90} & \textbf{0.88} & \textbf{0.86} & \textbf{0.84} & \textbf{0.81} & \textbf{0.79} & \textbf{0.76} & \textbf{0.74} & \textbf{0.71} & \textbf{0.68} & \textbf{0.64} & \textbf{0.60} & \textbf{0.56} & \textbf{0.52} & \textbf{0.46} & \textbf{0.41} & \textbf{0.35} & \textbf{0.28} & \textbf{0.20} & \textbf{0.11} & \textbf{0.01} \\
			\midrule
			\textbf{0.99} & 1.26 & 0.0 & 0.0 & 0.0 & 0.0 & 0.0 & 0.0 & 0.0 & 0.0 & 0.0 & 0.0 & 0.0 & 0.0 & 0.0 & 0.0 & 0.0 & 0.0 & 0.0 & 0.0 & 0.0 & 0.0 & 0.0 & 0.0 & 0.0 & 0.0 & 0.0 & 0.0 & 0 \\ 
			\textbf{0.97} & 1.26 & 1.33 & 0.0 & 0.0 & 0.0 & 0.0 & 0.0 & 0.0 & 0.0 & 0.0 & 0.0 & 0.0 & 0.0 & 0.0 & 0.0 & 0.0 & 0.0 & 0.0 & 0.0 & 0.0 & 0.0 & 0.0 & 0.0 & 0.0 & 0.0 & 0.0 & 0.0 & 0 \\ 
			\textbf{0.96} & 1.26 & 1.33 & 1.4 & 0.0 & 0.0 & 0.0 & 0.0 & 0.0 & 0.0 & 0.0 & 0.0 & 0.0 & 0.0 & 0.0 & 0.0 & 0.0 & 0.0 & 0.0 & 0.0 & 0.0 & 0.0 & 0.0 & 0.0 & 0.0 & 0.0 & 0.0 & 0.0 & 0 \\ 
			\textbf{0.95} & 0.0 & 1.33 & 1.4 & 1.47 & 0.0 & 0.0 & 0.0 & 0.0 & 0.0 & 0.0 & 0.0 & 0.0 & 0.0 & 0.0 & 0.0 & 0.0 & 0.0 & 0.0 & 0.0 & 0.0 & 0.0 & 0.0 & 0.0 & 0.0 & 0.0 & 0.0 & 0.0 & 0 \\ 
			\textbf{0.93} & 0.0 & 1.33 & 1.4 & 1.47 & 1.56 & 0.0 & 0.0 & 0.0 & 0.0 & 0.0 & 0.0 & 0.0 & 0.0 & 0.0 & 0.0 & 0.0 & 0.0 & 0.0 & 0.0 & 0.0 & 0.0 & 0.0 & 0.0 & 0.0 & 0.0 & 0.0 & 0.0 & 0 \\ 
			\textbf{0.91} & 0.0 & 0.0 & 1.44 & 1.56 & 1.56 & 1.65 & 0.0 & 0.0 & 0.0 & 0.0 & 0.0 & 0.0 & 0.0 & 0.0 & 0.0 & 0.0 & 0.0 & 0.0 & 0.0 & 0.0 & 0.0 & 0.0 & 0.0 & 0.0 & 0.0 & 0.0 & 0.0 & 0 \\ 
			\textbf{0.90} & 0.0 & 0.0 & 0.0 & 0.0 & 1.56 & 1.65 & 1.74 & 0.0 & 0.0 & 0.0 & 0.0 & 0.0 & 0.0 & 0.0 & 0.0 & 0.0 & 0.0 & 0.0 & 0.0 & 0.0 & 0.0 & 0.0 & 0.0 & 0.0 & 0.0 & 0.0 & 0.0 & 0 \\ 
			\textbf{0.88} & 0.0 & 0.0 & 0.0 & 0.0 & 0.0 & 1.65 & 1.74 & 1.85 & 0.0 & 0.0 & 0.0 & 0.0 & 0.0 & 0.0 & 0.0 & 0.0 & 0.0 & 0.0 & 0.0 & 0.0 & 0.0 & 0.0 & 0.0 & 0.0 & 0.0 & 0.0 & 0.0 & 0 \\ 
			\textbf{0.86} & 0.0 & 0.0 & 0.0 & 0.0 & 0.0 & 1.65 & 1.74 & 1.85 & 1.97 & 0.0 & 0.0 & 0.0 & 0.0 & 0.0 & 0.0 & 0.0 & 0.0 & 0.0 & 0.0 & 0.0 & 0.0 & 0.0 & 0.0 & 0.0 & 0.0 & 0.0 & 0.0 & 0 \\ 
			\textbf{0.84} & 0.0 & 0.0 & 0.0 & 0.0 & 0.0 & 0.0 & 1.74 & 1.85 & 1.97 & 2.1 & 0.0 & 0.0 & 0.0 & 0.0 & 0.0 & 0.0 & 0.0 & 0.0 & 0.0 & 0.0 & 0.0 & 0.0 & 0.0 & 0.0 & 0.0 & 0.0 & 0.0 & 0 \\ 
			\textbf{0.81} & 0.0 & 0.0 & 0.0 & 0.0 & 0.0 & 0.0 & 0.0 & 1.85 & 1.97 & 2.1 & 2.24 & 0.0 & 0.0 & 0.0 & 0.0 & 0.0 & 0.0 & 0.0 & 0.0 & 0.0 & 0.0 & 0.0 & 0.0 & 0.0 & 0.0 & 0.0 & 0.0 & 0 \\ 
			\textbf{0.79} & 0.0 & 0.0 & 0.0 & 0.0 & 0.0 & 0.0 & 0.0 & 0.0 & 1.97 & 2.1 & 2.24 & 2.4 & 0.0 & 0.0 & 0.0 & 0.0 & 0.0 & 0.0 & 0.0 & 0.0 & 0.0 & 0.0 & 0.0 & 0.0 & 0.0 & 0.0 & 0.0 & 0 \\ 
			\textbf{0.76} & 0.0 & 0.0 & 0.0 & 0.0 & 0.0 & 0.0 & 0.0 & 0.0 & 0.0 & 2.1 & 2.24 & 2.4 & 2.57 & 0.0 & 0.0 & 0.0 & 0.0 & 0.0 & 0.0 & 0.0 & 0.0 & 0.0 & 0.0 & 0.0 & 0.0 & 0.0 & 0.0 & 0 \\ 
			\textbf{0.74} & 0.0 & 0.0 & 0.0 & 0.0 & 0.0 & 0.0 & 0.0 & 0.0 & 0.0 & 2.1 & 2.24 & 2.4 & 2.57 & 2.76 & 0.0 & 0.0 & 0.0 & 0.0 & 0.0 & 0.0 & 0.0 & 0.0 & 0.0 & 0.0 & 0.0 & 0.0 & 0.0 & 0 \\ 
			\textbf{0.71} & 0.0 & 0.0 & 0.0 & 0.0 & 0.0 & 0.0 & 0.0 & 0.0 & 0.0 & 2.1 & 2.24 & 2.4 & 2.57 & 2.76 & 2.98 & 0.0 & 0.0 & 0.0 & 0.0 & 0.0 & 0.0 & 0.0 & 0.0 & 0.0 & 0.0 & 0.0 & 0.0 & 0 \\ 
			\textbf{0.68} & 0.0 & 0.0 & 0.0 & 0.0 & 0.0 & 0.0 & 0.0 & 0.0 & 0.0 & 2.1 & 2.24 & 2.4 & 2.57 & 2.76 & 2.98 & 3.22 & 0.0 & 0.0 & 0.0 & 0.0 & 0.0 & 0.0 & 0.0 & 0.0 & 0.0 & 0.0 & 0.0 & 0 \\ 
			\textbf{0.64} & 0.0 & 0.0 & 0.0 & 0.0 & 0.0 & 0.0 & 0.0 & 0.0 & 0.0 & 2.1 & 2.24 & 2.4 & 2.57 & 2.76 & 2.98 & 3.22 & 3.49 & 0.0 & 0.0 & 0.0 & 0.0 & 0.0 & 0.0 & 0.0 & 0.0 & 0.0 & 0.0 & 0 \\ 
			\textbf{0.60} & 0.0 & 0.0 & 0.0 & 0.0 & 0.0 & 0.0 & 0.0 & 0.0 & 0.0 & 2.1 & 2.24 & 2.4 & 2.57 & 2.76 & 2.98 & 3.22 & 3.49 & 3.8 & 0.0 & 0.0 & 0.0 & 0.0 & 0.0 & 0.0 & 0.0 & 0.0 & 0.0 & 0 \\ 
			\textbf{0.56} & 0.0 & 0.0 & 0.0 & 0.0 & 0.0 & 0.0 & 0.0 & 0.0 & 0.0 & 0.0 & 2.24 & 2.4 & 2.57 & 2.76 & 2.98 & 3.22 & 3.49 & 3.8 & 4.15 & 0.0 & 0.0 & 0.0 & 0.0 & 0.0 & 0.0 & 0.0 & 0.0 & 0 \\ 
			\textbf{0.52} & 0.0 & 0.0 & 0.0 & 0.0 & 0.0 & 0.0 & 0.0 & 0.0 & 0.0 & 0.0 & 2.24 & 2.4 & 2.57 & 2.76 & 2.98 & 3.22 & 3.49 & 3.8 & 4.15 & 4.56 & 0.0 & 0.0 & 0.0 & 0.0 & 0.0 & 0.0 & 0.0 & 0 \\ 
			\textbf{0.46} & 0.0 & 0.0 & 0.0 & 0.0 & 0.0 & 0.0 & 0.0 & 0.0 & 0.0 & 0.0 & 2.24 & 2.4 & 2.57 & 2.76 & 2.98 & 3.22 & 3.49 & 3.8 & 4.15 & 4.56 & 5.02 & 0.0 & 0.0 & 0.0 & 0.0 & 0.0 & 0.0 & 0 \\ 
			\textbf{0.41} & 0.0 & 0.0 & 0.0 & 0.0 & 0.0 & 0.0 & 0.0 & 0.0 & 0.0 & 0.0 & 2.24 & 2.4 & 2.57 & 2.76 & 2.98 & 3.22 & 3.49 & 3.8 & 4.15 & 4.56 & 5.02 & 5.56 & 0.0 & 0.0 & 0.0 & 0.0 & 0.0 & 0 \\ 
			\textbf{0.35} & 0.0 & 0.0 & 0.0 & 0.0 & 0.0 & 0.0 & 0.0 & 0.0 & 0.0 & 0.0 & 0.0 & 0.0 & 2.57 & 2.76 & 2.98 & 3.22 & 3.49 & 3.8 & 4.15 & 4.56 & 5.02 & 5.56 & 6.19 & 0.0 & 0.0 & 0.0 & 0.0 & 0 \\ 
			\textbf{0.28} & 0.0 & 0.0 & 0.0 & 0.0 & 0.0 & 0.0 & 0.0 & 0.0 & 0.0 & 0.0 & 0.0 & 0.0 & 2.57 & 2.76 & 2.98 & 3.22 & 3.49 & 3.8 & 4.15 & 4.56 & 5.02 & 5.56 & 6.19 & 6.93 & 0.0 & 0.0 & 0.0 & 0 \\ 
			\textbf{0.20} & 0.0 & 0.0 & 0.0 & 0.0 & 0.0 & 0.0 & 0.0 & 0.0 & 0.0 & 0.0 & 0.0 & 0.0 & 2.57 & 2.76 & 2.98 & 3.22 & 3.49 & 3.8 & 4.15 & 4.56 & 5.02 & 5.56 & 6.19 & 6.93 & 7.82 & 0.0 & 0.0 & 0 \\ 
			\textbf{0.11} & 0.0 & 0.0 & 0.0 & 0.0 & 0.0 & 0.0 & 0.0 & 0.0 & 0.0 & 0.0 & 0.0 & 0.0 & 2.57 & 2.76 & 2.98 & 3.22 & 3.49 & 3.8 & 4.15 & 4.56 & 5.02 & 5.56 & 6.19 & 6.93 & 7.82 & 8.89 & 0.0 & 0 \\ 
			\textbf{0.01} & 0.0 & 0.0 & 0.0 & 0.0 & 0.0 & 0.0 & 0.0 & 0.0 & 0.0 & 0.0 & 0.0 & 0.0 & 0.0 & 0.0 & 2.98 & 3.22 & 3.49 & 3.8 & 4.15 & 4.56 & 5.02 & 5.56 & 6.19 & 6.93 & 7.82 & 8.89 & 10.2 & 0 \\ 
			\textbf{0.00} & 0.0 & 0.0 & 0.0 & 0.0 & 0.0 & 0.0 & 0.0 & 0.0 & 0.0 & 0.0 & 0.0 & 0.0 & 0.0 & 0.0 & 2.98 & 3.22 & 3.49 & 3.8 & 4.15 & 4.56 & 5.02 & 5.56 & 6.19 & 6.93 & 7.82 & 8.89 & 10.2 & 15 \\ 
			\bottomrule
		\end{tabular}
		\end{adjustbox}
		\label{table:adjusted_euler_28}
	\end{table}	

	\begin{figure}
		\centering
		\begin{subfigure}[]{\includegraphics[width=0.99\textwidth]{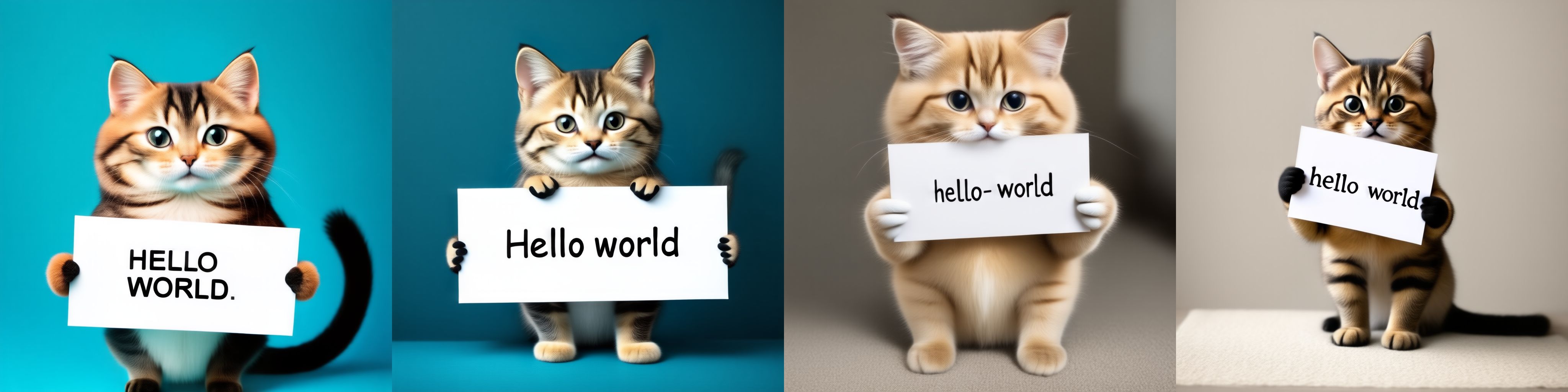}} \end{subfigure}
		\begin{subfigure}[]{\includegraphics[width=0.99\textwidth]{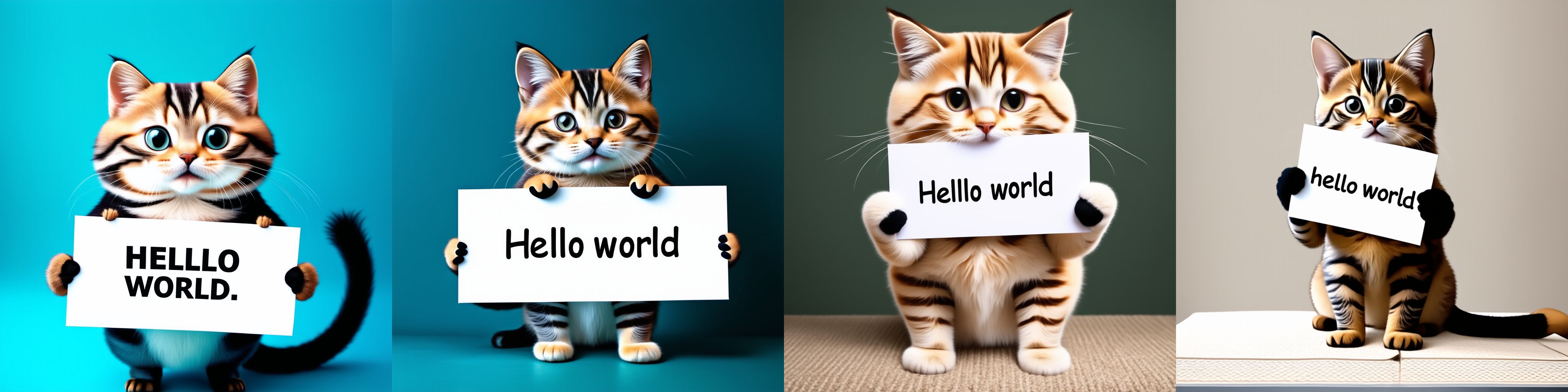}} \end{subfigure}
		\caption{(a) Result for original Euler sampling (Table \ref{table:euler_28}) \quad (b) Result for adjusted coefficient matrix (Table \ref{table:adjusted_euler_28})} 
		\label{figure:adjusted coeff matrix}
	\end{figure}

	\subsection{ Mid Self Guidance is better suited for large CFG} \label{section:large_cfg}

\begin{figure}
		\centering
		\begin{subfigure}[]{\includegraphics[width=0.98\textwidth]{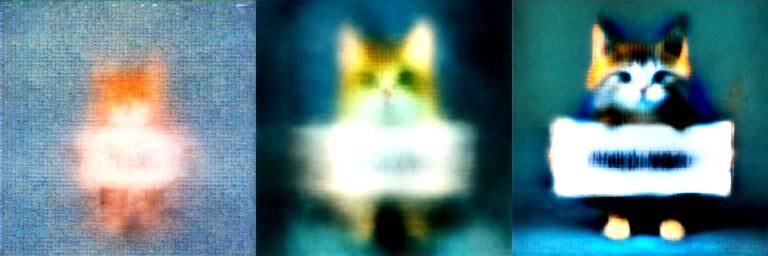}} \end{subfigure}
		\begin{subfigure}[]{\includegraphics[width=0.32\textwidth]{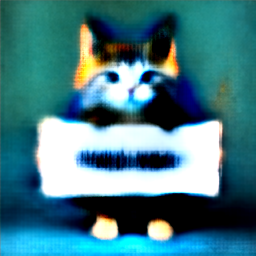}} \end{subfigure}
		\begin{subfigure}[]{\includegraphics[width=0.32\textwidth]{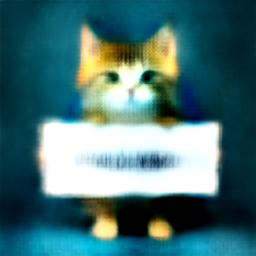}} \end{subfigure}
		\begin{subfigure}[]{\includegraphics[width=0.32\textwidth]{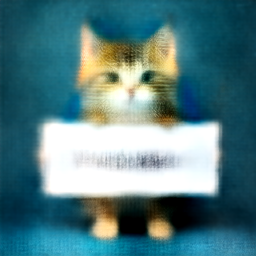}} \end{subfigure}
		\caption{(a) The first three output ($\mlg{x}^0_0$, $\mlg{x}^1_0$, $\mlg{x}^2_0$)  \qquad (b) $0.00\cdot\mlg{x}^0_0-0.17\cdot\mlg{x}^1_0+1.17\cdot\mlg{x}^2_0$  \newline (c) $0.00\cdot\mlg{x}^0_0+0.49\cdot\mlg{x}^1_0+0.51\cdot\mlg{x}^2_0$ \qquad (d) $0.32\cdot\mlg{x}^0_0+0.33\cdot\mlg{x}^1_0+0.35\cdot\mlg{x}^2_0$} 
		\label{figure:cfg_sen}
	\end{figure}

	Figure \ref{figure:cfg_sen} presents a small experiment demonstrating that, the output from Mid Self Guidance significantly outperforms Fore Self Guidance when using large CFG values. Figure \ref{figure:cfg_sen}(a) shows the outputs for the first three steps of SD3 (cfg=7), and Figures \ref{figure:cfg_sen}(b)$-$(d) show results from different linear combinations. (b) using negative coefficients, (c) using two positive coefficients, and (d) using three positive coefficients. It is clear that when there are more positive coefficients, the image quality is better.

	\FloatBarrier
	\subsection{Inference process visualization}
	
\begin{figure}[h]
		\centering \includegraphics[width=1.0\textwidth]{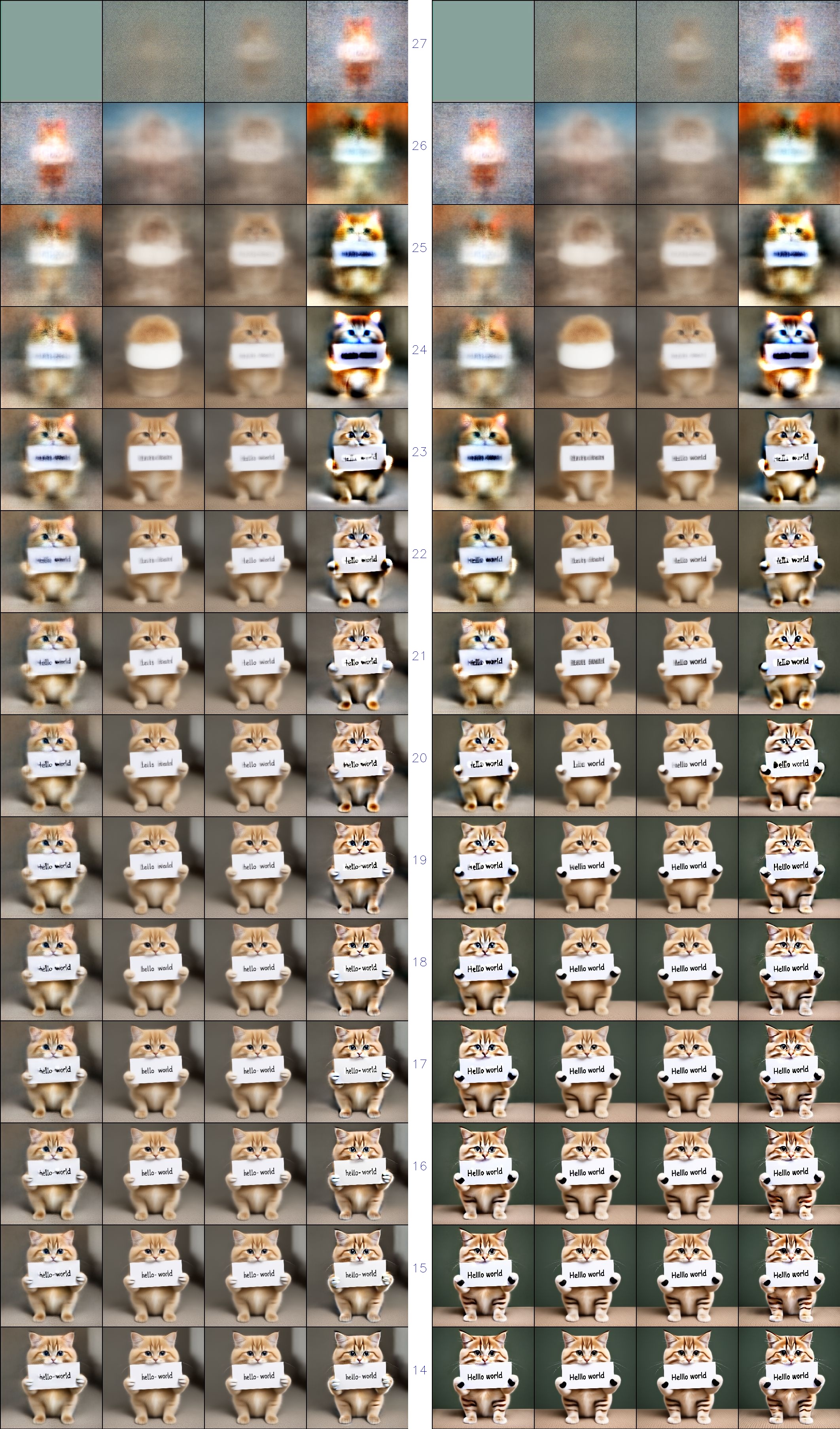} \centering \caption{Inference process visualization: first half.} \label{figure:sharp_process_0}
	\end{figure}

\begin{figure}[h]
		\centering \includegraphics[width=1.0\textwidth]{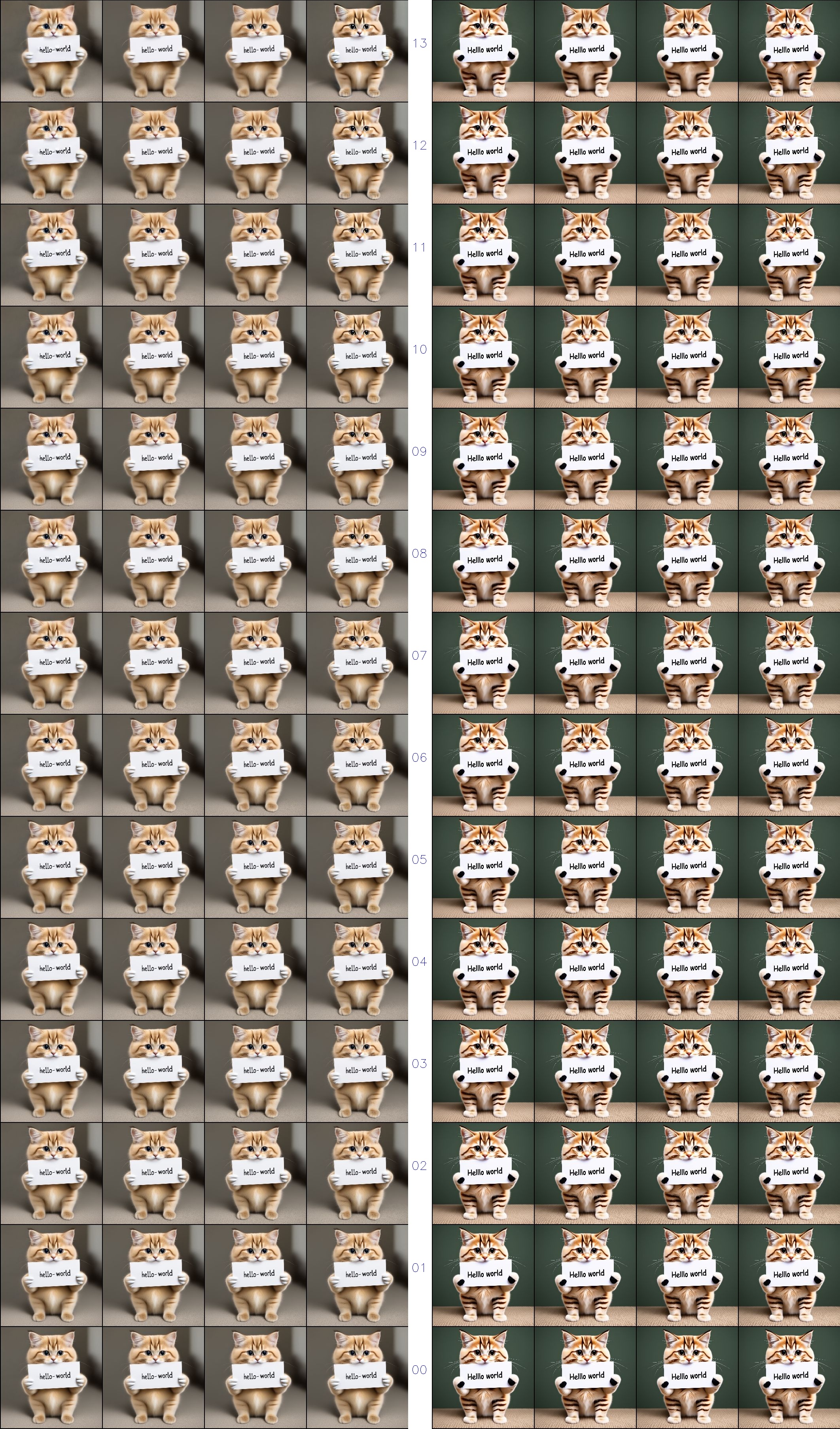} \centering \caption{Inference process visualization: second half} \label{figure:sharp_process_1}
	\end{figure}

	Figures \ref{figure:sharp_process_0} and \ref{figure:sharp_process_1} provide a visualization of the complete inference process. The left half shows the inference process using the coefficient matrix from Table \ref{table:euler_28}, and the right half shows the inference process using the coefficient matrix from Table \ref{table:adjusted_euler_28}. The first column shows the result of Self Guidance, which is also the input image signal to the model. The second column shows the model output without conditioning, the third column shows the conditioned model output, and the fourth column shows the result of Classifier Free Guidance. For each model operation, there is a clear image signal input and image signal output, which greatly enhances intuitive understanding of the operation's purpose and facilitates efficient debugging and problem analysis.
	
	\FloatBarrier
	\section{Optimized coefficient matrix for pretrained CIFAR10 model}
	
	To clearly understand the relative proportions of the coefficients in each row, the coefficients in Table \ref{table:opt_coeff_5} have been scaled to ensure that the diagonal elements equal 1. When using these coefficients, each row needs to be normalized to its corresponding Marginal Coefficient for each step. For example, the first row should be normalized to 0.118, and the second row should be normalized to 0.487. The same normalization process applies to Tables \ref{table:opt_coeff_10} and \ref{table:opt_coeff_15}.
	
	\begin{table}[!ht]
		\centering
		\caption{optimized coefficient matrix for 5 step} 
		\begin{tabular}{ccccccc}
			\toprule
			\textbf{time} & \textbf{1.000} & \textbf{0.650} & \textbf{0.375} & \textbf{0.176} & \textbf{0.051} & \textbf{marginal coeff} \\
			\midrule
			\textbf{0.650} & 1 & 0 & 0 & 0 & 0 & 0.118 \\ 
			\textbf{0.375} & -0.291 & 1 & 0 & 0 & 0 & 0.487 \\ 
			\textbf{0.176} & 0 & 0.133 & 1 & 0 & 0 & 0.85 \\ 
			\textbf{0.051} & 0 & 0 & -0.337 & 1 & 0 & 0.985 \\ 
			\textbf{0.001} & 0 & 0 & 0 & -0.583 & 1 & 1 \\
			\bottomrule 
		\end{tabular}
		\label{table:opt_coeff_5}
	\end{table}
	
	\begin{table}[!ht]
		\centering
		\caption{optimized coefficient matrix for 10 step} 
		\begin{adjustbox}{width=1\textwidth}
		\begin{tabular}{cccccccccccccc}
			\toprule
			\textbf{time} & \textbf{1.000} & \textbf{0.816} & \textbf{0.650} & \textbf{0.503} & \textbf{0.375} & \textbf{0.266} & \textbf{0.176} & \textbf{0.104} & \textbf{0.051} & \textbf{0.017} & \textbf{marginal coeff} \\
			\midrule
			\textbf{0.816} & 1 & 0 & 0 & 0 & 0 & 0 & 0 & 0 & 0 & 0 & 0.035 \\ 
			\textbf{0.650} & 0 & 1 & 0 & 0 & 0 & 0 & 0 & 0 & 0 & 0 & 0.118 \\ 
			\textbf{0.503} & 0 & -0.3 & 1 & 0 & 0 & 0 & 0 & 0 & 0 & 0 & 0.276 \\ 
			\textbf{0.375} & 0 & 0.52 & -0.3 & 1 & 0 & 0 & 0 & 0 & 0 & 0 & 0.487 \\ 
			\textbf{0.266} & 0 & 0.2 & 0.4 & -0.3 & 1 & 0 & 0 & 0 & 0 & 0 & 0.694 \\ 
			\textbf{0.176} & 0 & 0 & 0.2 & 0.4 & -0.2 & 1 & 0 & 0 & 0 & 0 & 0.85 \\ 
			\textbf{0.104} & 0 & 0 & 0 & 0 & 0.35 & -0.15 & 1 & 0 & 0 & 0 & 0.943 \\ 
			\textbf{0.051} & 0 & 0 & 0 & 0 & 0 & 0.37 & -0.2 & 1 & 0 & 0 & 0.985 \\ 
			\textbf{0.017} & 0 & 0 & 0 & 0 & 0 & 0 & 0.1 & -0.33 & 1 & 0 & 0.998 \\ 
			\textbf{0.001} & 0 & 0 & 0 & 0 & 0 & 0 & 0 & 0.05 & -0.34 & 1 & 1 \\ 
			\bottomrule
		\end{tabular}
		\end{adjustbox}
		\label{table:opt_coeff_10}
	\end{table}
	
	\begin{table}[!ht]
		\centering
		\caption{optimized coefficient matrix for 15 step} 
		\begin{adjustbox}{width=1\textwidth}
		\begin{tabular}{cccccccccccccccccc}
			\toprule
			\textbf{time} & \textbf{1.000} & \textbf{0.875} & \textbf{0.758} & \textbf{0.650} & \textbf{0.550} & \textbf{0.459} & \textbf{0.375} & \textbf{0.300} & \textbf{0.234} & \textbf{0.176} & \textbf{0.126} & \textbf{0.084} & \textbf{0.051} & \textbf{0.026} & \textbf{0.009} & \textbf{marginal coeff} \\ \midrule
			\textbf{0.875} & 1 & 0 & 0 & 0 & 0 & 0 & 0 & 0 & 0 & 0 & 0 & 0 & 0 & 0 & 0 & 0.021 \\ 
			\textbf{0.758} & 0.24 & 1 & 0 & 0 & 0 & 0 & 0 & 0 & 0 & 0 & 0 & 0 & 0 & 0 & 0 & 0.055 \\ 
			\textbf{0.650} & 0.1 & 0.48 & 1 & 0 & 0 & 0 & 0 & 0 & 0 & 0 & 0 & 0 & 0 & 0 & 0 & 0.118 \\ 
			\textbf{0.550} & 0 & 0.2 & 0.41 & 1 & 0 & 0 & 0 & 0 & 0 & 0 & 0 & 0 & 0 & 0 & 0 & 0.216 \\ 
			\textbf{0.459} & 0 & 0 & 0.2 & -0.2 & 1 & 0 & 0 & 0 & 0 & 0 & 0 & 0 & 0 & 0 & 0 & 0.343 \\ 
			\textbf{0.375} & 0 & 0.3 & -0.14 & 0.56 & -0.77 & 1 & 0 & 0 & 0 & 0 & 0 & 0 & 0 & 0 & 0 & 0.487 \\ 
			\textbf{0.300} & 0 & 0 & 0.23 & -0.06 & 0.3 & -0.85 & 1 & 0 & 0 & 0 & 0 & 0 & 0 & 0 & 0 & 0.629 \\ 
			\textbf{0.234} & 0 & 0 & 0 & 0.26 & -0.01 & 0.85 & -0.86 & 1 & 0 & 0 & 0 & 0 & 0 & 0 & 0 & 0.753 \\ 
			\textbf{0.176} & 0 & 0 & 0 & 0 & 0.25 & 0 & 0.82 & -0.78 & 1 & 0 & 0 & 0 & 0 & 0 & 0 & 0.85 \\ 
			\textbf{0.126} & 0 & 0 & 0 & 0 & 0 & 0.23 & -0.02 & 0.9 & -0.2 & 1 & 0 & 0 & 0 & 0 & 0 & 0.919 \\ 
			\textbf{0.084} & 0 & 0 & 0 & 0 & 0 & 0 & 0.2 & -0.04 & 0.7 & -0.4 & 1 & 0 & 0 & 0 & 0 & 0.961 \\ 
			\textbf{0.051} & 0 & 0 & 0 & 0 & 0 & 0 & 0 & 0.17 & -0.07 & 0.62 & -0.66 & 1 & 0 & 0 & 0 & 0.985 \\ 
			\textbf{0.026} & 0 & 0 & 0 & 0 & 0 & 0 & 0 & 0 & 0.14 & -0.09 & 0.57 & -0.88 & 1 & 0 & 0 & 0.995 \\ 
			\textbf{0.009} & 0 & 0 & 0 & 0 & 0 & 0 & 0 & 0 & 0 & 0.11 & -0.11 & 0.31 & -0.49 & 1 & 0 & 0.999 \\ 
			\textbf{0.001} & 0 & 0 & 0 & 0 & 0 & 0 & 0 & 0 & 0 & 0 & 0.08 & -0.11 & 0.24 & -0.31 & 1 & 1 \\ 
			\bottomrule
		\end{tabular}
		\end{adjustbox}
		\label{table:opt_coeff_15}
	\end{table}

\end{document}